\documentclass[a4paper,12pt]{article}
\usepackage{lineno,hyperref}
\usepackage{array,multirow,graphicx}
\usepackage{float}
\usepackage{graphicx}
\usepackage{afterpage}
\usepackage{mathrsfs}  
\usepackage{caption}
\usepackage{amsmath,amssymb,amsthm}
\usepackage{subcaption}
\usepackage{color}
\usepackage{rotating,booktabs}
\usepackage{amsfonts}
\usepackage{indentfirst}
\usepackage{mathptmx}
\usepackage{amsmath}
\usepackage{lmodern}
\usepackage{relsize,exscale}
\usepackage[doublespacing]{setspace}
\usepackage{lipsum}
\usepackage{titlesec}
\titleformat{\section}
    [block]{\normalfont\fontfamily{phv}\fontsize{16}{19}\bfseries\Large}{\rlap{\thesection}}{0em}
    {\hspace{.05\linewidth}\begin{minipage}[t]{.95\textwidth}}[\end{minipage}]
    
\titleformat*{\subsection}{\normalfont\fontsize{12}{13}\bfseries}
\titleformat*{\subsubsection}{\normalfont\fontsize{12}{13}\selectfont}

\usepackage{relsize,exscale}

\newtheorem{theorem}{Theorem}
\newtheorem{corollary}{Corollary}
\newtheorem{lemma}{Lemma}

\newcommand{\eqdef}{\overset{\mathrm{def}}{=\joinrel=}}

\usepackage{titlesec}
\titleformat*{\section}{\centering\normalfont\bfseries}
\makeatletter
\renewcommand\@biblabel[1]{}
\makeatother
\modulolinenumbers[5]

\usepackage{amsmath}

\newcommand{\argmax}{\mathop{\mathrm{argmax}}\limits}

\usepackage[symbol*]{footmisc}
\DefineFNsymbolsTM{myfnsymbols}{
	\textdagger    \dagger
	\textdaggerdbl \ddagger
	\textsection   \mathsection
	\textbardbl    \|%
	\textparagraph \mathparagraph
	\textasteriskcentered *
}
\setfnsymbol{myfnsymbols}

\usepackage{fancyhdr}
\fancyheadoffset{0cm}

\fancypagestyle{plain}{%
	\fancyhf{}%
	\fancyhead[R]{\thepage}%
}
\usepackage{blindtext}
\usepackage{algorithm2e}

\parskip=\medskipamount

\begin{document}

\thispagestyle{empty}
\begin{center}
{\fontsize{14}{17} \selectfont 
\textbf{A Theory of Dichotomous Valuation with  \\ Applications to Variable Selection\footnote{To cite this article: Xingwei Hu (2020): A theory of dichotomous valuation with applications to
		variable selection, Econometric Reviews, DOI: 10.1080/07474938.2020.1735750}\footnote{The views expressed herein are those of the author and should not be attributed to the IMF, its Executive Board, or its management.}}
}

\vspace{2cm}
Xingwei Hu\footnote{
	Email: xhu@imf.org; \
	Phone: 202-623-8317; \
	Fax: 202-589-8317.
} 

\vskip .5cm
International Monetary Fund \\
700 19th St NW, Washington, DC 20431, USA
\date{}
\end{center}

\newpage
\noindent \textbf{Abstract:}

\vspace{.2cm}\noindent
An econometric or statistical model may undergo a marginal gain if we admit a new variable to the model, and a marginal loss if we remove an existing variable from the model. Assuming equality of opportunity among all candidate variables, we derive a valuation framework by the expected marginal gain and marginal loss in all potential modeling scenarios. However, marginal gain and loss are not symmetric; thus, we introduce three unbiased solutions. When used in variable selection, our new approaches significantly outperform several popular methods used in practice. The results also explore some novel traits of the Shapley value. 

\vspace{.6cm}
\begin{center}
\begin{minipage}{5.2in}
\begin{itemize}
\setlength\itemsep{-6pt}
\item[\texttt{Keywords}: ]
variable selection;\
feature selection;\
Shapley value; \
endowment bias;\
over-fitting
\item[\texttt{JEL Codes}:]
C11,\ C52,\ C57,\ C71,\ D81
\end{itemize}
\end{minipage}
\end{center}




\newpage

\section{Introduction}

\noindent The problem we address relates to the following typical situation: 
when modeling data, we attempt to use a formula to simplify the underlying process that has been operating to produce the data. 
To be useful, the formula or model should not only help us better understand the underlying structure of the variables in the specific past, but also be predictive in a non-specific situation in the future.
Ideally, the model should perform well under multiple scenarios. 
In general, the variables may be interdependent. 
Some variables may even explain another one, making the explained one superfluous. 
Other variables may be entirely irrelevant to the underlying process. 
Selecting the right variables or features is one of the most fundamental processes in statistics, econometrics, and machine learning.
Also, people subconsciously select data for their daily personal or business decision-making. 
This paper argues that we can improve the selection accuracy by dichotomously evaluating the behavioral bias in the data selection process.

The results in this paper arise from two interlinked considerations. 
We first separate any marginal effect into  either a marginal gain or a marginal loss by an ownership relation.
For a motivating example, consider the marginal effect  of a bachelor's  degree (abbreviated ``BD") on the 
annual income of an individual aged $40$. The individual may have a BD or not.
For an individual with a BD, the marginal gain is computed as the difference between his current annual income and
his estimated annual income, assuming he had no BD, \textit{ceteris paribus}.  
For an individual with no BD, on the other hand, the marginal loss is computed as the difference between 
his estimated annual income, assuming  he had a BD, and his current annual income, \textit{ceteris paribus}. 
We note that the possession and the worth of a BD interweave
with other factors, such as his profession and length of relevant work experience, which also affect his income. 
We measure the value of a BD by the expected marginal gain and marginal loss, 
incorporating the ownership uncertainty and interdependence with other factors.
Our second consideration is the implicitly embedded bias when we admit a candidate variable to the model or remove it from the model. 
In the above example, a BD holder may show some endowment bias, valuing the BD more than those who do not have one.
We define the bias as the expected difference between the marginal gain and 
the marginal loss. We then design mechanisms to mitigate the bias. 
In the mechanism design, the end goals are set as constraint equations for the valuation solutions.

The approach we employ in this paper is Bayesian and game-theoretic.
For a given performance or objective function, our goal is to find a small set of variables that have high importance concerning  the performance criterion. 
We first use the performance function as the payoffs on all subsets of candidate variables to set up a coalitional game. 
In the game, we evaluate each player.
Secondly, unlike many algorithm-based approaches that search for a subset of variables with optimal collective performance, we incorporate rather than ignore model uncertainty. 
The search for a single model, however, ignores the model uncertainty (e.g., Clyde and George, 2004). 
Thus, we directly evaluate the \textit{overall performance} of each variable in a broad set of modeling scenarios. 
Then, we use their performance to select variables and institute the model. 
In this process, some prior beliefs or distributions are required to specify the possibility of modeling scenarios. 
In this sense, our approach is also Bayesian and uses model averaging (e.g., George and McCulloch, 1997;  O'Hara and Sillanpaa, 2009).
For a particular class of non-informative priors, the overall performance happens to be the renown Shapley value (1953) or the Banzhaf value (1965) in the coalitional game. 

The advantage of our approach is threefold. First, 
by dichotomizing the marginal effect and acknowledging the model uncertainty, we generalize the
Shapley value and the Banzhaf value under the same framework. 
They differ in prior distributions, which quantify the likelihood of modeling scenarios, but
their prior distributions are not unique.
Next, by  also using the dichotomization, we introduce a concept of bias and discover the symmetry in the Banzhaf value and asymmetry in the Shapley value. We
suggest a simple way to adjust the asymmetry. 
Simulations show that our bias-adjusted solutions perform exceptionally well, compared with a few variable selection methods used in practice, such as
stepwise regression, information-based subset search, Lasso (Tibshirani, 1996), and its adaptive version (Zou, 2006).
Lastly, we introduce a new value solution based on binomial distributions,
which embraces the Banzhaf value and which links the Shapley value by compounding.
It is as tractable in expression and calculation as the Shapley value and the Banzhaf value. 
The new value allows the expected model size from the probability distribution to be consistent with the actual model.

Our research aims at applying economic theories to machine learning, statistics, and econometrics.
These theories include Keynes' principle of indifference, cooperative game theory, diminishing marginality, and behavioral economics.
In return, we develop the Shapley value in a data modeling and forecast context; we also supply solutions to reduce endowment biases in behavioral economics.
Recently, the use of the Shapley value and cooperative game have gained popularity in modeling data, partly due to its simplicity and generality
(e.g.,  Lipovetsky and Conklin, 2001; Cohen et al., 2007; Israeli, 2007; Gromping, 2007; Devicienti, 2010; Strumbelj and Kononenko, 2010 \& 2014; Raja et al., 2016). 
Moreover, the Shapley value allows heterogeneous or nonlinear marginal contributions in different coalitional situations.
The vast literature also provides many variations on the Shapley value (e.g., Monderer and Samet, 2002; Winter, 2002).
Our research offers not only a few new derivations and new interpretations of the Shapley value, but also
a theoretical foundation for appropriately using the value concept in variable selection and related fields.
Besides, Theorem \ref{thm:sv} in this paper was previously proved in Hu (2002).
The first two theorems of this paper are generalizations of the results in Hu (2006), in which unanimous support of the bill to vote is not necessary, and the payoff function is binary. 
A closely related work in Hu (2018) applies a similar valuation approach to fairly allocate unemployment benefits and to derive a fair real-time tax rate.

We organize the remainder of the paper as follows: 
Section 2 formulates the essential ideas of dichotomized marginal effect and relates them to the Shapley value and the Banzhaf value. 
Section 3 analyzes the difference, called \textit{endowment bias}, between the expectations of marginal gain and marginal loss.
This section also proposes three unbiased solutions.
Section 4 discusses how to implement the basic ideas with a sequential algorithm and how to modify the algorithm to select the right features. 
In Section 5, we compare the simulation performance of our four new solutions with eight other variable selection methods.
We also study the accuracy of the new methods in situations with correlated covariates, correlated residuals, or large sample sizes.
Finally, in Section 6, we relate this framework to several other topics in economics, political science, and statistics. 
We also suggest a few directions extend the framework.
Our exposition is self-contained, and the proofs are in the Supplemental Appendix.

\section{Evaluation of Candidate Predictors}\label{sect:two_marginals}

\noindent Before our formal discussions, we introduce a few notations.
Let $\mathbb{N} = \{1, 2,\cdots,n\}$ denote the set of all candidate variables, indexed as $1,2,..., n$. 
For any $T\subseteq \mathbb{N}$, let $v(T)$ be a performance measure or performance function when we model the data using the variables in $T$.
In particular, for the empty set $\emptyset$, $v(\emptyset)$  is the performance  
when the model does not involve any variables from $\mathbb{N}$.
We use  ``$\setminus$" for set subtraction and ``$\cup$" for set union. 
The vinculum (overbar) is used in naming the elements of a set; for example,  ``$\overline{i}$"  for the singleton set $\{ i \}$ and ``$\overline{ik}$" for the set $\{i, k \}$.
For any subset $T\subseteq \mathbb{N}$, let $|T|$ denote its cardinality.

For a simple and concrete example, we consider a linear regression model for the dependent variable $Y_j$ where $j$ is the index for the observations in the data.
Let there be $n$ candidate explanatory variables $X_{ji}$, $i=1,2,...,n$.
The performance measure could be a model fit statistic, the variance explained, the forecast accuracy, 
the cost involved, the probability of avoiding fatal errors, or any combination of the above. 
In particular, $v(\emptyset)$ is the performance when $Y_j$ is modeled by a constant.
Without loss of generality, we could even extend $v$ to be a vector-valued set function, which does not weaken the results in the next three sections.  
As a consequence, variable selection eventually becomes a multi-criteria decision analysis,  mitigating the discrepancy between model accuracy in estimation and model usefulness in the forecast.

\subsection{\textit{Model Uncertainty}}
\noindent Data modeling is subject to both objective and subjective uncertainty.
For the objective one, the models and assumptions we impose on the data are inevitably misspecified in some way.
Hence, an econometric or statistical model is merely an approximation and simplification of reality. 
There could be many good approximations under various criteria. 
This multiplicity requires us to look into the data in various scenarios before reaching a decisive decision.
For the subjective uncertainty, we do not know exactly which specific variables to choose before performing a selection analysis.
We may have a class of subjective probabilities for it, derived from our judgment, opinions, experience, or even fairness assumptions.
We make fairness assumptions in this paper and form inference by addressing each modeling scenario separately.
To address the model uncertainty, we let the random subset $\mathbf{S}\subseteq \mathbb{N}$ consist of variables 
in the actual model.

To formulate general selection methods, we assume no prior knowledge about the relation between the covariates.
Without being given any specific prior knowledge, we have no reason to believe that one set of 
candidate variables is more likely to be $\mathbf{S}$ than another set of the same size. We should not discriminate between sets of variables
having the same size. 
Given the size of $\mathbf{S}$, in other words, we have no reason to select one variable and reject another; the consequence is that
each variable is equally likely to be selected. 
This equality of opportunity or rule of epistemic probabilities can be formally justified by Leibniz's \textit{principle of insufficient reason} or Keynes' \textit{principle of indifference}. 

We let $\mu$ be the probability distribution of $\mathbf{S}$ and let $P_{_T}$ be the probability of $\mathbf{S} = T$ for any $T\subseteq \mathbb{N}$. 
Let us define a class of probability distributions with an equal opportunity:

\vskip .3cm
\begin{center}
$\mathscr{F}\ \eqdef$\  {\Large \{} $\mu | P_{_T}$ is a function depending only on the size of $T$ {\Large \}}.
\end{center}
\vskip .3cm

\noindent In the above, we use the notation ``$\eqdef$" for definition. 
For any $t=0, 1, \cdots, n$, let 
$$
\delta_t \ \eqdef \ \sum\limits_{T\subseteq \mathbb{N}: |T|=t} \ P_{_T}
$$
which is the probability of the model size $|{\mathbf S}|$ being $t$. 
For any $\mu \in \mathscr{F}$, we can easily find the probability density $P_{_T} = \frac{(|T|)!(n-|T|)!}{n!} \delta_{_{|T|}}$, as there are 
$\frac{n!}{(|T|)!(n-|T|)!}$ subsets of size $|T|$.
Thus, when $\mu \in \mathscr{F}$, the set of $\{ \delta_t \}_{t=0}^n$ determines a specific prior $\mu$.
However, unlike other Bayesian methods, we do not specify a particular prior distribution. 
Instead, we merely set up a prior rule which corresponds to a particular class of prior distributions for $\mu$.
In Bayesian probability, this is a set of non-informative priors.

Let us introduce three special subclasses of $\mathscr{F}$ indexed by $\eta$,
$$
\begin{array}{rcll}
\mu[\eta; \mathrm{SV}] 	&:& P_{_T} = \frac{(|T|)!(n-|T|)!}{(n+1)!} + (-1)^{|T|}\eta, \quad &\forall \ \ T\subseteq \mathbb{N}, \\

\mu[\eta; \mathrm{BV}] 	&:& P_{_T} = \frac{1}{2^n} + (-1)^{|T|} \eta, \quad & \forall \ \ T\subseteq \mathbb{N}, \\

\mu[\eta; \mathrm{BN}] 	&:& P_{_T} =  \eta^{|T|} (1-\eta)^{n-|T|}, \quad & \forall \ \ T\subseteq \mathbb{N}.
\end{array}
$$
For each subclass, $\eta$ is chosen such that  all $P_{_T}$ are non-negative. 
Also, it is not hard to verify that $P_{_T}$ over all $T\subseteq \mathbb{N}$ sum to $1$ in the subclass.
Thus, $\mu[\eta; \mathrm{SV}]$, $\mu[\eta;\mathrm{BV}]$, 
or $\mu[\eta; \mathrm{BN}]$ is a probability distribution, respectively.
Here,  SV, BV, and BN are shorthands for ``Shapley value", ``Banzhaf value", and ``Binomial", respectively.
For the prior $\mu[\eta; \mathrm{SV}]$, $\delta_t = \frac{1}{n+1} + (-1)^t \eta
\left (
\begin{array}{c}
n \\
t
\end{array}
\right )$ for all $t=0,1,\cdots, n$. 
The distribution $\mu[\eta; \mathrm{BV}]$ is a variant of $\mu[\frac{1}{2}; \mathrm{BN}]$ with
$\delta_t = \left [ \frac{1}{2^n}+(-1)^t \eta \right ]
\left (
\begin{array}{c}
n \\
t
\end{array}
\right ).
$
For the prior $\mu[\eta; \mathrm{BN}]$, the model size
$|\mathbf{S}|$ has a binomial distribution with parameters $(n,\eta)$, i.e., 
$\delta_t = 
\left (
\begin{array}{c}
n \\
t
\end{array}
\right )
\eta^t (1-\eta)^{n-t}
$.
For its own sake, it extends $\mu[0; \mathrm{BV}]$.

\subsection{\textit{Dichotomous Valuation}}
\noindent
Hu (2002, 2006) introduced the dichotomized marginal effect into coalitional games to provide a new characterization of the Shapley value and the Shapley-Shubik power index (Shapley and Shubik, 1954).
In the power index setting, a voter may have two pivotal situations: either turning a losing vote to winning or vice versa. 
The probability of such pivotal situations measures the power of the voter. 
In the current setting, however, the performance function $v$ takes values not merely equal to $0$ or $1$.

Let us formally analyze the two-sided marginal effect for each candidate variable in $\mathbb{N}$.
For the indeterminate model $\mathbf{S}$, we could add one variable from $\mathbb{N}\setminus \mathbf{S}$ to the model; we could also remove an existing variable from the model $\mathbf{S}$. 
The variable's marginal effect to $v(\mathbf{S})$ explains the worth of the addition or removal; the effect is dependent on the cooperation between $i$ and the members in $\mathbf{S}$.
There are two jointly exhaustive and mutually exclusive scenarios between $i$ and $\mathbf{S}$:

\begin{center}
\begin{minipage}{5.8in}
\begin{itemize}\itemsep2pt
\item Scenario 1: $i \in {\mathbf S}$. Then,
$i$'s marginal effect is $v({\mathbf S}) - v( {\mathbf S} \setminus \overline{i})$, called \textit{marginal gain}, in that
it contributes $v({\mathbf S}) - v( {\mathbf S} \setminus \overline{i})$ 
due to its existence in the model ${\mathbf S}$. 
The expected marginal gain is 
\begin{equation}\label{eq:gamma}
\gamma_i[v;\mu] \
\eqdef \ \mathbb{E} \left [v({\mathbf S}) - v({\mathbf S} \setminus \overline{i}) \right ].
\end{equation}
The notation ``$\mathbb{E}[\cdot]$" here is for expectation under the probability measure $\mu$.

\item 
Scenario 2: $i \not \in {\mathbf S}$. Then, the marginal effect is
$v({\mathbf S}\cup \overline{i}) - v({\mathbf S})$ in that ${\mathbf S}$ faces a \textit{marginal loss} or opportunity cost
$v({\mathbf S}\cup \overline{i}) - v({\mathbf S})$ without variable $i$ in the model ${\mathbf S}$. 
In other words, the variable could have increased the collective performance by $v({\mathbf S}\cup \overline{i} ) - v({\mathbf S})$ if we had added
it to $\mathbf{S}$. The expected marginal loss, due to $i$'s absence from ${\mathbf S}$, is 
\begin{equation}\label{eq:lambda}
\lambda_i[v;\mu] \
\eqdef \ \mathbb{E} \left [v({\mathbf S}\cup \overline{i}) - v({\mathbf S}) \right ].
\end{equation}
\end{itemize}
\end{minipage}
\end{center}

\noindent In either case, the marginal effect of $i$ to $v(\mathbf{S})$ 
can be written as $v({\mathbf S}\cup \overline{i}) - v({\mathbf S}\setminus \overline{i})$.
Combining these two exclusive marginals, we define variable $i$'s \textit{dichotomous value}  $\psi_i[v;\mu]$ (hereinafter, ``\textit{D-value}"), as a functional function of $v$,
by its expected marginal effect under the probability distribution $\mu$:
\begin{equation} \label{eq:D-Value}
\psi_i[v;\mu] \
\eqdef \ \gamma_i[v;\mu] + \lambda_i[v;\mu] 
=\mathbb{E} \left [v({\mathbf S}\cup \overline{i}) - v({\mathbf S}\setminus \overline{i}) \right ].
\end{equation}

One additional justification for the dichotomization is that the performance of $i$ is not a constant; it is conditional on with whom $i$ works. 
The D-value $\psi_i[v;\mu]$ quantifies variable $i$'s expected conditional performance under the distribution $\mu$,
which specifies the likelihood for the situational conditions of ${\mathbf S}$. 
When fairness is stipulated in the equal opportunity assumption for these scenarios,
the D-value and its variants fairly weight the conditional performance.
In the following exposition, $\psi[v; \mu]$ ($\gamma[v; \mu]$ or $\lambda[v; \mu]$) denotes the $n$-by-$1$ column vector with elements $\psi_i[v; \mu]$ ($\gamma_i[v; \mu]$ or $\lambda_i[v; \mu]$, respectively), $i\in \mathbb{N}$.

As a function of functions, the functional $\psi[v; \mu]$ ($\gamma [v;\mu]$ or $\lambda [v;\mu]$, respectively) takes the function $v$ as its input argument.
The parameter $\mu$ provides \textit{de facto} a weighting scheme for the valuation rule $\psi$ ($\gamma$ or $\lambda$, respectively).
In Sections \ref{sect:eBias} and \ref{subsect:beyongSVBV}, we re-weight or even de-weight the weighting scheme such that the expected marginal gain and loss are balanced, or their sums satisfy certain functional equations.
Formally, one could derive (\ref{eq:D-Value}) from the following two axioms:
\begin{center}
\begin{minipage}{5.8in}
\begin{itemize}\itemsep1pt
\item Marginality: given $P_{_T}=1$, $\psi_i[v;\mu] = v(T\cup \overline{i})-v(T\setminus \overline{i})$;

\item Linearity: given any probability distributions $\mu_1$ and $\mu_2$ on $2^\mathbb{N}$, 
$$
\psi[v;c \mu_1+(1-c)\mu_2] = c \psi[v;\mu_1]+(1-c)\psi[v;\mu_2]
$$
for any $0\le c \le 1$.
\end{itemize}
\end{minipage}
\end{center}
\vskip .4cm

Two attributes distinguish the definition of D-value from that of probabilistic value (e.g., Weber, 1988) or semivalue (e.g., Dubey et al., 1981). 
First, the D-value separates the marginal gain and the marginal loss so that a bias can be formulated in the next section.
Secondly, the probability distribution $\mu$ is placed on the subsets of $\mathbb{N}$, not on the subsets of $\mathbb{N}\setminus \overline{i}$.
Nevertheless, the D-value could be written as a semivalue or a probabilistic value for some special $\mu$, and vice versa.

In the proofs in the Supplemental Appendix, we often capitalize on the relation that $i$'s marginal gain in $T$ equals its marginal loss to $T\setminus \overline{i}$ for any $i\in T \subseteq \mathbb{N}$.
Similarly, $j$'s marginal loss to $T$ equals its marginal gain in $T\cup \overline{j}$ for any $j\not \in T \subseteq \mathbb{N}$. In addition,
a dummy player, who has zero marginal loss to all $T\subseteq \mathbb{N}$,  has zero D-value no matter what the $\mu$. 
Two symmetric players $i$ and $j$ (i.e., $v(T\cup \overline{i}) = v(T\cup \overline{j})$ for all $T\subseteq \mathbb{N}$ such that $i, j\not \in T$) may  have different D-values;
when $\mu \in \mathscr{F}$, however, $\gamma_i[v;\mu]=\gamma_j[v;\mu]$, $\lambda_i[v;\mu]=\lambda_j[v;\mu]$, and $\psi_i[v;\mu]=\psi_j[v;\mu]$ (cf. Hu, 2018).

\subsection{\textit{Beyond the Shapley and Banzhaf Values}}\label{subsect:beyongSVBV}
\noindent One application of the D-value $\psi[v;\mu]$ is to distribute the grand collective performance $v(\mathbb{N})-v(\emptyset)$ 
among all candidate variables. This requires that the weighting scheme $\mu$ satisfies the functional equation of
\begin{equation}\label{eq:totality}
\sum\limits_{i\in \mathbb{N}} \psi_i [v;\mu] \equiv v(\mathbb{N})-v(\emptyset).
\end{equation}
We use ``$\equiv$" to denote a functional equation which holds equal for any payoff function $v$.
Thus, the portion $\psi_i [v;\mu]$ of $v(\mathbb{N})-v(\emptyset)$ is explained by variable $i$.
The following theorem and its corollary relate the D-value $\psi [v;\mu]$ with 
the Shapley value $\Psi[v]$ in the coalitional game $(\mathbb{N}, v)$, defined as (Shapley, 1953):
\begin{equation}\label{eq:shapley_value}
\Psi_i[v]\ \eqdef \ \sum\limits_{T\subseteq \mathbb{N}: \ i \in  T} \frac{(|T|-1)!(n-|T|)!}{n!} \left [v(T) - v(T\setminus \overline{i}) \right ].
\end{equation}

\begin{theorem}\label{thm:sv}
For any $\mu \in \mathscr{F}$ which satisfies (\ref{eq:totality}),  $\psi[v; \mu] \equiv \Psi[v]$ and $\mu = \mu[\eta; \mathrm{SV}]$
for some $\eta$.
\end{theorem}

\noindent \textit{Proof}: See Supplementary Appendix SA-1.

\begin{corollary} \label{cor:svinverse}
If  $\mu = \mu[\eta; \mathrm{SV}]$ for some $\eta$, then $\psi[v;\mu] \equiv \Psi[v]$.
\end{corollary}

\noindent \textit{Proof}: See Supplementary Appendix SA-2.

As $\psi[v;\mu]$ is linear and satisfies the dummy player property, 
it is not surprising that $\psi[v;\mu]$ equals $\Psi[v]$ for some particular class of symmetric $\mu$.
With equation (\ref{eq:totality}), Theorem \ref{thm:sv} accurately figures out these peculiar distributions. 
As shown in the theorem and its corollary, $\mu$ is not even unique. 
In deriving the proofs, we assume neither $v(\emptyset) = 0$, nor monotonicity of $v$, 
nor super-additivity of $v$ (i.e., $v(S \cup T) \ge v(S)+v(T)$ for any disjoint subsets $S$ and $T$ of $\mathbb{N}$).
All these assumptions are highly artificial in a practical data modeling situation. 
For example, if $v(T)$ is the F-statistic when modeling $Y_j$ by the regressors in $T$,
then neither $v(\emptyset)=0$ nor super-additivity is a valid assumption.

In the game theory literature, the Banzhzf value  (1965) does not satisfy (\ref{eq:totality}). It is defined as 
\begin{equation}\label{eq:banzhaf_value}
b_i[v]\ \eqdef \ \frac{1}{2^{n-1}} \sum\limits_{T\subseteq \mathbb{N}:\ i \in T} \left [v(T) - v(T\setminus \overline{i}) \right ].
\end{equation}
In the next theorem, we associate the Banzhaf value with the D-value $\psi[v;\mu]$ through 
the subclass $\mu[\eta; \mathrm{BV}]$.
\begin{theorem}\label{thm:bv} 
$\psi [v;\mu] \equiv b [v]$ if and only if $\mu = \mu[\eta;\mathrm{BV}]$ for some $\eta$ with $|\eta|\le \frac{1}{2^n}$.
\end{theorem}

\noindent \textit{Proof}: See Supplementary Appendix SA-3. 

For both subclasses $\mu[\eta; \mathrm{SV}]$ and $\mu[\eta;\mathrm{BV}]$, 
the model $\mathbf{S}$ has an expected size $\frac{n}{2}$.
This expectation could be highly unrealistic, particularly when the D-value is applied to big data analytics where the model size is relatively small compared to a large number of candidate variables.
In contrast, for a $\mu[\eta; \mathrm{BN}]$, the model $\mathbf{S}$ has an expected size of $n\eta$. 
In Theorem \ref{thm:binomial}, we write the D-value and its components for the subclass $\mu[\eta; \mathrm{BN}]$.

\begin{theorem}\label{thm:binomial} 
If $\mu=\mu[\eta; \mathrm{BN}]$ for some $\eta\in (0,1)$, then
\begin{equation} \label{eq:binomialPsi}
\begin{array}{rcl}
\psi_i [v; \mu[\eta; \mathrm{BN}]] 
&=& \frac{1}{\eta} \gamma_i [v; \mu[\eta; \mathrm{BN}]] \\
&=& \frac{1}{1-\eta} \lambda_i [v;\mu[\eta; \mathrm{BN}]] \\
&=& \sum\limits_{T\subseteq \mathbb{N}: i \in T} \eta^{|T|-1} (1-\eta)^{n-|T|} \left [ v(T) - v(T\setminus \overline{i}) \right ]. 
\end{array}
\end{equation}
\end{theorem}

\noindent \textit{Proof}: See Supplementary Appendix SA-4.

However, the fair-division solutions to (\ref{eq:totality}) may not fit well to the model identification problem when $\mathbb{N}$ is not the exact model variables.
Either over-fitting or under-fitting is a counterpart of the exact identification of the actual model.
When irrelevant variables join with relevant ones to model the data generating process, the benefit would be merely a highly inflated fitting statistic without any further explanatory power.
However, the price is the loss of significance to the true model variables and the curse of dimensionality.
On the other hand, a direct consequence of under-fitting is the omitted-variable bias brought to the right model variables.
Thus, we consider an alternative to (\ref{eq:totality}), and distribute $\mathbb{E} v(\mathbf{S}) - v(\emptyset)$ 
among the variables in $\mathbb{N}$; $\mathbb{E} v(\mathbf{S}) - v(\emptyset)$ is the expected collective performance of the actual model.
We expect that the irrelevant variables in $\mathbb{N}\setminus \mathbf{S}$ would have little disturbance to $v(\mathbf{S})$,
compared with their disturbance to $v(\mathbb{N})$. 

Let us try using the functionals $\psi[v;\mu]$, $\gamma[v;\mu]$, and $\lambda[v;\mu]$ to share the expected model performance $\mathbb{E} v(\mathbf{S}) - v(\emptyset)$
among the players in $\mathbb{N}$. The respective functional equations for the distribution of $\mathbb{E} v(\mathbf{S}) - v(\emptyset)$ are
\begin{equation}\label{eq:divide_expectation_psi}
\sum\limits_{i\in \mathbb{N}} \psi_i[v;\mu] \equiv  \mathbb{E} v(\mathbf{S}) - v(\emptyset),
\end{equation}
\begin{equation}\label{eq:divide_expectation_gamma}
\sum\limits_{i\in \mathbb{N}}  \gamma_i[v;\mu] \equiv  \mathbb{E} v(\mathbf{S}) - v(\emptyset),
\end{equation}
\begin{equation}\label{eq:divide_expectation_lambda}
\sum\limits_{i\in \mathbb{N}} \lambda_i[v;\mu] \equiv  \mathbb{E} v(\mathbf{S}) - v(\emptyset).
\end{equation}

Solutions exist in $\mathscr{F}$ to two of the three functional equations, according to Theorem \ref{thm:divide_expectation_lambda}.
(\ref{eq:divide_expectation_lambda}) has a unique solution $\mu \in \mathscr{F}$,
and the solution is the marginal loss component of the Shapley value. Thus,
we should prefer $\lambda[v;\mu[0; \mathrm{SV}]]$ to the Shapley value when addressing the variable selection issue.
Unfortunately, there is no $\mu \in \mathscr{F}$, which satisfies (\ref{eq:divide_expectation_psi}).
In contrast, (\ref{eq:divide_expectation_gamma}) has infinite solutions in $\mathscr{F}$; but any of the solutions concentrates its probability density on  $\delta_0$ and $\delta_1$.
In this case, the fair-division rule is: the winner takes it all.
In theory, only the winner contributes to $v$; in a real data analysis, however, $v(T\cup \overline{i}) \not = v(T)$ for all $i\in \mathbb{N}$ and $T\not \ni i$, due to the sampling error.
Thus, we seek one essential feature from $\mathbb{N}$, which takes the largest share from $\mathbb{E} v(\mathbf{S}) - v(\emptyset)$.
After selecting the essential feature or the most valuable player (MVP), we could continue looking for the next most valuable one from the remaining features, which takes the largest from the reminding share.
We continue this iteration until  the leftover is no longer significant enough to be divided by the remaining features.
By accepting features in succession from the most valuable one to the last acceptable one, we indeed attempt to divide $\mathbb{E} v(\mathbf{S}) - v(\emptyset)$ among the members of $\mathbf{S}$, not the members of $\mathbb{N}$ as specified in (\ref{eq:divide_expectation_psi})$-$(\ref{eq:divide_expectation_lambda}).
Algorithms \ref{alg:UShapley} and \ref{alg:BN} in Section \ref{sect:est_algorthms} reveal the idea of successive acceptance of MVPs.
A straightforward implication is that $\gamma[v;\mu]$, together with the algorithms, could also solve the model identification problem.

\begin{theorem}\label{thm:divide_expectation_lambda} 
Assume $\mu \in \mathscr{F}$. 
Then:
(\ref{eq:divide_expectation_lambda}) holds if and only if $\mu = \mu[0; \mathrm{SV}]$;
(\ref{eq:divide_expectation_gamma}) holds if and only if $\delta_0 +\delta_1 = 1$;
there exists no $\mu \in \mathscr{F}$ which solves (\ref{eq:divide_expectation_psi}).
\end{theorem}

\noindent \textit{Proof}: See Supplementary Appendix SA-5.

Theorem \ref{thm:divide_expectation_lambda} implies that both $\lambda [v; \mu]$ and $\gamma[v;\mu]$ could be fair-division solutions. 
The fairness comes from the equality of opportunity or Keynes' \textit{principle of indifference}.
The payoff to be divided should be $v(\mathbf{S})-v(\emptyset)$; but as it is random, we use its expectation. 
Besides, we revise (\ref{eq:divide_expectation_psi}) so that $ v(\mathbf{S}) - v(\emptyset)$ is fairly shared by the members of $\mathbf{S}$ only. 
By Theorem \ref{thm:sv}, the D-value solution is the Shapley value $\Psi [v_{_\mathbf{S}}]$ where the set function $v_{_Z}(\cdot)$ is defined as 
$$
v_{_Z} (T) \eqdef v(Z \cap T), \quad \mathrm{for}\ \mathrm{any} \ T\subseteq \mathbb{N}.
$$
Any $i$ outside of $\mathbf{S}$ is a dummy player to $v_{_\mathbf{S}}$; hence, it has zero Shapley value in $\Psi [v_{_\mathbf{S}}]$.
Therefore, the expected vector of $\Psi [v_{_\mathbf{S}}]$ also fairly distributes $\mathbb{E} v(\mathbf{S}) - v(\emptyset)$ to the members in $\mathbb{N}$.
By Theorem \ref{thm:expected_SV}, the loss component of the Shapley value is the expectation of $\Psi [v_{_\mathbf{S}}]$
where $\mathbf{S}$ has the prior $\mu[0; \mathrm{SV}]$.

\begin{theorem}\label{thm:expected_SV} 
Given the prior $\mu = \mu[0;\mathrm{SV}]$ for $\mathbf{S}$, then the expected Shapley value 
$$
\mathbb{E} \Psi_i [v_{_\mathbf{S}}] 
= 
\sum\limits_{T\subseteq \mathbb{N}: \ i \in T} \frac{(|T|-1)!(n-|T|+1)!}{(n+1)!} \left [v(T) - v(T\setminus \overline{i}) \right ]
=
\lambda_i[v;\mu[0;\mathrm{SV}]].
$$
And the weights for the marginal gain $\left [ v(T) - v(T\setminus \overline{i})\right ]$ tally to $\frac{1}{2}$.
\end{theorem}

\noindent \textit{Proof}: See Supplementary Appendix SA-6.

Both the marginal gain and loss components de-weight and re-weight the ingredients of $\Psi[v]$. 
Compared with (\ref{eq:shapley_value}), $\lambda_i[v;\mu[0;\mathrm{SV}]]$ (cf. Theorem \ref{thm:expected_SV}) underweights each marginal gain $\left [v(T) - v(T\setminus \overline{i}) \right ]$ by $\frac{|T|}{n+1}$, and the total weights are $\frac{1}{2}$.
When $T$ is close to $\mathbb{N}$, $i$'s contribution to $v(T)$ is almost negligible when $i$ correlates with other covariates, and no matter $i$ is a true model variable or not. 
Thus, $\lambda[v;\mu[0;\mathrm{SV}]]$ heavily underweights these scenarios.
In contrast, we use (\ref{eq:gamma}) and $\mu[0;\mathrm{SV}]$ to write 
$$
\gamma_i[v;\mu[0;\mathrm{SV}]] = \sum\limits_{T\subseteq \mathbb{N}: \ i \in T} \frac{(|T|)!(n-|T|)!}{(n+1)!} \left [v(T) - v(T\setminus \overline{i}) \right ].
$$
Compared with (\ref{eq:shapley_value}), $\gamma_i[v;\mu[0;\mathrm{SV}]]$ underweights each marginal gain $\left [v(T) - v(T\setminus \overline{i}) \right ]$ by $1-\frac{|T|}{n+1}$, and the weights also add up to $\frac{1}{2}$. 
When $T$ is close to $\overline{i}$, $i$'s contribution to $v(T)$ is subject to omitted-variable bias unless $\mathbf{S}=\overline{i}$. 
Thus, $\gamma[v;\mu[0;\mathrm{SV}]]$ heavily underweights these situations.

\section{Endowment Bias}\label{sect:eBias}
\noindent 
In our valuation paradigm in Section \ref{sect:two_marginals}, we classify two types of marginal effect 
according to the ownership: either $i \in {\mathbf S}$ or $i \not \in {\mathbf S}$.
In practice, people tend to ascribe more worth to things they own, 
rather than ones they do not own --- even when things are exchangeable in value. 
This subjective bias is called the \textit{endowment effect} or \textit{endowment bias} (e.g., Thaler, 1980; Kahneman et al., 1990).
When the covariates in $T\subseteq \mathbb{N}$ are used to model the dependent variable $Y_j$, for example,
the significance statistics for each $i\in T$, such as t-statistic and its p-value, are 
vulnerable to selection bias and omitted-variable bias, among others. 
Both selection bias and omitted-variable bias ultimately associate with the mere ownership between individual covariates and the correct model.
In this section, we first define the endowment bias, then analyze the implied positive bias in the Shapley value and the implied zero bias in the Banzhaf value. 
The test statistics in a modeling situation, however, tend to demonstrate negative bias.
Finally, we propose solutions that mitigate these endowment biases.

\subsection{\textit{Zero Bias in the Banzhaf Value}}
\noindent For each $i\in \mathbb{N}$, we define its \textit{endowment bias} $\kappa_i[v;\mu]$ as the difference between 
its expected marginal gain and its expected marginal loss,
$$
\kappa_i[v;\mu] \
\eqdef \ \gamma_i[v;\mu] - \lambda_i[v;\mu].
$$

In the above definition, the uncertainty is only about the relationship between $i$ and $\mathbf{S}$.
In a subjective valuation, we sometimes need to consider another type of risk, which is the subjective uncertainty of the function $v$. 
For example, consider the marginal production of some person $i$ to a firm $\mathbf{S}$. 
On the one hand, if $i$ has already been an employee of the firm for a while (i.e., $i\in \mathbf{S}$), then he or she or the firm would have a clear idea about how much the marginal production is. 
On the other hand, if $i$ is a potential employee (i.e., $i\not \in \mathbf{S}$), the marginal production he or she would bring to the firm is uncertain. 
Given $\mathbf{S}$, therefore, the subjective uncertainty in $v(\mathbf{S} \cup \overline{i})-v(\mathbf{S})$ and $v(\mathbf{S})-v(\mathbf{S} \setminus \overline{i})$ differs:
the former one  is likely to be far more significant than the latter.
Thus, a positive endowment bias $\kappa_i [v; \mu]$ is preferred in a risk-averse valuation, other things remaining unchanged. 
This risk in marginal loss partially explains the vast existence of positive endowment bias. 
Uncertainty about $v$ and subjective valuation, however, are beyond the scope of the current research.

Lemma \ref{lm:bias} provides a method to directly calculate the biases without involving the expected marginal gain or loss.
In this lemma, the weights on $v(T)$ sum to zero, i.e., $\sum\limits_{T\subseteq \mathbb{N}} \left [2 P_{_T} - P_{_{T \cup \overline{i}}} - P_{_{T \setminus \overline{i}}} \right ] = 0$; 
but the bias itself may not.
\begin{lemma}\label{lm:bias}
For any $i\in \mathbb{N}$,
$
\kappa_i [v;\mu] = \sum\limits_{T\subseteq \mathbb{N}} \left [2 P_{_T} - P_{_{T \cup \overline{i}}} - P_{_{T \setminus \overline{i}}} \right ] v(T)
$
\end{lemma}

\noindent \textit{Proof}: See Supplementary Appendix SA-7.

In this section, we consider three forms of endowment unbiasedness. In a strong form, $\kappa_i [v;\mu] \equiv 0$  for all $i\in \mathbb{N}$.
The following theorem characterizes this type of endowment unbiasedness by the Banzhaf value.
For two other subclasses of $\mathscr{F}$, there is only one parameter $\eta$, and
it could be unattainable for the parameter to make all $\kappa_i [v;\mu]$ to zeros. 
We thus seek a weak form of unbiasedness (\ref{eq:unbiasedValue}) by unevenly weighting the two components of $\kappa_i [v;\mu]$ and setting the aggregate bias to zero, 
i.e., $\sum\limits_{i\in \mathbb{N}} \kappa_i [v;\mu] \equiv 0$. 
When $\mu = \mu[\eta; \mathrm{BN}]$, an explicit formula is available for this form of unbiasedness (cf. Theorem \ref{thm:binomialbias}), which is indexed by $\eta$.
To smoothen the dependence on the unknown $\eta$, we compound the $\eta-$indexed formulas to obtain the third form in (\ref{eq:adjustBinomialBeta}) and (\ref{eq:unbiased_Shapley}).

\begin{theorem}\label{thm:bias_Banzhaf} 
The endowment bias $\kappa [v;\mu] \equiv 0$ if and only if  $\mu = \mu[0; \mathrm{BV}]$. 
Besides, given $\mu \in \mathscr{F}$, the aggregate endowment bias
 $\sum\limits_{i\in \mathbb{N}} \kappa_i [v;\mu] \equiv 0$ if and only if $\mu = \mu[0; \mathrm{BV}]$.
\end{theorem}

\noindent \textit{Proof}: See Supplementary Appendix SA-8.

\subsection{\textit{Diminishing Marginality}}
\noindent In general, the candidate variables in $\mathbb{N}$ are more or less, significantly or insignificantly, correlated with each other.
Consequently, for any $i\not \in T$, its explanatory and predictive power is
partially abated by the members of $T$. Moreover, the larger the set $T$, the more likely the abating.
Therefore, the super-additivity assumption is highly artificial, and diminishing marginality is broadly pervasive in our data modeling and forecast situations.

\begin{figure}[!t]
	\centering
	\includegraphics[height=4.5cm, width=7cm]{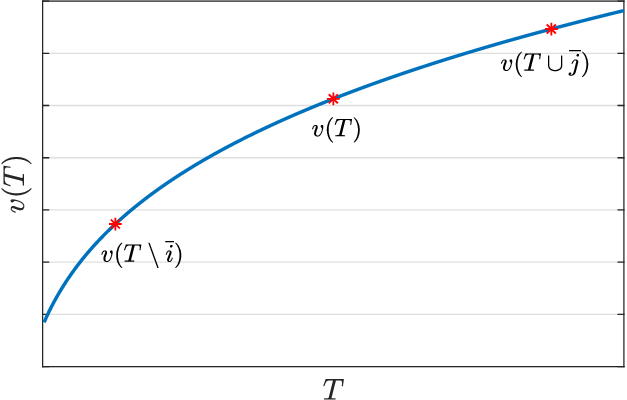}
	\caption{Diminishing Marginality of $v$ where $i\in T$ and $j\not \in T$}\label{fig:diminshing_marginal}
\end{figure}

Let us formally define the diminishing marginality using a model averaging strategy.
Hypothetically, we expect the inequality
$$
v(T) - v(T\setminus \overline{i}) \ge v(T\cup \overline{j}) - v(T)
$$
holds for a typical  $T\subseteq \mathbb{N}$,  $i\in T$ and  $j\not \in T$, as shown in Figure \ref{fig:diminshing_marginal}.
However, it is impractical for the inequality to hold for every $T \subseteq \mathbb{N}$, $i\in T$ and $j\not \in T$.
Thus, we average both sides of the inequality for all $T$ of size $t$, all $i\in T$ and all $j\not \in T$.
The averages on the left side and that on the right side, respectively, are
$$
\left \{
\begin{array}{rcll}
\omega_t(v) 
&\eqdef& 
\frac{ (t-1)! (n-t)!}{n!} \sum\limits_{T\subseteq \mathbb{N}: |T|=t} \
\sum\limits_{i\in T} [v(T) - v(T\setminus \overline{i}) ], &t=1,2,\cdots,n;\\

\pi_t (v) 
&\eqdef& 
\frac{ t! (n-t-1)!}{n!} \sum\limits_{T\subseteq \mathbb{N}: |T|=t} \
\sum\limits_{j\in \mathbb{N} \setminus T} [v(T\cup \overline{j}) - v(T) ], &t=0,1,\cdots,n-1.
\end{array}
\right .
$$

In terms of model averaging, we say $v$ has a \textit{diminishing marginal effect} if  $\omega_t(v) \ge \pi_t(v)$ for $t=1,2,\cdots,n-1$.
We also say $v$ has a \textit{diminishing marginal gain} or \textit{diminishing marginal loss} if 
$\omega_t (v)$ or $\pi_t (v)$, respectively, is a decreasing function of $t$.
Theorem \ref{thm:pi_omega} establishes the equivalence among these types  of 
diminishing marginality.

\begin{theorem}\label{thm:pi_omega}
$\pi_t(v) \equiv \omega_{t+1}(v)$ for $t=0,1,\cdots, n-1$.
Consequently, the following statements are equivalent:
$v$ has a diminishing marginal effect;
$v$ has a diminishing marginal gain;
$v$ has a diminishing marginal loss.
\end{theorem}

\noindent \textit{Proof}: See Supplementary Appendix SA-9.

When used in modeling and forecast, the Shapley value, 
however, tends to demonstrate substantial negative endowment bias, due to the diminishing marginal effect of $v$. 
This counterintuitive issue could bring users undesirable inference from the data and eventually limits the usage of the value concept. 
This negative bias is formally claimed in the first part of Theorem \ref{thm:diminishing}. 
The second part states that super-additivity is also a theoretical explanation for positive endowment effects.
When generating value, $\mathbb{N}$ is often believed to be greater than the sum of its parts.
When modeling data, however, this super-additivity belief contradicts the diminishing marginality of $v$.

\begin{theorem}\label{thm:diminishing}
Assume $\mu = \mu[0; \mathrm{SV}]$. 
If $v$ has a diminishing marginal effect, then the aggregate bias $\sum\limits_{i\in \mathbb{N}} \kappa_i[v;\mu]\le 0$;
if $v$ is super-additive, then $\sum\limits_{i\in \mathbb{N}} \kappa_i[v;\mu]\ge 0$.
\end{theorem}

\noindent \textit{Proof}: See Supplementary Appendix SA-10.

\subsection{\textit{Mitigation of Aggregate Bias}}
\noindent In an objective valuation, such as feature selection, either positive or negative endowment bias should be avoided. 
To systematically eliminate the aggregate bias $\sum\limits_{i=1}^n \kappa_i[v;\mu]$, one could unevenly weight the dichotomous marginal effects and
study a weighted \textit{unbiased D-value} defined as
\begin{equation} \label{eq:unbiasedValue}
\tilde \psi_i[v;\mu] \
\eqdef \ (1-\alpha) \gamma_i[v;\mu] + (1+\alpha) \lambda_i[v;\mu]
\end{equation}
where 
\begin{equation} \label{eq:def_alpha}
\alpha \ \eqdef \ \frac{\sum\limits_{j \in \mathbb{N}} \kappa_j[v;\mu]}{\sum\limits_{j \in \mathbb{N}} \psi_j[v;\mu]}
\end{equation}
is the ratio between the aggregate endowment bias and the aggregate D-value. 
We call $\alpha$ the \textit{endowment bias ratio}.
In the definition (\ref{eq:unbiasedValue}), the weighted marginal gain is $(1-\alpha) \gamma_i[v;\mu] $ while the weighted marginal loss is $ (1+\alpha) \lambda_i[v;\mu]$. 
Clearly, there is no more aggregate endowment bias in this unbiased D-value $\tilde \psi [v;\mu]$:
$$
\sum\limits_{i\in \mathbb{N}} (1-\alpha) \gamma_i[v;\mu] - \sum\limits_{i\in \mathbb{N}} (1+\alpha) \lambda_i[v;\mu] 
=
\sum\limits_{i\in \mathbb{N}} \kappa_i[v;\mu] - \alpha \sum\limits_{i\in \mathbb{N}} \psi_i[v;\mu]
=
0.
$$

In general, the endowment bias ratio $\alpha$ lies between $-1$ and $1$ when $v$ is a monotonic set function. 
If $\alpha$ is close to $0$, then there is no aggregate endowment bias. If $\alpha$ is significantly positive, 
then $\sum\limits_{j\in \mathbb{N}} \gamma_j [v;\mu] > \sum\limits_{j\in \mathbb{N}} \lambda_j [v;\mu]$; we should place more weight on the marginal loss. 
In contrast, if $\alpha$ is significantly negative, then we should place more weight on the marginal gain.
Also, in general, (\ref{eq:unbiasedValue}) does not eliminate the bias for all $i\in \mathbb{N}$, and $\alpha$ in (\ref{eq:def_alpha}) depends on both $v$ and $\mu$. 
For $\mu[\eta; \mathrm{BN}]$, however, (\ref{eq:unbiasedValue}) can entirely remove the endowment bias 
for all $i\in \mathbb{N}$, and the endowment bias ratio $\alpha$ relies only upon $\eta$, as stated in Theorem \ref{thm:binomialbias}.

\begin{theorem}\label{thm:binomialbias}
If $\mu = \mu[\eta; \mathrm{BN}]$, then $\kappa[v;\mu] \equiv (2\eta-1) \psi[v; \mu]$. 
Thus, the endowment bias ratio $\alpha = 2\eta-1$. 
Furthermore, the unbiased D-value is
$$
\begin{array}{rcl}
\tilde \psi_i[v; \mu[\eta; \mathrm{BN}]] 
&=& 4(1-\eta)\gamma_i[v; \mu[\eta; \mathrm{BN}] ] \\
&=& 4\eta\lambda_i[v; \mu[\eta; \mathrm{BN}] ]  \\
&=& 4\eta (1-\eta)\psi_i[v; \mu[\eta; \mathrm{BN}]] \\
&=& 4 \sum\limits_{T\subseteq \mathbb{N}: i\in T} \eta^{|T|} (1-\eta)^{n-|T|+1} \big[ v(T)-v(T\setminus \overline{i}) \big].
\end{array}
$$
\end{theorem}

\noindent \textit{Proof}: See Supplementary Appendix SA-11.

For $\psi_i[v; \mu [\eta; \mathrm{BN}]]$ in Theorem \ref{thm:binomial}, the weights on the marginal gain $\big[ v(T)-v(T\setminus \overline{i}) \big]$ sum to $1$.
As a consequence of Theorem \ref{thm:binomialbias}, the weights in $\tilde \psi_i[v; \mu [\eta; \mathrm{BN}]]$ sum to $4 \eta (1-\eta)$, which has a maximum value $1$ at $\eta=.5$.

\subsection{\textit{An Unbiased Shapley Value}}
\noindent One advantage of the binomial prior $\mu [\eta; \mathrm{BN}]$ is the flexibility of $\eta$.
When independence is assumed, $\eta$ is the probability of any given variable being a true model variable.
If $\mu = \mu [\eta; \mathrm{BN}]$ and $\eta$ itself has a Beta distribution with parameters $(\theta,\rho)$, $\theta>0$ and $\rho>0$, then
\begin{equation} \label{eq:probabilitydensity}
P_{_T} 
=
\mathop{\mathlarger{\int_0^1}}
\frac{\eta^{\theta-1}(1-\eta)^{\rho-1}}{\beta (\theta, \rho)} \eta^{|T|} (1-\eta)^{n-|T|} \mathrm{d} \eta 
=
\frac{\beta(\theta+|T|,\rho+n-|T|)}{\beta(\theta,\rho)}.
\end{equation}
In this integral, $\beta (\cdot,\cdot)$ is the 2-parameter beta function. Thus, model $\mathbf{S}$ 
has an expected size $\mathbb{E} [n \eta] = n \mathbb{E} [\eta]= \frac{n \theta}{\theta+\rho}$. 
Clearly, $\mu \in \mathscr{F}$. We marginalize out $\eta$ in (\ref{eq:binomialPsi}) to get the D-value: 
\begin{equation}\label{eq:binomialBeta}
\begin{array}{rcl}
\psi_i [v;\mu] 
&=& 
\sum\limits_{T\subseteq \mathbb{N}: i \in T}  
\mathop{\mathlarger{\int_0^1}}
\frac{\eta^{\theta-1}(1-\eta)^{\rho-1}}{\beta(\theta,\rho)}  \eta^{|T|-1} (1-\eta)^{n-|T|} 
\left [ v(T) - v(T\setminus \overline{i}) \right ] \mathrm{d}\eta \\

&=& \sum\limits_{T \subseteq \mathbb{N}: i \in T} \frac{\beta(\theta+|T|-1, \rho+n-|T|)}{\beta(\theta,\rho)} \big[ v(T) - v(T\setminus \overline{i}) \big].
\end{array}
\end{equation}
Using the definition of (\ref{eq:D-Value}) can also verify that (\ref{eq:binomialBeta}) is indeed the D-value for the probability density (\ref{eq:probabilitydensity}).
Likewise, by Theorem \ref{thm:binomialbias}, the  unbiased D-value for (\ref{eq:probabilitydensity}) is
\begin{equation}\label{eq:adjustBinomialBeta}
\begin{array}{rcl}
 \tilde \psi_i[v; \mu]
&=& 
4 \sum\limits_{T\subseteq \mathbb{N}: i \in T} 
\mathop{\mathlarger{\int_0^1}}
\frac{\eta^{\theta-1} (1-\eta)^{\rho-1}}{\beta(\theta,\rho)} \eta^{|T|} (1-\eta)^{n-|T|+1} 
\left [ v(T)-v(T\setminus \overline{i}) \right ] \mathrm{d} \eta\\

&=& 
4 \sum\limits_{T\subseteq \mathbb{N}: i \in T} 
\frac{\beta(\theta+|T|,\rho+n-|T|+1)}{\beta(\theta,\rho)} \big[ v(T)-v(T\setminus \overline{i}) \big].
\end{array}
\end{equation}
In fact, (\ref{eq:adjustBinomialBeta}) and (\ref{eq:binomialBeta}) are compounds of  $\tilde \psi_i[v; \mu [\eta; \mathrm{BN}] ]$ and $\psi_i[v; \mu [\eta; \mathrm{BN}] ]$, respectively, according to the beta density function of $\eta$. 
Also, one may choose $v$-specific $\theta$ and $\rho$ such that $\sum\limits_{i\in \mathbb{N}} \tilde \psi_i[v; \mu] = v(\mathbb{N})-v(\emptyset)$, but not ``$\equiv v(\mathbb{N})-v(\emptyset)$."
There are infinitely many choices.

As a special case, when $\theta=\rho=1$ (i.e., $\eta$ has the uniform distribution in the interval $(0,1)$), 
then (\ref{eq:binomialBeta}) reduces to the Shapley value $\Psi_i[v]$. In light of this fact, using (\ref{eq:adjustBinomialBeta}) with $\theta=\rho=1$, we define
the unbiased Shapley value as
\begin{equation}\label{eq:unbiased_Shapley}
\tilde \Psi_i[v] \
\eqdef \ 4 \sum\limits_{T\subseteq \mathbb{N}: i \in T} \frac{(|T|)! (n-|T|+1)!}{ (n+2)!} \big[ v(T)-v(T\setminus \overline{i}) \big].
\end{equation}

The weighting in (\ref{eq:shapley_value}) and (\ref{eq:unbiased_Shapley}) differs in two aspects.
First, for any $i\in \mathbb{N}$,  the weights on its marginal gain $v(T)-v(T\setminus \overline{i})$ sum to $1$ in
both the Shapley value (\ref{eq:shapley_value}) and the Banzhaf value (\ref{eq:banzhaf_value}). 
In the unbiased Shapley value (\ref{eq:unbiased_Shapley}), however, the sum reduces to $\frac{2}{3}$, as stated in the first part of Theorem \ref{thm:UShapleyValueWeights}. 
This one-third de-weighting is due to the inequality $\sum\limits_{i\in\mathbb{N}} \tilde \Psi_i[v]  \not \equiv v(\mathbb{N})-v(\emptyset)$.
Secondly, using $\frac{4(|T|)! (n-|T|+1)!}{ (n+2)!} = \frac{4 |T| (n-|T|+1)}{ (n+1)(n+2)!} \frac{(|T|-1)! (n-|T|)!}{n!}$ to compare the weights in (\ref{eq:shapley_value}) and (\ref{eq:unbiased_Shapley}), we find that large under-weighting occurs when
$|T|$ is far from $\frac{n+1}{2}$. As the expected model size for $\mu [\eta; \mathrm{SV}]$ is $\frac{n}{2}$, such a $T$ is likely either under-fit or over-fit. 
Hence, we should heavily underweight these situations.
In contrast, as indicated in Section \ref{subsect:beyongSVBV}, $\gamma[v;\mu[0,\mathrm{SV}]]$ only heavily underweights small-sized $T$, and $\lambda[v;\mu[0,\mathrm{SV}]]$ only heavily underweights large-sized $T$.

\begin{theorem} \label{thm:UShapleyValueWeights}
For any $i\in \mathbb{N}$, the weights on its marginal gain $[v(T)-v(T\setminus \overline{i})]$ in the unbiased Shapley value $\tilde \Psi_i[v]$, defined in (\ref{eq:unbiased_Shapley}), sum to $\frac{2}{3}$.
Besides,
\begin{equation} \label{eq:total_unbiased_shapley}
\sum\limits_{i \in \mathbb{N}} \tilde \Psi_i[v] 
= 
\sum\limits_{T \subseteq \mathbb{N}: |T| > \frac{n}{2}}  \frac{4(|T|)!(n-|T|)!(2|T|-n)}{(n+2)!} \left [ v(T) - v(\mathbb{N} \setminus T)\right ].
\end{equation}
\end{theorem}

\noindent \textit{Proof}: See Supplementary Appendix SA-12.

The unbiased Shapley value (\ref{eq:unbiased_Shapley}) is an approximate fair-division solution with noise reduction.
As previously stated, the total amount $v(\mathbb{N})-v(\emptyset)$ to be distributed in (\ref{eq:totality}) is inflated when $\mathbf{S}$ is a proper subset of $\mathbb{N}$. 
Similarly, when $T$ is a proper superset of $\mathbf{S}$, $v(T)$ is also over-fit by the irrelevant variables in $T\setminus \mathbf{S}$. 
If we use the irrelevant variables in $\mathbb{N} \setminus T$ to partially offset the over-fit noise in $v(T)$, 
then the net amount $v(T) - v(\mathbb{N} \setminus T)$ would likely be closer to $v(\mathbf{S})$. 
In (\ref{eq:total_unbiased_shapley}), the unbiased Shapley value sums to a weighted average of $v(T) - v(\mathbb{N} \setminus T)$.
When its size is close to $n$, $T$ likely contains all $\mathbf{S}$, particularly when $\frac{|\mathbf{S}|}{n}$ is relatively small. 
Thus, (\ref{eq:total_unbiased_shapley}) places heavy weights on large-sized $T$.
The weights increase with $|T|$ and range from $0$ to $\frac{4}{n}$; the sum of the weights is close to $1$.
If an algorithm seeks only one feature from $\mathbb{N}$, the sought one likely lies in a large-sized $T$. 
For the MVP feature to emerge from the rest ones, we filter out $v(N\setminus T)$ from $v(T)$.
Both Algorithms \ref{alg:UShapley} and \ref{alg:BN} in Section \ref{sect:est_algorthms}  seek one feature at one time.

Finally, among the family of  Beta-Binomial distributions for the model size, 
Theorem \ref{thm:ShapleyValue} provides a few equivalent characterizations of the Shapley value. Not surprisingly, 
it reassures the suitability of the marginal loss component $\lambda[v;\mu]$ in distributing the expected model performance $\mathbb{E} v(\mathbf{S}) - v(\emptyset)$.

\begin{theorem} \label{thm:ShapleyValue}
Assume $\mu = \mu[\eta; \mathrm{BN}]$ where $\eta$ follows a Beta distribution with
parameters $(\theta,\rho)$. Then
the following are equivalent:
(i)  $\psi [v;\mu] \equiv \Psi [v]$;
(ii)  $\sum\limits_{i\in \mathbb{N}}  \psi_i[v; \mu] \equiv v(\mathbb{N}) - v(\emptyset)$;
(iii) $\sum\limits_{i\in \mathbb{N}}  \lambda_i [v;\mu] \equiv \mathbb{E} v(\mathbf{S}) - v(\emptyset)$;
(iv) $\sum\limits_{i\in \mathbb{N}}  \gamma_i [v;\mu] \equiv v(\mathbb{N}) - \mathbb{E} v(\mathbf{S})$;
(v)  $\theta=\rho=1$.
\end{theorem}

\noindent \textit{Proof}: See Supplementary Appendix SA-13.

\section{Estimation}\label{sect:est_algorthms}
\noindent 
In this section, we modify the sequential approach in Shapley (1953) to estimate the D-value, its components, and its unbiased version.
For a large $n$, exact calculation of the D-value $\psi[v;\mu]$ is not practical; 
thus, we seek random sampling techniques to approximate it.
An easy way to randomly approximate the D-value $\psi[v; \mu]$ 
is to randomly draw a large sample of $\mathbf{S}$ and then average the marginal gain and marginal loss
in the sample. For a large $n$, however, some members in $\mathbb{N}$ 
could be much less represented in the sample than other members. 
Moreover,
the distribution of $\mathbf{S}$ may not be unique and depend on the unknown parameter $\eta$, impairing the random sampling of $\mathbf{S}$.
This section studies a random ordering in which each member appears precisely once in the ordering. 
We then average the weighted conditional performance for all members in a large number of random orderings.

\subsection{\textit{Sequential Implementation}}
\noindent Let $\Omega$ be the set of orderings of all candidate variables in $\mathbb{N}$.
There are $n!$ orderings in total. We randomly take an ordering $\tau$ from $\Omega$:
$$
\tau: \quad \emptyset\ \longrightarrow \  i_1\ \longrightarrow \ \cdots\  \longrightarrow \ i\ \longrightarrow \ \cdots\ \longrightarrow \ i_n.
$$
Let 
$\Xi_i^\tau$ be the set of variables in $\mathbb{N}$ which precede $i$ in the ordering $\tau$, and let
$\phi^\tau_i$ be $i$'s entry marginal contribution in $\tau$, i.e.,
$$
\phi^\tau_i \ \eqdef \ v(\Xi_i^\tau \cup \overline{i}) - v(\Xi_i^\tau).
$$
$\phi^\tau_i$ is $i$'s performance in $\tau$ when $\Xi_i^\tau$ are the first-movers ahead of $i$.
This conditional performance $\phi^\tau_i$ ignores the behavior of $\mathbb{N} \setminus \Xi_i^\tau \setminus \overline{i}$;
it also ignores the internal ordering in $\Xi_i^\tau$.
Shapley (1953) showed that $\Psi_i [v] = \mathbb{E} [\phi^\tau_i] $ 
where the expectation is under the uniform distribution on $\Omega$.

To estimate $\gamma[v;\mu]$, $\lambda[v;\mu]$, and $\psi[v;\mu]$ for any $\mu$, we  
bind the probability density $\mu$ to the conditional performance $\phi^\tau_i$ by letting
\begin{equation}\label{eq:variant_tau_phi}
\tilde \phi^\tau_i \
\eqdef \
\frac{(P_{_{\Xi_i^\tau}}+P_{_{\Xi_i^\tau \cup \overline{i}}}) n!}{(|\Xi_i^\tau|)!(n-|\Xi_i^\tau| - 1)!}\phi^\tau_i.
\end{equation}
Thus, $\tilde \phi^\tau$ is a weighted or scaled $\phi^\tau$, and the weight is not a constant.
When $\mu \in \mathscr{F}$, $\tilde \phi^\tau_i$ reduces to
$$
\tilde \phi^\tau_i = \left [ (n-|\Xi_i^\tau|)\delta_{_{|\Xi_i^\tau|}} + (|\Xi_i^\tau|+1) \delta_{_{|\Xi_i^\tau|+1}} \right ] \phi^\tau_i. 
$$

\begin{theorem}\label{thm:sq}
Assume $\tau$ has the uniform distribution on $\Omega$. Then
$\mathbb{E} [\tilde \phi^\tau_i] = \psi_i[v;\mu]$, and
\begin{equation}\label{eq:sequel_phi}
\left \{
\begin{array}{rcl}
\gamma_i[v;\mu]
&=& 
\mathbb{E} \frac{n! P_ {_{\Xi_i^\tau \cup \overline{i}}}}{(|\Xi_i^\tau|)!(n-|\Xi_i^\tau| - 1)!} \phi^\tau_i, \\ 

\lambda_i[v;\mu]
&=&
\mathbb{E} \frac{n! P_{_{\Xi_i^\tau}}}{(|\Xi_i^\tau|)!(n-|\Xi_i^\tau| - 1)!} \phi^\tau_i.
\end{array}
\right .
\end{equation}
When $\mu \in \mathscr{F}$, the two components are
\begin{equation}\label{eq:sequel_phi_noprior}
\left \{
\begin{array}{rcl}
\gamma_i[v;\mu]
&=&
\mathbb{E} (|\Xi_i^\tau|+1) \delta_{_{|\Xi_i^\tau|+1}} \phi^\tau_i, \\ 

\lambda_i[v;\mu]
&=&
\mathbb{E} (n-|\Xi_i^\tau|)\delta_{_{|\Xi_i^\tau|}} \phi^\tau_i.
\end{array}
\right .
\end{equation}
In particular,  the Shapley value consists of the following two parts
\begin{equation}\label{eq:ShapleyGainLoss}
\left \{
\begin{array}{rcl}
\gamma_i[v;\mu[0,\mathrm{SV}]]&=& \mathbb{E} \frac{|\Xi_i^\tau|+1}{n+1} \phi^\tau_i , \\
\lambda_i[v;\mu[0,\mathrm{SV}]]&=& \mathbb{E}\frac{n-|\Xi_i^\tau|}{n+1} \phi^\tau_i .
\end{array}
\right .
\end{equation}
Also, the unbiased Shapley value equals
\begin{equation}\label{eq:unbiasedShapleyValue}
\tilde \Psi_i[v]= 4 \mathbb{E} \frac{(|\Xi_i^\tau|+1)(n-|\Xi_i^\tau|)}{(n+1)(n+2)} \phi^\tau_i .
\end{equation}
\end{theorem}

\noindent \textit{Proof}: See Supplementary Appendix SA-14.

Theorem \ref{thm:sq} provides a generic calculation tool for the formulas defined in the last two sections.
To estimate the D-value, for example, we take a large sample of random orderings from $\Omega$, say $\tau_1,\tau_2, \cdots, \tau_s$. 
Then, we use the average of $\tilde \phi^{\tau_j}$ to estimate $\psi[v;\mu]$.
The average converges as the sample size $s$ increases, according to the large sample theory.
At the same time, we could also estimate the variance-covariance of  $\left (\tilde \phi^\tau_1, \tilde \phi^\tau_2,\cdots, \tilde \phi^\tau_n \right )'$.
This information, together with the convergence rate in the multidimensional Central Limit Theorem, helps us determine a sample size $s$ such
that the sampling error in the average is, at a high confidence level, within an acceptable tolerance.
Additionally, we can extract the sample median, confidence interval, and other robust statistics 
of $\psi[v;\mu]$ from $\tilde \phi^{\tau_j}$, $j=1,2, \cdots, s$.
Similar computational procedures apply to $\tilde \Psi[v]$, $\gamma[v;\mu]$, and $\lambda[v;\mu]$, respectively.

We can also determine the sample size $s$ once the change of the average series is within a small tolerance. For example, consider calculating $\psi[v;\mu]$ with a given tolerance level $\xi$.
Let the $s$-th average be
$$
\psi^{(s)}[v;\mu] \
\eqdef \
\frac{1}{s} \ \sum\limits_{j=1}^s \tilde \phi^{\tau_j}.
$$
Then, we have the recursive relation
$$
\psi^{(s)}[v;\mu]
=
\frac{s-1}{s} \psi^{(s-1)} [v;\mu] \
+ \
\frac{1}{s} \tilde \phi^{\tau_s}.
$$
We stop any further randomly drawing $\tau$ if
$$
\left | \left | \psi^{(s)}[v;\mu] \
 - \
 \psi^{(s-1)}[v;\mu] \right | \right |_\infty \le \xi
$$
where $||\cdot||_\infty$ is the $L_\infty$ norm. Equivalently,
$$
\frac{1}{s} \ \left | \left | \tilde\phi^{\tau_s}  \ - \ \psi^{(s-1)}[v;\mu] \right | \right |_\infty \ \le \  \xi.
$$
Finally, the estimated value of  $\psi[v;\mu]$ is $\psi^{(s)}[v;\mu]$.

\subsection{\textit{Cut-off by Statistical Significance}}\label{subsect:cutoff_sig}

\noindent 
In the above sections, we have laid down the foundation of dichotomous valuation and its estimation. 
In the context of variable selection, dropping or selecting a candidate variable is two sides of the same coin, justifying a concrete exercise of dichotomous valuation.
To apply the dichotomous valuation to variable selection, we need two ingredients: an appropriate set function $v$ or entry performance $\phi^\tau_i$ in the random ordering $\tau$;
a desirable simple estimation method of $\tilde \psi[v;\mu]$.
To make the binary decision of $i$ being relevant or irrelevant, we set a rule which separates the two exclusive decisions by a critical value or cut-off value.
Thus, we only need to know if the estimated $\tilde \psi_i[v;\mu]$ is larger or less than the cut-off value, without precisely calculating it.

We drop or select variable $i$ based on its expectation of weighted entry statistical significance in $\tau$.
Using statistical significance in selection decisions is a common practice in statistics and econometrics.
When $i$ is irrelevant to $v$, we expect its contribution in $\tau$ is of less significance, especially when most members of $\mathbf{S}$ are already in $\Xi_i^\tau$, or $i$ correlates with variables in $\Xi_i^\tau$.
It might still be significant, due to the sampling error,  if it is one of the few first entrants in $\tau$. 
But, $i$'s significance is too complicated to be explicitly expressed (e.g., Freedman, 1983). Thus, we apply the model averaging mechanism.
For simplicity, we use $i$'s absolute value of t-statistic as its entry significance or conditional performance in $\tau$. 
Then, the cut-off is the one-sided critical value of a $t$-distribution, say $t_{_{dof,\xi}}$.
The degree of freedom $dof$ ranges from the sample size of observations to the sample size minus $n$; the significance level $\xi$ could be $.01$, $.05$, or $.1$.
The $t$-statistic is widely used for significance tests; it is easy to compute and is available in most econometric and statistical software packages.
Also, for the same reason, using the absolute t-statistic as the conditional performance in $\tau$, the variable with the maximum unbiased D-value is likely the most statistically significant.
Alternatively, one could use the incremental F-statistic or likelihood ratio in the nested models in $\tau$, which involves higher computational cost.

With the entry significance acting as the conditional performance $\phi^{\tau}_i$ in $\tau$, the computation could be much simpler than that implied in Theorem \ref{thm:sq}. 
On the one hand, $i$'s actual performance is that in $\mathbf{S}$:
$i$'s conditional performance is diluted when $\Xi_i^\tau$ is a proper superset of $\mathbf{S}\setminus \overline{i}$; it is inflated when $\Xi_i^\tau$ is a proper subset of $\mathbf{S}\setminus \overline{i}$.
Thus, if some $j\in \mathbb{N}$ is already identified as a relevant variable, we should place it before $i$ in the random ordering $\tau$;
if $j\in \mathbb{N}$ is previously identified as irrelevant, we should place it after $i$ in the random ordering.
Hence, we could first target the most valuable player (MVP) in $\mathbb{N}$.
This MVP has the highest expected performance in all modeling scenarios.
After this variable is selected, we then seek the next most significant one conditional on the first selection. 
This new one is the most valuable partner (MVP) of the first selected MVP.
We continue this fashion until no more MVP can make a significant contribution.
On the other hand, when identifying the MVP variable in the remaining player set, say $\mathbb{R}$, we only need to know, at high confidence, if the largest value in $\tilde \psi[v_{_\mathbb{R}};\mu]$ is greater than or less than the cut-off value $t_{_{dof,\xi}}$. 
Thus, high precision is generally not required for $\tilde \psi[v_{_\mathbb{R}};\mu]$, and we could use much fewer random orderings than that implied in Theorem \ref{thm:sq}.
Finally, Algorithm \ref{alg:UShapley} realizes the above ideas by iteratively applying $\tilde \Psi[v_{_\mathbb{R}}]$ to find the
MVPs until no more MVP is significantly valuable. 
Similar MVP algorithms work for $\Psi[v]$,  $\gamma[v;\mu[0;\mathrm{SV}]]$, and $\lambda[v;\mu[0;\mathrm{SV}]]$, respectively.

\begin{figure}[htb]
\centering
\begin{minipage}{.9\linewidth}
\begin{algorithm}[H]\label{alg:UShapley}
\SetAlgoLined
$\mathbb{R} \longleftarrow \mathbb{N}$, \ $\hat{\mathbf{S}} \longleftarrow \emptyset$\;
\While{$\mathbb{R} \not = \emptyset$}{
	Apply Theorem \ref{thm:sq} to estimate $\tilde \Psi[v_{_\mathbb{R}}]$, while keeping the variables in $\hat{\mathbf{S}}$ as additional regressors in the regression\;
	$j \longleftarrow \argmax_{i \in \mathbb{R}} \tilde \Psi_i [v_{_\mathbb{R}}]$\;
	\eIf{ $\tilde \Psi_j[v_{_\mathbb{R}}] > t_{_{dof,\xi}}$}{
		$\hat{\mathbf{S}} \longleftarrow \hat{\mathbf{S}} \cup \overline{j}$,\ 
		$\mathbb{R} \longleftarrow \mathbb{R} \setminus \overline{j}$\;
	}{
		Return $\hat{\mathbf{S}}$\;
	}
}
\caption{MVP Implementation of Variable Selection by $\tilde \Psi[v]$}
\end{algorithm}
\end{minipage}
\end{figure}

\vskip .5cm
The prior belief about $\mathbf{S}$ is dynamically updated in each iteration of Algorithm \ref{alg:UShapley}. 
At the inception, each variable is equally likely to be a relevant one. 
After the first MVP variable, say $x_{(1)}$, is included in $\hat{\mathbf{S}}$, all other variables in $\mathbb{N}\setminus \{ x_{(1)}\}$ are equally likely to be selected.
After the second MVP variable, say $x_{(2)}$, is included in $\hat{\mathbf{S}}$, all other variables in $\mathbb{N}\setminus \{ x_{(1)}, x_{(2)}\}$ are equally likely to be selected. 
And so on.

The sequential approach is different from the classical stepwise regression procedure. 
In the sequential approach, we apply weighted averaging to the conditional performance $\phi^\tau_i$.
The stepwise procedure admits and drops variables based on their single significance test; 
however, the exact significance level cannot be calculated (e.g., Freedman, 1983). 
In fact, because of the diminishing marginality, variable $i$'s  significance
tends to become smaller as the size of $\Xi_i^\tau$ increases;
consequently, starting from different models could lead to different selected models.
In particular, the sequential approach for the unbiased D-value is designed to soften the first-mover advantage.
Another drawback of the stepwise procedure is that it heavily relies on a single criterion, such as the $F$-statistic or $t$-statistic, in a specific modeling scenario. 
In contrast,  in the sequential approach, 
the weighted conditional performance $\phi^\tau_i$ converges as the number of random orderings increases. Even though $i\in \mathbb{N}$ is subject to selection bias and omitted-variable bias in a single random ordering,
these biases partially offset each other in the weighted model averaging process.

\subsection{\textit{$\psi[v; \mu[\eta, \mathrm{BN}]]$ and $\tilde \psi[v; \mu[\eta, \mathrm{BN}]]$ with an Unknown $\eta$}} \label{subsect:estimate_BN_D-Value}

\noindent 
In essence, Algorithm \ref{alg:UShapley} adopts up to one MVP variable at a time, focusing on one feature from the remaining ones in $\mathbb{R}$. 
When $\mu = \mu [\eta; \mathrm{BN}]$ with an unknown $\eta \in (0,1)$, the estimated model $\hat{\mathbf{S}}$  indeed depends on the choice of $\eta$. 
It would be challenging to estimate both $\eta$ and $\mathbf{S}$ at the same time.
To avoid estimating $\eta$, we adapt Algorithm \ref{alg:UShapley} such that it targets only the MVP variable from $\mathbb{R}$ when calculating $\tilde \psi[v_{_\mathbb{R}}; \mu[\eta, \mathrm{BN}]]$. 
The modified algorithm is:

\begin{figure}[htb]
\centering
\begin{minipage}{.9\linewidth}
\begin{algorithm}[H]\label{alg:BN}
\SetAlgoLined
$\mathbb{R} \longleftarrow \mathbb{N}$, \ $\hat{\mathbf{S}} \longleftarrow \emptyset$\;
\While{$\mathbb{R} \not = \emptyset$}{
	Apply Theorem \ref{thm:sq} to estimate $\tilde \psi \left [v_{_\mathbb{R}};\mu \left [\frac{1}{|\mathbb{R}|};\mathrm{BN} \right ] \right ]$, 
	while keeping the variables in $\hat{\mathbf{S}}$ as additional regressors in the regression\;
	$j \longleftarrow \argmax_{i \in \mathbb{R}} \tilde \psi_i \left [v_{_\mathbb{R}};\mu \left [\frac{1}{|\mathbb{R}|};\mathrm{BN} \right ] \right ]$\;
	\eIf{$\tilde \psi_j \left [v_{_\mathbb{R}};\mu \left [\frac{1}{|\mathbb{R}|};\mathrm{BN} \right ] \right ]> t_{_{dof,\xi}}$}{
		$\hat{\mathbf{S}} \longleftarrow \hat{\mathbf{S}} \cup \overline{j}$,\
		$\mathbb{R} \longleftarrow \mathbb{R} \setminus \overline{j}$ \;
	}{
		Return $\hat{\mathbf{S}}$\;
	}
}
\caption{MVP Implementation of Variable Selection by $\tilde \psi[v;\mu[\eta,\mathrm{BN}]]$}
\end{algorithm}
\end{minipage}
\end{figure}

\vskip .3cm
In Algorithm \ref{alg:BN}, we use (\ref{eq:sequel_phi_noprior}) to estimate either $\gamma \left [v_{_\mathbb{R}};\mu \left [\frac{1}{|\mathbb{R}|};\mathrm{BN} \right ] \right ]$ 
or $\lambda \left [v_{_\mathbb{R}};\mu \left [\frac{1}{|\mathbb{R}|};\mathrm{BN} \right ] \right ]$.
Then, $\tilde \psi \left [v_{_\mathbb{R}};\mu \left [\frac{1}{|\mathbb{R}|};\mathrm{BN} \right ] \right ]$ follows from Theorem \ref{thm:binomialbias}.
Initially, we set $\eta = \frac{1}{n}$. After $k$ variables enter into $\hat{\mathbf{S}}$, $\eta$ becomes $\frac{1}{n-k}$.
A similar algorithm works for $\psi[v; \mu[\eta, \mathrm{BN}]]$. 
The $dof$ in both MVP algorithms is the sample size minus the number of explanatory variables in the regeression, including the constant intercept.

It is worth mentioning that $\tilde \psi_j[v_{_\mathbb{R}};\mu]$ in both Algorithms \ref{alg:UShapley} and \ref{alg:BN} is conditional on the previously selected variables in $\mathbb{N}\setminus \mathbb{R}$. 
Computing the exact unconditional $\tilde \psi[v;\mu]$ for a selected variable remains a challenge in these algorithms;
a good approximation is $\tilde \psi [v_{\hat{\mathbf{S}}}; \mu ]$.
Nevertheless, as mentioned previously, we do not have to calculate the conditional $\tilde \psi_j[v_{_\mathbb{R}};\mu]$ precisely. 
By the large sample theory, thirty or more random orderings are highly recommended. 
Also, when the estimated $\tilde{\psi}_j[v_{_\mathbb{R}};\mu]$ and the cut-off value $t_{_{dof,\xi}}$ are too close, we should increase the number of random orderings.
When the largest two values in $\tilde \psi[v_{_\mathbb{R}};\mu]$ are too close, we may also add more random orderings.
In general, the computational cost of Algorithms \ref{alg:UShapley} and \ref{alg:BN} is of order $O(n|\mathbf{S}|)$.

\section{Simulation Studies} \label{sect:simulation}
\noindent
In this section, we conduct seven simulation experiments to study the performance of the selection methods proposed in the previous sections. 
We first compare the performance of four new methods with six popular ones used in practice.
In terms of exact identification of the actual models, the new approaches are twice as accurate as of others on average.
In contrast to their biased opponents, the two unbiased D-value methods, $\tilde \Psi[v]$ and $\tilde\psi[v;\mu[\eta;\mathrm{BN}]]$, reduce a 90\% chance of over-fitting.
Then, we analyze the selection accuracy upon occasions of homogeneous correlation between the covariates, homogeneous correlation between the residuals, large sample sizes of the observations,
and varying sizes of $\mathbf{S}$.

\subsection{\textit{Comparison with Other Selection Methods}}\label{subsect:comparison}

\noindent In this subsection, we compare the performance of twelve variable selection methods in 1,000 simulated linear models.
For convenience, we group these methods into three categories:
the four new ones (NEW4): $\tilde \Psi[v]$, $\gamma[v;\mu[0;\mathrm{SV}]]$, $\lambda[v;\mu[0;\mathrm{SV}]]$, and $\tilde \psi[v;\mu[\eta;\mathrm{BN}]]$;
the six popular ones used in practice (POP6):  Lasso (Tibshirani, 1996), adaptive Lasso or aLasso (Zou, 2006), stepwise regression, and three information-criterion approaches (AIC, BIC, and Hannan-Quinn);
the two biased ones (BIS2): the Shapley value $\Psi[v]$, and the D-value $\psi[v;\mu[\eta;\mathrm{BN}]]$ with a binomial model size.

\subsubsection{\textit{The Data and Models}}
\noindent 
In this experiment, each model has a dependent variable $Y_j$ and $20$ candidate explanatory variables $\left \{ X_{ji} \right \}_{i=1}^{20}$. 
The real relationship is
\begin{equation} \label{eq:simulate_data}
Y_j = \zeta_0 + \sum\limits_{i=1}^5 \zeta_i X_{ji} + \sum\limits_{i=6}^{20} 0 \times X_{ji} + \epsilon_j
\end{equation}
for some model-specific unknown coefficients $\zeta_i$ and white noise $\epsilon_j$. Thus, the actual model variables are $\mathbf{S} = \{ X_{j1}, \cdots, X_{j5} \}$.
Each model has a dataset of 100 simulated observations. 
In generating a dataset, we first simulate 20 independent variables using the normal random number generator. 
To remove the normality, we transform the raw data by a nonlinear function, such as exponential, square, cubic, or logarithmic of absolute value. 
We then simulate a set of $6$ coefficients $\zeta_i$, including the constant intercept $\zeta_0$. 
Finally, we apply (\ref{eq:simulate_data}) to calculate the dependent variable $Y_j$.

A selection method attempts to drop all the irrelevant variables in $\mathbb{N}\setminus \mathbf{S}$ and keep only the relevant ones in $\mathbf{S}$.
Hopefully, the estimated model $\hat{\mathbf{S}}$ should exactly match the actual model $\mathbf{S}$, i.e., $\hat{\mathbf{S}} = \mathbf{S}$. 
In reality, however, the method could make two types of error:
if  $\hat{\mathbf{S}}\setminus \mathbf{S} \not = \emptyset$, then some irrelevant variables are falsely selected; 
if $\mathbf{S}\setminus \hat{\mathbf{S}} \not = \emptyset$, then some true model variables fail to be selected. 
In a large number of simulated models, an excellent selection method
should correctly identify a large percentage of the right models and should make a small percentage of errors. 
For each simulated dataset, we apply any of the twelve variable selection methods to find an estimated model $\hat{\mathbf{S}}$.
After that, we compute the discrepancy between  $\hat{\mathbf{S}}$ and  $\mathbf{S}$. 
The discrepancy includes the number of not-selected true model variables $\mathbf{S} \setminus \hat{\mathbf{S}}$,
and the number of selected irrelevant variables $\hat{\mathbf{S}} \setminus \mathbf{S}$.
Finally, we aggregate the discrepancy statistics for all 1,000 datasets.

In this and other experiments, we use the following options: 100 random orderings in both NEW4 and BIS2;
$.05$ or $.01$ significance level for both NEW4 and BIS2;
for the AIC, BIC, or H-Q (Hannan-Quinn) approach, search all $2^n$ subsets of $\mathbb{N}$ and choose the subset with the least information criterion as the estimated model $\hat{\mathbf{S}}$;
for the Lasso or adaptive Lasso method, apply tenfold cross-validation. 
We use multiple implementations of Lasso and aLasso by the software packages MatLab and R; the best results are reported in Tables \ref{tb:compareselectionmethods}$-$\ref{tb:size_of_S}.
There are many stepwise regression methods in the literature; we adopt the best results produced by the software EViews.

\subsubsection{\textit{The Comparison}}
\noindent 
We first compare NEW4 and POP6. Rows 2 through 7 in Table \ref{tb:compareselectionmethods} summarize the discrepancy statistics for all these twelve methods in the 1,000 simulated models. 
The second row is the numbers of correctly identified models --- the NEW4 has an average accuracy of 90.3\%, compared to 44.4\% for the POP6.  
Rows 3 through 4 are the numbers of estimated models that reject relevant variables. These numbers measure the under-fitting of the selection methods. 
All the NEW4 are barely better than the POP6 at resolving the under-fitting issue. 
On average, they turn down relevant variables with a 7.8\% chance, while the POP6 has a chance of 8.2\%.
In rows 5 through 7, we list the numbers of estimated models, which include irrelevant variables. 
These numbers measure the over-fitting of the selection methods. 
The NEW4 successfully blocks the irrelevant variables from being selected. 
The average chance to make this type of error is only 2.1\%, a 95.7\% reduction from the 48.3\% over-fitting rate for the POP6.

Next, the advantage of unbiasedness is evidenced by the contrast between the BIS2 and their unbiased counterparts. 
The unbiased methods, $\tilde \Psi[v]$ and $\tilde \psi[v;\mu[\eta; \mathrm{BN}]]$, outperform the BIS2 by 28.8\% in terms of exact identification.
The unbiasedness also reduces 91.5\% over-fit models.

\begin{sidewaystable}
\caption{Comparison of Twelve Variable Selection Methods in 1,000 Simulated Models$^*$}
\label{tb:compareselectionmethods}
\centering
{ \small
\begin{tabular}{cc||c||c|c|c|c|c|c|c|c|c|c|c|c} \hline \hline
&&&Lasso&aLasso&Stepwise&AIC&BIC&H-Q&$\psi[\mathrm{BN}]$&$\tilde\psi[\mathrm{BN}]$&$\Psi[v]$&$\tilde\Psi[v]$&$\gamma[\mathrm{SV}]$&$\lambda[\mathrm{SV}]$\\ \hline \hline
\parbox[t]{2mm}{\multirow{6}{*}{\rotatebox[origin=c]{90}{$|\mathbf{S}|=5$}}} & \parbox[t]{2mm}{\multirow{6}{*}{\rotatebox[origin=c]{90}{independent}}}
& $\hat{\mathbf{S}}=\mathbf{S}$			         &691&698&477&42 &539&216&654&876&737&915&908&909\\  
&&$|\mathbf{S}\setminus\hat{\mathbf{S}}|=1$	     &85 &233&42 &18 &40 &25 &49 &81 &49 &79 &73 &75 \\
&&$|\mathbf{S}\setminus\hat{\mathbf{S}}|\ge 2$	 &2  & 45&0  &0  &0  &0  &0  &4  &0  &0  &0  &0  \\
&&$|\hat{\mathbf{S}}\setminus \mathbf{S}|=1$     &162&13 &330&165&302&339&231&36 &182&7  &19 &18  \\
&&$|\hat{\mathbf{S}}\setminus \mathbf{S}|=2$     &40 &8  &133&237&106&223&64 &3  &38 &0  &1  &0  \\
&&$|\hat{\mathbf{S}}\setminus \mathbf{S}|\ge 3$	 &29 &5  &38 &555&31 &214&17 &0  &6  &0  &0  &0  \\ \hline
\parbox[t]{2mm}{\multirow{6}{*}{\rotatebox[origin=c]{90}{$|\mathbf{S}|=5$}}} & \parbox[t]{2mm}{\multirow{6}{*}{\rotatebox[origin=c]{90}{correlated}}}
& $\hat{\mathbf{S}}=\mathbf{S}$              	 &96 &76 &477&60 &521&229&657&938&765&982&970&970  \\ 
&&$|\mathbf{S}\setminus\hat{\mathbf{S}}|=1$		 &6  &6  &6  &2  &4  &3  &6  &15 &6  &14 &13 &11  \\
&&$|\mathbf{S}\setminus\hat{\mathbf{S}}|\ge 2$   &0  &0  &0  &0  &0  &0  &0  &0  &0  &0	 &0  &0  \\ 
&&$|\hat{\mathbf{S}}\setminus \mathbf{S}|=1$     &48 & 24&332&144&338&307&235&46 &202&4  &16  &18  \\
&&$|\hat{\mathbf{S}}\setminus \mathbf{S}|=2$     &36 & 28&133&213&109&245&87 &2  &22 &0  &1  &1  \\
&&$|\hat{\mathbf{S}}\setminus \mathbf{S}|\ge 3$  &816&868&56 &583&30 &218&16 &0  &6  &0  &0  &0  \\  \hline
\parbox[t]{2mm}{\multirow{6}{*}{\rotatebox[origin=c]{90}{$|\mathbf{S}|=12$}}}&\parbox[t]{2mm}{\multirow{6}{*}{\rotatebox[origin=c]{90}{correlated}}}
& $\hat{\mathbf{S}}=\mathbf{S}$				     &12 &  7&660&181&659&382&814&912&852&964&962&958  \\
&&$|\mathbf{S}\setminus\hat{\mathbf{S}}|=1$      &24 & 27&15 &14 &15 &15 &17 &30 &19 &34 &30 &29 \\
&&$|\mathbf{S}\setminus\hat{\mathbf{S}}|\ge 2$   &0  & 12&1  &0  &1  &0  &1  &11 &1  &1  &1  &1\\
&&$|\hat{\mathbf{S}}\setminus \mathbf{S}|=1$	 &66 & 43&251&319&265&373&150&43 &114&1  &8  &14 \\
&&$|\hat{\mathbf{S}}\setminus \mathbf{S}|=2$	 &130& 71&62 &275&55 &178&21 &6  &17 &0  &0  &0  \\
&&$|\hat{\mathbf{S}}\setminus \mathbf{S}|\ge 3$  &788&878&16 &223&10 &65 &0  &0  &0  &0	 &0  &0  \\ \hline\hline	
\multicolumn{15}{l}{\small{*: \ $n=20$; the actual model size $|\mathbf{S}|$ is either $5$ or $12$; the covariates are independent or heterogeneously correlated.}}  \\
\multicolumn{15}{l}{\small{*: \ $\gamma[\mathrm{SV}]$ for $\gamma[v;\mu[0;\mathrm{SV}]]$, \ $\lambda[\mathrm{SV}]$ for $\lambda[v;\mu[0;\mathrm{SV}]]$, $\tilde\psi[\mathrm{BN}]$ for $\tilde\psi[v;\mu[\eta;\mathrm{BN}]]$, and $\psi[\mathrm{BN}]$ for $\psi[v;\mu[\eta;\mathrm{BN}]]$.}}  \\
\multicolumn{15}{l}{\small{*: \ We denote ``$\hat{\mathbf{S}}=\mathbf{S}$" the exact identification of the true model, ``$|\mathbf{S}\setminus\hat{\mathbf{S}}|$" the number of relevant variables }} \\
\multicolumn{15}{l}{\small{\hspace{.5cm} which are not selected, ``$|\hat{\mathbf{S}}\setminus \mathbf{S}|$" the number of irrelevant variables which are selected.}}  \\			
\end{tabular} 
}
\end{sidewaystable}

\subsubsection{\textit{Robustness Checks}}
\noindent 
In this section so far, the data and the actual models are generated by Monte Carlo simulation so that the real models are known by design. 
For simplicity, we assume orthogonalization among the candidate variables; but, cross-products, endogeneity, and lags are not allowed.
In this subsection, we conduct two robustness checks. 
Both checks reconfirm the effectiveness of our new methodologies.

In the first robustness check, we apply a linear transformation to the 20 independent variables after being created by the random number generator and before being applied by the nonlinear transformations. 
By applying the linear transformation, we make the candidate variables correlated. 
For each dataset, the transform matrix is also randomly generated. 
We tabulate the discrepancy statistics in rows 8 through 13 in Table \ref{tb:compareselectionmethods}.
Not surprisingly,  all the NEW4 perform even better in the situations with correlated candidate regressors and with even stronger diminishing marginality.

In the second robustness check, we increase the size of the true model variables to twelve, i.e.,
the real model changes to
$$
Y_j = \zeta_0 + \sum\limits_{i=1}^{12} \zeta_i X_{ji} + \sum\limits_{i=13}^{20} 0 \times X_{ji} + \epsilon_j.
$$
We still allow the random correlation among the candidate regressors.
Rows 14 through 19 in Table \ref{tb:compareselectionmethods} list the discrepancy statistics. 
The results are consistent with those in rows 2 through 13.

Looking at rows 8 through 19 together, we find that the NEW4's accuracy of exact identification is 95.7\% on average, compared to POP6's 28.0\% and BIS2's 77.2\%.
Also, the average rate of over-fitting is only 2.0\% for the NEW4, compared to 71.6\% for the POP6 and 21.6\% for the BIS2.
Based on Table \ref{tb:compareselectionmethods}, the BIS2 surpasses the POP6 in extracting the real features and blocking the false ones, primarily due to the dichotomous valuation of each feature and the MVP algorithms of successive acceptance.
Within the POP6, the BIC has the best accuracy on average, and the stepwise regression comes next. 
Lasso and adaptive Lasso are among the best choices when the covariates are independent.
The independence assumption (Tibshirani, 1996, p.268; Zou, 2006, p.1418) seems a necessary condition to maintain the performance, according to the studies of robustness check.

\subsection{\textit{Accuracy of Exact Identification}}
\noindent
In this subsection, we use the exact identification rate as the accuracy measurement for the selection methods NEW4
and analyze the accuracy's dependency on the sample size of observations, the correlation among the regressors, the correlation among the residuals, and the size of relevant features.
In all of these experiments, $\gamma [v;\mu[0; \mathrm{SV}]]$, $\lambda [v;\mu[0; \mathrm{SV}]]$, and $\tilde \Psi [v]$ behave similarly. 
$\tilde \psi [v;\mu[\eta; \mathrm{BN}]]$ has a lower precision when the covariates or the error terms are homogeneously correlated, but it has the highest accuracy, along with the BIC method, when the sample size is large.

\subsubsection{\textit{Homogeneous Correlation Between the Covariates}}
\noindent
In rows 8 through 19 in Table \ref{tb:compareselectionmethods}, the covariates are randomly correlated. Thus, the same high correlation among all covariates is not likely to occur.
In this experiment, we investigate how the accuracy of NEW4 changes with the degree of correlation between the covariates.
We simulate 100,000 datasets using (\ref{eq:simulate_data}) such that the expected correlation between any two different covariates is constant in each dataset. 
Let the constant be $\frac{1}{100,001}, \frac{2}{100,001},\cdots,$ and $\frac{100,000}{100,001}$, respectively. For each dataset, we apply the NEW4 to identify the exact models. 

\begin{figure}[!t]
	\centering
	\includegraphics[height=5cm, width=9cm]{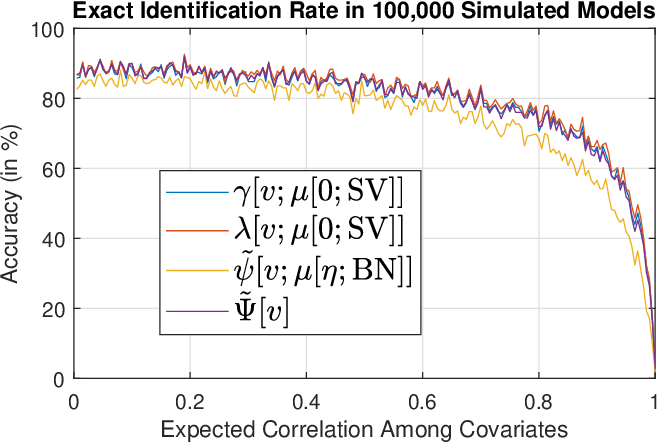}
	\caption{Accuracy $vs$ Expected Correlation Among Regressors}\label{fig:correlated_covariates}
\end{figure}

The accuracy is plotted in Figure \ref{fig:correlated_covariates}.
$\gamma [v;\mu[0; \mathrm{SV}]]$, $\lambda [v;\mu[0; \mathrm{SV}]]$, and  $\tilde \Psi [v]$ have an accuracy of over 80\% when the expected correlation is less than $.7$.
When the expected correlation is close to $1$, the distinction between the covariates becomes blurred, and the selection methods cannot discriminate their contribution. 
Thus, the precision drops sharply to zero.
In contrast,  the precision of $\tilde \psi [v;\mu[\eta; \mathrm{BN}]]$ is about 5.1\% lower.

\subsubsection{\textit{Correlated Errors in Regression}}
\noindent
In this experiment, we investigate how the performances of the NEW4 would change when the error terms in the regression are correlated.
We simulate 100,000 datasets using (\ref{eq:simulate_data}) such that the covariates are randomly correlated.
In each dataset, the expected correlation between any two different residuals is constant. Let the constant be
$\frac{1}{100,001}, \frac{2}{100,001},\cdots,$ and $\frac{100,000}{100,001}$, respectively.
In the regression, we estimate the coefficients $\zeta_i$ using the generalized least squares method to handle the correlation of residuals.
Essentially, we transform both the dependent variable $Y_j$ and the covariates $X_{ji}$ by a linear transformation. 
The transformation matrix is the inverse of the Cholesky component of the variance-covariance matrix of the residuals.

\begin{figure}[!t]
	\centering
	\includegraphics[height=5cm, width=9cm]{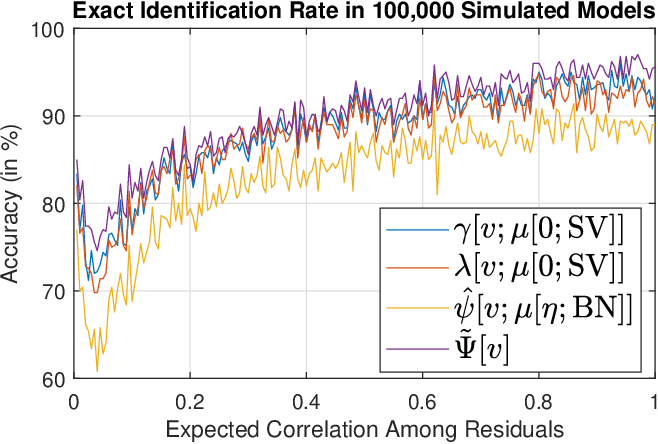}
	\caption{Accuracy $vs$ Expected Correlation Among Residuals}\label{fig:correlated_residuals}
\end{figure}

The exact identification rates are plotted in Figure \ref{fig:correlated_residuals}. 
As the linear transformation causes an alignment problem to the covariates and the dependent variable, the homogeneous correlation of residuals has a negative effect on the identification accuracy.
When the correlation is close to $1$, however, we can regard the residuals as a new variable plus white noise; thus, the negative effect is mitigated, and the identification precision increases.
In general, $\gamma [v;\mu[0; \mathrm{SV}]]$, $\lambda [v;\mu[0; \mathrm{SV}]]$, and $\tilde \Psi [v]$ perform very closely, with an average accuracy of 88.7\%;
$\tilde\psi [v;\mu[\eta; \mathrm{BN}]]$ has a 5.5\% lower accuracy.

\subsubsection{\textit{Sample-size Effect}}

\noindent
In this experiment, we simulate data using (\ref{eq:simulate_data}) and allowing heterogeneous correlation among the covariates. 
We let the sample size of observations to be $100$, $400$, $700$, $\cdots$, $10000$, $40000$, $70000$, and $100000$, respectively.
For each sample size, we generate $1,000$ datasets.
Besides the NEW4, we also apply the BIC, stepwise, Lasso, and adaptive Lasso approaches to the same datasets.

Table \ref{tb:samplesize} lists the numbers of accurately identified models for these sample sizes.
For $\tilde \psi[v;\mu[\eta;\mathrm{BN}]]$, the accuracy reaches 99.7\% when the sample size is larger than 1,000.
For $\tilde \Psi[v]$, the exact identification rate is around 99.5\% when the sample has over 2,000 observations;
it is about 98.5\% for both $\gamma [v;\mu[0; \mathrm{SV}]]$ and $\lambda [v;\mu[0; \mathrm{SV}]]$.
For all the NEW4, the accuracies are robust with varying large sample sizes.
The robustness also indicates that Algorithms \ref{alg:UShapley} and \ref{alg:BN} may not have the oracle properties.

Noticeably, the stepwise procedure has a stable accuracy of around 47\% when the sample size increases. 
The BIC method performs exceptionally well when the sample size is over 10,000. 
However, its computational cost is remarkably high, compared with that for the NEW4 and Lasso. Lastly, Zou (2006) indicates that the adaptive Lasso holds the oracle properties when the covariates are independent. 
The last column of Table \ref{tb:samplesize} implies that independence may be necessary for the oracle properties. 
Correlation obscures the uniqueness of the covariates, confusing the Lasso and adaptive Lasso identification. 
Moreover, the increasing number of observations reinforces the correlation relation. 
Therefore, identification rates decrease.

\begin{table}
\caption{Numbers of Exact Identification in 1,000 Simulated Models$^*$}
\label{tb:samplesize}
\centering
\begin{tabular}{c || c |c |c |c |c |c |c |c}
\hline \hline
Sample Size& $\gamma[\mathrm{SV}]$& $\lambda[\mathrm{SV}]$& $\tilde\psi[\mathrm{BN}]$ & $\tilde\Psi[v]$& BIC & Stepwise& Lasso& aLasso \\ \hline
100&  970&  970&  968&  982&  521&  477&  76&  96\\
400&  982&  985&  988&  992&  814&  469&  37&  40\\
700&  986&  988&  993&  994&  858&  481&  26&  27\\
1,000&  984&  985&  990&  992&  885&  470&  12&  12\\
1,400&  984&  988&  998&  994&  920&  476&  7&  8\\
1,700&  985&  987&  998&  994&  895&  469&  7&  8\\
2,000&  989&  990&  998&  996&  921&  487&  5&  5\\
2,400&  987&  988&  999&  994&  891&  455&  4&  5\\
2,700&  986&  985&  998&  995&  933&  471&  4&  7\\
3,000&  984&  982&  999&  995&  946&  469&  5&  7\\
7,000&  985&  985&  999&  995&  967&  477&  2&  3\\
10,000&  990&  988&  1,000&  995& 981&  480&  1&  1\\
40,000&  985&  989&  1,000&  993&  1,000&  473&  1& 0\\
70,000&  989&  987&  999&  995&  1,000&  460&  0& 1\\
100,000&  987&  981&  999&  994&  1,000&  451&  1& 0\\ \hline \hline
\multicolumn{9}{l}{\small{*:\ In each model, $n=20$, $|\mathbf{S}|=5$, and the covariates are heterogeneously correlated.}}  \\
\multicolumn{9}{l}{\small{*:\ $\gamma[\mathrm{SV}]$ for $\gamma[v;\mu[0;\mathrm{SV}]]$; \ $\lambda[\mathrm{SV}]$ for $\lambda[v;\mu[0;\mathrm{SV}]]$; and $\tilde\psi[\mathrm{BN}]$ for $\tilde\psi[v;\mu[\eta;\mathrm{BN}]]$.}}  \\
\end{tabular}
\end{table}

\subsubsection{\textit{Sizes of Relevant Features}}
\noindent In this last experiment, we let the size of $\mathbf{S}$ vary from $0$ to $n$ while keeping $n=20$.
We still allow that the covariates randomly correlate.

Table \ref{tb:size_of_S} lists the numbers of correctly identified models.
In general, the NEW4 has a slight decreasing accuracy as $|\mathbf{S}|$ increases from $0$ to $n$.
Nevertheless, all have over 95\% accuracy when the ratio of $|\mathbf{S}|$ and $n$ is less than 20\%.
Except for $\tilde\psi[v;\mu[\eta;\mathrm{BN}]]$, the three other NEW4 members maintain above 90\% precision.
In contrast, the accuracies of both BIC and stepwise methods increase with $|\mathbf{S}|$;
they are below 50\% when the relevant features are sparse among the candidates.
Lasso and adaptive Lasso perform exceptionally well when $|\mathbf{S}|$ is $n$ or near $0$. 

Clearly, Algorithm \ref{alg:BN} does not work when $\mathbf{S}=\mathbb{N}$, resulting in a N/A value in Table \ref{tb:size_of_S}. 
An easy fix is to replace $\frac{1}{|\mathbb{R}|}$ with $\frac{1}{2}$ whenever $|\mathbb{R}|=1$ in the algorithm.

\begin{table}
\caption{Exact Identification with Varying $|\mathbf{S}|$ in 1000 Models$^*$}
\label{tb:size_of_S}
\centering
\begin{tabular}{c || c |c |c |c |c |c |c |c}
\hline \hline
$|\mathbf{S}|$& $\gamma[\mathrm{SV}]$& $\lambda[\mathrm{SV}]$& $\tilde\psi[\mathrm{BN}]$ & $\tilde\Psi[v]$& BIC & Stepwise& Lasso& aLasso \\ \hline
0&  982&  979&  993&  989&  385& 378& 540&0 \\
1&  981&  979&  987&  991&  445& 397& 996&996 \\
2&  979&  979&  975&  991&  458& 415& 159&159 \\
3&  980&  979&  978&  993&  483& 433& 143&138 \\
4&  969&  975&  957&  977&  508& 453& 131&126 \\
6&  975&  974&  934&  978&  527& 510&  55&63 \\
7&  974&  976&  941&  977&  559& 531&  27&38 \\
8&  973&  973&  931&  978&  572& 550&  20&28 \\
9&  959&  964&  925&  962&  593& 584&  17&20 \\
10& 956&  962&  918&  963&  607& 604&  14&14 \\
11& 957&  956&  922&  959&  631& 631&  12&9 \\
13& 958&  957&  871&  957&  672& 690&   3&2 \\
14& 937&  939&  781&  934&  675& 711&   0&0 \\
15& 940&  939&  707&  941&  699& 741&   1&1 \\
16& 926&  926&  625&  918&  766& 781&   1&2 \\
17& 937&  943&  270&  936&  820& 823&   0&0 \\
18& 940&  941&  347&  937&  859& 878&   0&0 \\
19& 925& 922&   982&  919&  917& 915&   0&0 \\
20& 926& 926&   N/A&  918&  965& 963& 999&1000 \\ \hline \hline
\multicolumn{9}{l}{\small{*:\ $n=20$; sample size=100; the covariates are heterogeneously correlated.}}  \\
\multicolumn{9}{l}{\small{*:\ $\gamma[\mathrm{SV}]$ for $\gamma[v;\mu[0;\mathrm{SV}]]$; \ $\lambda[\mathrm{SV}]$ for $\lambda[v;\mu[0;\mathrm{SV}]]$; $\tilde\psi[\mathrm{BN}]$ for $\tilde\psi[v;\mu[\eta;\mathrm{BN}]]$.}}  \\
\end{tabular}
\end{table}

\section{Conclusions}

\noindent 
In this paper, we provide a top-down economic-theoretical approach to address a fundamental issue in machine learning, econometrics, and statistics.
When observing the performance $v$ over all subsets of the candidate variables in $\mathbb{N}$, we obtain a coalitional game $(\mathbb{N}, v)$.
For this game, this research supplies three new properties to the Shapley value. 
First, we decompose the value into an expected marginal loss and an expected marginal gain, by assuming a particular type of non-informative priors. 
The Shapley value justly attributes the grand performance $v(\mathbb{N})-v(\emptyset)$, 
but we argue that the variable selection context should attribute the expected performance $\mathbb{E} v(\mathbf{S})-v(\emptyset)$ .
Fortunately, both the marginal loss and marginal gain are fair-division solutions to the expected performance.
Secondly, the expected marginal loss is not half of the Shapley value. 
For a given model size $|\mathbf{S}|$, the average marginal loss is no more than the average marginal gain, due to the diminishing marginality.
To mitigate the asymmetry, we define an unbiased value by unequally weighting the gain and loss.
Lastly, the Shapley value or its unbiased version can be represented as a compound of another class of D-value. 
In this class, the expected model size could be either small or large, compared to the number of available candidates.

Based on the D-value and its unbiased version, we formulate four new feature selection methods.
Our simulations indicate that these methods well succeed in reducing over-fitting and thus identifying the real models.
The success comes from two aspects: simultaneously evaluating the gain and loss of each variable, reducing the first-mover advantage.
The reduction of over-fitting is by deliberate mechanism design, not by the penalty or the trade-off between the model fitting and the number of parameters.
By design, we impose certain functional equations or zero bias in the valuation.
Besides, Algorithms \ref{alg:UShapley} and \ref{alg:BN} are devised to reduce the computational cost in Theorem \ref{thm:sq} and to avoid estimating the nuisance parameter $\eta$.
By consecutively accepting the MVP features, the algorithms acknowledge the dependence of one's performance on already accepted features, focus on the crucial feature at one time, and
realize the opportunity equality by a sequel of conditional equality.
In the algorithms, we make a selection decision by comparing the weighted significance statistics with a cut-off value $t_{_{dof,\xi}}$. 
Thus, the key is the weights. All four methods have different weighting systems derived from different viewpoints.
Unlike a conventional Bayesian model averaging technique, the weights in each method do not sum to $1$. 
The reason for the de-weighting is that the real collective performance $v(\mathbf{S}) - v(\emptyset)$ or its expectation is different from $v(\mathbb{N}) - v(\emptyset)$. 
By de-weighting, we deflate the over-fitting. 
Furthermore, more weights are reduced when the expected model size deviates from the perceived one in the priors.

The framework we provide here opens new areas of applications and suggests improvements on how to apply the Shapley value.  
Our nonparametric methods make few assumptions on the models, the performance function, and the prior knowledge.
It allows for the explanatory variables' heterogeneous effects in different modeling situations, which is the starting point for the D-value. 
Also, our approach addresses the asymmetric sides of the marginal effects.
In a practical application, the estimated model consists of variables that have significant unbiased D-value $\tilde \psi[v;\mu]$.
The selected variables $\hat{\mathbf{S}}$, however, may not have the best collective performance under a specific scenario, e.g., the historical data.
The model with the best performance under the scenario, however, 
does not necessarily produce the best performance under other scenarios, e.g., the future. 
This dilemma leads us to evaluate the variables, not in one specific modeling situation, but various modeling scenarios.

Depending on specific contexts, similar ideas can be applied to other areas in economics, 
political science, and statistics. For example, Hu (2006) applied $\psi[v, [\eta, \mathrm{SV}]]$ and $\psi[v, [0, \mathrm{BV}]]$ to simple games in order to measure the voting power in the games.
In a labor market, Hu (2018) used the aggregate marginal gain and aggregate marginal loss to allocate the net profit between the employed and the unemployed workers.
In detecting outlier observations in statistics, one could apply the unbiased D-value $\tilde \psi [v;\mu]$ to evaluate the overall performance of each observation.
For example, consider the MCD estimator of robust covariance (e.g., Rousseeuw, 1985). 
Among all covariances of subsamples of a given size,
the MCD estimator has the minimum determinant. We let $\mathbb{N}$ be the full sample of observations and
$T$ be a subsample. Let $v(T)$ be the determinant of the covariance of the subsample $T$.
We can reasonably assume no prior discrimination on any subsample among all subsamples of the same size.
A $\mu[\eta; \mathrm{BN}]$ with $\eta \in [.01,.05]$ could be a proper distribution choice 
for the set of outliers observations, and outliers would have significant values in $\tilde \psi[v; \mu[\eta;\mathrm{BN}]]$. 
This example is also an application to unsupervised learning, in which we separate the outliers from non-outliers.

We could extend the D-value from different angles. One way is structural valuations.
In a real valuation situation, it could be proper to specify a context-specific $\mu$.
For example, in a voting context, let $|\mathbf{S}|$ have a uniform distribution on the integers in $\lfloor n/4, 3n/4 \rfloor$, rather than on the integers in $[0,n]$.
A structural valuation can also be done by placing restrictions on $\mu$.
For example, if players $i$ and $j$  never cooperate in the voting, then the probability of  
$\overline{ij} \in \mathbf{S}$ should be zero.
Though the analytic formula is unlikely available for $\psi[v;\mu]$ under a restricted $\mu$, 
the Monte Carlo method is generally feasible.
In the Monte Carlo simulations, we ignore the restricted cases from the sample of
randomly simulated $\mathbf{S}$.
Another type of extension is to alleviate the functional equation (\ref{eq:totality}),  (\ref{eq:divide_expectation_psi}), (\ref{eq:divide_expectation_gamma}), or (\ref{eq:divide_expectation_lambda}),
which requires $2^n$ identity restrictions. 
For example, Hu (2018) only requires the equality holds at the specific $\mathbf{S}$, which does happen.
For another example, we could seek a unique $\mu [\eta, \mathrm{BN}]$, which minimizes the discrepancy between the weights on the left and right sides of (\ref{eq:divide_expectation_psi}),
(\ref{eq:divide_expectation_gamma}), or (\ref{eq:divide_expectation_lambda}).
Finally, when there are more covariates than observations in the regression, then $v(\mathbb{N})$ fails to exist, but the above $n$-player weighting systems are still there. 
The MVP algorithms may use a truncated random ordering,
in which $|\Xi_i^\tau|+1$ has an upper bound set by the sample size of observations.
However, modification of (\ref{eq:variant_tau_phi}) worths further development. 
A similar development may also worth being considered when $\mathbb{N}$ does not contain all relevant features, for example, latent variables.

Efficient computation of the D-value and its unbiased counterpart, however, remains a challenge when $n$ is large. 
Many algorithms in the literature compute the Shapley value; variants of them, especially parallel or graphical computing, could be used to calculate $\psi[v;\mu]$, $\tilde \psi[v;\mu]$, and their components.
Besides, we could also refine the MVP algorithms in Section \ref{sect:est_algorthms}. 
First, one could dynamically determine the number of random orderings based on the already calculated sample mean and sample variance of $\tilde \phi^{\tau}$.
Next, when the computational cost of $\phi^\tau_i$ is high, we could compute it once and save it in a data structure (e.g., a hash table).
The data structure is also usable in calculating $\tilde \psi [v_{_\mathbb{R}}, \mu]$.
Thirdly, when $\phi^\tau_i$ is the absolute t-statistic, we could save all absolute $t$-statistics to the data structure after running the regression on the covariates $\Xi^\tau_i \cup \overline{i}$.
For any variable $j$ in $\Xi^\tau_i \cup \overline{i}$, its absolute t-statistic can be $\phi^{\tau^*}_j$ in some other random ordering $\tau^*$.
Thus, a random ordering can output $\frac{n(n+1)}{2}$ values to the data structure; all of them, scaled by their respective weights, can be used in averaging $\tilde \phi^{\tau}$.
Fourthly, if the reversal of a new random ordering is re-used as the next ordering, then each $i$ \textit{de facto} harvests $n+1$ scaled $\phi^\tau_i$, though correlated, in these two random orderings.
This correlation is not an issue when we estimate the mean of scaled $\phi^\tau_i$.
Finally, in the MVP algorithms  of Section \ref{sect:est_algorthms}, we successively admit one new variable or admit nothing. We could also drop one or drop nothing at the same time.
Accepting and dropping multiple variables at one time could also be possible if their estimated values in $\tilde \psi [v_{_\mathbb{R}}; \mu]$ are far from the cut-off value $t_{_{dof,\xi}}$.
When combined, these modifications reduce the computational cost of the MVP algorithms to $O(|\mathbf{S}|)$. This linear cost allows users to deal with a large number of covariates.
By targeting one feature at one time, the modified MVP algorithms are particularly cost-efficient for a very sparse and high dimensional feature set.

\vskip 1cm
\noindent \textbf{Acknowledgments}

\noindent The author is indebted to the editor, the associate editor, and the referees for their insightful comments and detailed suggestions on the initial submission.
The author is grateful to Lloyd S. Shapley, Stephen Legrand, and Bruce Moses for their advice. 
The author also thanks seminar participants at Central South University, Stony Brook University, and International Monetary Fund for their suggestions.

\vskip 1cm
\noindent \textbf{ORCID}

\noindent Xingwei Hu http://orcid.org/0000-0001-8454-3974

\vskip 1cm
\noindent \textbf{References}

\hangindent=1em
\hangafter=1
\noindent  Banzhaf, J.F. (1965).
Weighted voting doesn't work: a mathematical analysis.
\textit{Rutgers Law Rev.} 19:317-343.

\hangindent=1em
\hangafter=1
\noindent  Clyde, M.,  George, E. I.  (2004).
Model uncertainty.
\textit{Statist. Science} 19:81-94.

\hangindent=1em
\hangafter=1
\noindent  Cohen, S.,  Dror, G.,   Ruppin, E. (2007).
Feature selection via coalitional game theory.
\textit{Neural Computation} 19:1939-1961.

\hangindent=1em
\hangafter=1
\noindent  Devicienti, F. (2010).
Shapley-value decomposition of changes in wage distributions: a note.
\textit{J.  Econ. Inequality}  8:35-45.

\hangindent=1em
\hangafter=1
\noindent Dubey, P.,   Neyman, J.A.,  Weber, R.J. (1981).
Value  without efficiency.
\textit{Math. Operations Research}  6:122-128.

\hangindent=1em
\hangafter=1
\noindent  Freedman, D.A. (1983).
A note on screening regression equations.
\textit{American Statistician} 37:152-155.

\hangindent=1em
\hangafter=1
\noindent  George, E.I.,   McCulloch, R.E. (1997).
Approaches for Bayesian variable selection.
\textit{Statistica Sinica} 7:339-373.

\hangindent=1em
\hangafter=1
\noindent  Gromping, U. (2007).
Estimators of relative importance in linear regression based on variance decomposition.
\textit{American Statistician} 61:1-9.

\hangindent=1em
\hangafter=1
\noindent  Hu,  X. (2002).
Value of loss in n-person games.
UCLA Computational and Applied Mathematics Reports 02-53.

\hangindent=1em
\hangafter=1
\noindent  Hu, X. (2006).
An asymmetric Shapley-Shubik power index.
\textit{Intl.  J.  Game Theory} 34:229-240.

\hangindent=1em
\hangafter=1
\noindent  Hu, X. (2018).
A dichotomous analysis of unemployment benefits.
arXiv:1808.08563.

\hangindent=1em
\hangafter=1
\noindent  Israeli, O. (2007).
A Shapley-based decomposition of the R-Square of a linear regression.
\textit{J.  Econ. Inequality} 5:199-212.

\hangindent=1em
\hangafter=1
\noindent Kahneman, D.,  Knetsch, J. L.,  Thaler, R.H.  (1990).
Experimental tests of the endowment effect and the Coase theorem.
\textit{J.  Polit. Economy} 98:1325-1348.

\hangindent=1em
\hangafter=1
\noindent  Lipovetsky, S.,   Conklin, M. (2001).
Analysis of regression in game theory approach.
\textit{Applied Stochastic Models in Business Industry} 17:319-330.

\hangindent=1em
\hangafter=1
\noindent  Megiddo, N. (1988).
On finding additive, superadditive and subadditive set-function subject to linear inequalities,
RJ 6329, IBM Almaden Research Center: San Jose, California.

\hangindent=1em
\hangafter=1
\noindent   Monderer, D.,   Samet, D. (2002).
Variations on the Shapley value.
In:  \textit{Handbook of Game Theory} Vol 2, edited by R.J. Aumann and S. Hart.
Netherlands: Elsevier Science, pp.2055-2076.

\hangindent=1em
\hangafter=1
\noindent   O'Hara, R.B.,  Sillanpaa, M.J. (2009).
A review of Bayesian variable selection methods: what, how and which.
\textit{Bayesian Analysis} 4:85-118.

\hangindent=1em
\hangafter=1
\noindent Raja, P., Pashyanthi, V., Nandini, U.,  Chakravarthy,  L.S.  (2016).
A novel methodology for classification using Shapley values of cooperative game theory.
\textit{Intl.  J. Computer Science and Mobile Computing} 5(5):60-66.

\hangindent=1em
\hangafter=1
\noindent  Rousseeuw, P.J. (1985).
Multivariate estimation with high break-down point.
In: \textit{Mathematical Statistics and Applications} Vol. B, edited by I. Vincze, W. Grossman, G. Pflug, and W. Wertz.
Budapest: Akademiai Kiado, pp.283-297. 

\hangindent=1em
\hangafter=1
\noindent  Shapley, L.S. (1953).
A value for n-person games.
In: \textit{Annals of Mathematics Studies}, Vol. 28, edited by H. Kuhn and A. Tucker.
Princeton, New Jersey: Princeton University Press, pp.307-317.

\hangindent=1em
\hangafter=1
\noindent Shapley, L.S. Shubik, M. (1954).
A method for evaluating the distribution of power in a committee system. 
\textit{Amer. Polit. Sci. Rev.} 48:787-792.

\hangindent=1em
\hangafter=1
\noindent Strumbelj, E.,  Kononenko, I.  (2010).
An efficient explanation of individual classifications using game theory.
\textit{J.  Machine Learning Research} 11: 1-18.

\hangindent=1em
\hangafter=1
\noindent  Strumbelj, E.,   Kononenko, I. (2014).
Explaining prediction models and individual predictions with feature contributions.
\textit{Knowledge and Information Systems} 41:647-665.

\hangindent=1em
\hangafter=1
\noindent  Thaler, R. (1980).
Toward a positive theory of consumer choice. 
\textit{J.  Econ. Behavior \& Organization} 1:39-60. 

\hangindent=1em
\hangafter=1
\noindent  Tibshirani, R. (1996).
Regression shrinkage and selection via the LASSO.
\textit{J. Roy. Statist. Soc}  B  58:267-288. 

\hangindent=1em
\hangafter=1
\noindent Weber, R.J. (1988).
Probabilistic values for games.
In: \textit{The Shapley Value}, edited by A. E. Roth.
Cambridge, UK: Cambridge University Press, pp. 101-119.

\hangindent=1em
\hangafter=1
\noindent  Winter, E. (2002).
The Shapley value,
In: \textit{Handbook of Game Theory} Vol 2,  edited by R.J. Aumann and S. Hart.
Netherlands: Elsevier Science, pp.2025-2054.

\hangindent=1em
\hangafter=1
\noindent  Zou, H. (2006).
The adaptive Lasso and its Oracle properties.
\textit{J. Amer. Stat. Assoc.} 101(476): 1418-1429.


\clearpage
\newpage
\setcounter{page}{1}
\begin{center}
\Large{
Supplementary Appendix to: \textit{A Theory of Dichotomous Valuation With Applications to Variable Section}
}

\vskip 1cm
Xingwei Hu\footnote{International Monetary Fund, 700 19th St NW, Washington, DC 20431, USA. Email: xhu@imf.org; Phone: (202)623-8317; Fax: (202)589-8317.}
\end{center}

\subsection*{SA-1. Proof of  Theorem \ref{thm:sv}}
\noindent
For any fixed $i\in \mathbb{N}$,  we re-write (\ref{eq:D-Value})  as
\begin{equation}\label{eq:rewrite}\tag{A.1}
\begin{array}{rcl}
\psi_i[v;\mu]&=&
\sum\limits_{_{T\subseteq \mathbb{N}: i \in T}}P_{_T}[v(T)-v(T\setminus\overline{i})]
	+\sum\limits_{_{Z\subseteq \mathbb{N}\setminus \overline{i}}}P_{_Z}[v(Z\cup \overline{i})-v(Z)]\\

&\stackrel{Q=T\setminus \overline{i}}{=}& \sum\limits_{_{T\subseteq \mathbb{N}: i \in  T}} P_{_T} v(T)
	-\sum\limits_{_{Q\subseteq \mathbb{N}\setminus \overline{i}}}  P_{_{Q\cup \overline{i}}} v(Q)
	+\sum\limits_{_{Z\subseteq \mathbb{N}\setminus \overline{i}}}P_{_Z}[v(Z\cup \overline{i})-v(Z)]\\

&\stackrel{Z=Q}{=}& \sum\limits_{_{T\subseteq \mathbb{N}: i \in T}} P_{_T} v(T)
	+ \sum\limits_{_{Z\subseteq \mathbb{N}\setminus \overline{i}}}P_{_Z} v(Z\cup \overline{i})
	-\sum\limits_{_{Z\subseteq \mathbb{N}\setminus \overline{i}}}  [ P_{_Z} + P_{_{Z\cup \overline{i}}}]v(Z)\\

&\stackrel{T=Z\cup \overline{i}}{=}& \sum\limits_{_{T\subseteq \mathbb{N}: i \in T}} P_{_T} v(T)
+ \sum\limits_{_{T\subseteq \mathbb{N}: i \in T}}P_{_{T\setminus \overline{i}}} v(T)
-\sum\limits_{_{Z\subseteq \mathbb{N}\setminus \overline{i}}}  [ P_{_Z} + P_{_{Z\cup \overline{i}}}]v(Z)\\

&\stackrel{T=Z}{=}&\sum\limits_{_{T\subseteq \mathbb{N}: i \in T}}v(T)[P_{_T}+P_{_{T\setminus \overline{i}}}] 
	- \sum\limits_{_{T\subseteq \mathbb{N}\setminus \overline{i}}}v(T)[P_{_T}+P_{_{T\cup\overline{i}}}].
\end{array}
\end{equation}
Therefore, by (\ref{eq:totality}), 
\begin{equation}\label{eq:total_variability}\tag{A.2}
\begin{array}{rcl}
v(\mathbb{N})-v(\emptyset)
&\equiv&\sum\limits_{i\in \mathbb{N}} \ \sum\limits_{T\subseteq \mathbb{N}: i \in T}v(T)[P_{_T} + P_{_{T\setminus \overline{i}}}] 
	- \sum\limits_{i\in \mathbb{N}} \ \sum\limits_{_{T\subseteq \mathbb{N}\setminus \overline{i}}}v(T)[P_{_T} + P_{_{T\cup \overline{i}}}]\\

&\equiv&\sum\limits_{_{T\subseteq \mathbb{N}}} v(T) \sum\limits_{_{i\in T}}[P_{_T} + P_{_{T\setminus \overline{i}}}]
		- \sum\limits_{_{T\subseteq \mathbb{N}}} v(T) \sum\limits_{_{i \in \mathbb{N} \setminus T}}[P_{_T} + P_{_{T\cup \overline{i}}}]\\

&\equiv&\sum\limits_{_{T\subseteq \mathbb{N}}} v(T) \left [|T| P_{_T} + \sum\limits_{_{i\in T}} P_{_{T\setminus \overline{i}}}
			- (n-|T|)P_{_T} - \sum\limits_{_{i\in \mathbb{N} \setminus T}} P_{_{T\cup \overline{i}}} \right ]\\

&\equiv& \sum\limits_{T\subseteq \mathbb{N}} v(T) \left [ \sum\limits_{i\in T} P_{_{T\setminus \overline{i}}} + (2|T|-n)P_{_T}
			- \sum\limits_{i \in \mathbb{N} \setminus T} P_{_{T\cup \overline{i}}} \right ].
\end{array}		
\end{equation}

We compare the coefficients of $v(\mathbb{N})$ and $v(\emptyset)$ in (\ref{eq:total_variability}) to get
\begin{equation}\label{eq:total_value}\tag{A.3}
\left \{
\begin{array}{l}
n\delta_n + \delta_{n-1} = 1, \\
-n\delta_0 - \delta_1  = -1.
\end{array}
\right .
\end{equation}
For any $T\subseteq \mathbb{N}$ such that $T\not = \mathbb{N}$ and $T \not = \emptyset$, the coefficient of
$v(T)$ in (\ref{eq:total_variability}) implies that
$$
\sum\limits_{i\in T}P_{_{T\setminus \overline{i}}} + (2|T|-n) P_{_T} - \sum\limits_{i \in \mathbb{N} \setminus T}P_{_{T\cup \overline{i}}} = 0.
$$

As $\mu \in \mathscr{F}$, $P_{_T} = \frac{(|T|)!(n-|T|)!}{n!} \delta_{_{|T|}}$.
Let $|T|=t$,  $1\le t \le n-1$, and re-write the above equation in terms of $\delta_t$,
$$
t  \frac{(t-1)! (n-t+1)!}{n!} \delta_{_{t-1}}
+ (2t-n) \frac{t! (n-t)!}{n!} \delta_{_t}
- (n-t)   \frac{(t+1)! (n-t-1)!}{n!} \delta_{_{t+1}} 
= 0.
$$
Or simply, for $1\le t \le n-1$,
\begin{equation}\label{eq:adjacentXt}\tag{A.4}
(n-t+1)\delta_{t-1} + (2t-n) \delta_t - (t+1)\delta_{t+1} =0.
\end{equation}

Using (\ref{eq:total_value}), (\ref{eq:adjacentXt}) and $\sum\limits_{t=0}^n \delta_t=1$, we have $n+2$ linear equations of $n+1$ unknowns as
$$
\left [ 
\begin{array}{cccccccccc}
-n&-1            &              &      &      &      &              &  \\
n &\quad 2-n\quad&-2            &      &      &      &              &  \\
  &n-1           &\quad 4-n\quad&-3    &      &      &              &  \\
  &              &              &\ddots&\ddots&\ddots&              &  \\
  &              &              &      &      &2     &\quad n-2\quad&-n\\
  &              &              &      &      &      &1             &n \\
1 &1             &1             &\cdots&\cdots&1     &1             &1
\end{array}
\right ] 
\left (
\begin{array}{c}
\delta_0 	\\
\delta_1 	\\
\delta_2 	\\
\vdots \\
\delta_{n-2} \\
\delta_{n-1} \\
\delta_n \\
\end{array}
\right )
= 
\left (
\begin{array}{c}
-1 \\
0 \\
0 \\
\vdots \\
0 \\
1 \\
1 \\
\end{array}
\right ).
$$

In the $(n+2)\times (n+1)$ coefficient matrix, 
we add the 1st row to the 2nd row,  add the new 2nd row to the 3rd row,  add the new 3rd row to the 4th row, et cetera, and add the new $n$-th row to the $n$+1st row.
Finally, we add $\frac{1}{n}$ of the sum of the first $n$ new rows to the last row. 
In the result matrix, the last two rows are zeros. After removing the last two rows, we have the new system of equations
$$
\left [ 
\begin{array}{cccccccccc}
-n&-1      &              &      &      &     &   &  \\
  &1-n &-2 &    &      &      &               &  \\
  &    &2-n&-3  &      &      &               &  \\
  &             &              &\ddots&\ddots &   &  \\
  &              &              &      &      &-1 &-n
\end{array}
\right ] 
\left (
\begin{array}{c}
\delta_0 	\\
\delta_1 	\\
\vdots \\
\delta_{n-1} \\
\delta_n \\
\end{array}
\right )
= 
\left (
\begin{array}{c}
-1 \\
-1 \\
\vdots \\
-1\\
-1
\end{array}
\right ).
$$

The new $n\times (n+1)$ coefficient matrix has rank $n$ and thus we may write $\delta_0 = \frac{1}{n+1} + \eta$ for some indeterminate $\eta$.
Using mathematical induction, we find the general solution to the system is
\begin{equation} \label{eq:x_i}\tag{A.5}
\delta_t = \frac{1}{n+1} + (-1)^t \left ( \begin{array}{c}n \\ t \end{array} \right ) \eta.
\end{equation}
Any $\eta$ with $|\eta|\le \min\limits_{s=0, 1, ..., n} \frac{s!(n-s)!}{(n+1)!} 
= \frac{(\lfloor \frac{n+1}{2}\rfloor )!  (n - \lfloor \frac{n+1}{2}\rfloor )!}{(n+1)!}$
guarantees the non-negativity of all $\delta_t$ and $\mu[\eta; \mathrm{SV}]$ being a
probability distribution. 
By (\ref{eq:x_i}), for any $T\subseteq \mathbb{N}$,
$$
P_{_T} = \frac{(|T|)!(n-|T|)!}{(n+1)!} + (-1)^{|T|}\eta.
$$

Therefore, 
\begin{equation}\label{eq:prob1}\tag{A.6}
\left \{
\begin{array}{rcll}
P_{_T}+P_{_{T\setminus \overline{i}}}
&=&  \frac{(|T|-1)!(n-|T|)!}{n!}, \quad  & \forall \ T\subseteq \mathbb{N}, \ T\not = \emptyset,\ \forall i \in T; \\

P_{_T}+P_{_{T\cup\overline{i}}} 
&=& \frac{(|T|)!(n-|T|-1)!}{n!}, \quad & \forall \ T\subseteq \mathbb{N}, \ T\not = \mathbb{N}, \ \forall i \not \in T.
\end{array}
\right .
\end{equation}
	
Finally, we plug (\ref{eq:prob1}) into (\ref{eq:rewrite}), 
\begin{equation} \label{eq:combined3}\tag{A.7}
\begin{array}{rcl}
\psi_i[v;\mu]
&=&
\sum\limits_{T\subseteq \mathbb{N}: i \in T} \frac{(|T|-1)!(n-|T|)!}{n!} v(T)
	- \sum\limits_{T\subseteq \mathbb{N} \setminus \overline{i}} \frac{(|T|)!(n-|T|-1)!}{n!} v(T) \\
&\stackrel{Z=T\cup \overline{i}}=& 
\sum\limits_{T\subseteq \mathbb{N}: i \in T} \frac{(|T|-1)!(n-|T|)!}{n!} v(T)
 	- \sum\limits_{Z\subseteq \mathbb{N}: i \in Z} \frac{(|Z|-1)!(n-|Z|)!}{n!} v(Z\setminus \overline{i}) \\
&\stackrel{T=Z}=&
\sum\limits_{T\subseteq \mathbb{N}: i \in T} \frac{(|T|-1)!(n-|T|)!}{n!}[v(T) 
	- v(T\setminus \overline{i})] \\

&=&
\Psi_i [v].  \hspace{1cm} \blacksquare
\end{array}
\end{equation}

\subsection*{SA-2. Proof of  Corollary \ref{cor:svinverse}}
\noindent
If $\mu = \mu[\eta; \mathrm{SV}]$ for some $\eta$, then $P_{_T} = \frac{(|T|)!(n-|T|)!}{(n+1)!} + (-1)^{|T|}\eta$ for all $T \subseteq \mathbb{N}$. 
We repeat (\ref{eq:prob1})$-$(\ref{eq:combined3}) to get $\psi_i[v;\mu] = \Psi_i[v]$ for all $i\in \mathbb{N}$.
 \hspace{1cm} $\blacksquare$

\subsection*{SA-3. Proof of  Theorem \ref{thm:bv}}
\noindent
If $\mu=\mu[\eta; \mathrm{BV}]$ for some $\eta$, then $P_{_T} = \frac{1}{2^n}+(-1)^{|T|} \eta$ for any $T\subseteq \mathbb{N}$. Thus, 
$P_{_T}+ P_{_{T \setminus \overline{i}}} = P_{_T} + P_{_{T \cup \overline{j}}} = \frac{1}{2^{n-1}}$ 
for any $i\in T$ and any $j\not \in T$. By (\ref{eq:rewrite}),
$$
\begin{array}{rcl}
\psi_i[v;\mu] 
&=& \sum\limits_{T\subseteq \mathbb{N}: i \in T} \frac{1}{2^{n-1}} v(T) 
- \sum\limits_{T\subseteq \mathbb{N}: i \not \in T} \frac{1}{2^{n-1}} v(T) \\

&\stackrel{Z=T \cup\overline{i}}{=}&  
\frac{1}{2^{n-1}} \sum\limits_{T\subseteq \mathbb{N}: i \in T} v(T) -  \frac{1}{2^{n-1}} \sum\limits_{Z\subseteq \mathbb{N}: i \in Z} v(Z\setminus \overline{i}) \\

&\stackrel{Z=T}{=}& 
\frac{1}{2^{n-1}} \sum\limits_{T\subseteq \mathbb{N}: i \in T} [v(T) - v(T\setminus \overline{i})] \\ 

&=&
b_i[v].
\end{array}
$$

On the other hand, if $\psi_i [v;\mu] \equiv b_i[v]$ for all $i\in \mathbb{N}$,  then
$$
\begin{array}{rcl}
\psi_i [v;\mu] 
&\equiv& 
\frac{1}{2^{n-1}} \sum\limits_{T\subseteq \mathbb{N}} [v(T)-v(T\setminus \overline{i} )] \\

&\equiv& 
\sum\limits_{T\subseteq \mathbb{N}: i \in T} \frac{1}{2^{n-1}} v(T) - \sum\limits_{T\subseteq \mathbb{N}: i \not \in T} \frac{1}{2^{n-1}} v(T).
\end{array}
$$
By (\ref{eq:rewrite}), 
\begin{equation} \label{eq:TTi}\tag{A.8}
P_{_T}+P_{_{T\setminus \overline{i}}} = \frac{1}{2^{n-1}}, \hspace{.5cm} \forall \ i\in T.
\end{equation}

Without loss of generality, let $P_{_{\emptyset}} = \frac{1}{2^n} + \eta$ for some $\eta$ with $|\eta| \le \frac{1}{2^n}$.
Then, a simple implication of (\ref{eq:TTi}) is $P_{_{\overline{i}}} = \frac{1}{2^{n-1}} - P_{_{\emptyset}} = \frac{1}{2^n} - \eta$ for any $i\in \mathbb{N}$.
For any $T\subseteq \mathbb{N}$ and $T\not = \emptyset$, we write $T$ as $T = \overline{i_1  i_2 i_3 \cdots i_{_{|T|}}}$. 
We consecutively drop two elements from $T$ using (\ref{eq:TTi}),
$$
P_{_T} 
= P_{_{\overline{i_3 i_4 \cdots i_{_{|T|}}}}} 
= \cdots 
= \left \{
\begin{array}{ll}
P_{_{\overline{i_{_{|T|}}}}} = \frac{1}{2^n} - \eta, \quad &\mathrm{if} \ |T| \ \mathrm{is} \ \mathrm{odd};\\
P_{_{\emptyset}} = \frac{1}{2^n} + \eta, \quad &\mathrm{if} \ |T| \ \mathrm{is} \ \mathrm{even}.
\end{array}
\right .
$$
Therefore,
$
P_{_T} = \frac{1}{2^n} +(-1)^{|T|} \eta
$
for any $T\subseteq \mathbb{N}$. \hspace{1cm} $\blacksquare$

\subsection*{SA-4. Proof of  Theorem \ref{thm:binomial}}
\noindent
Using (\ref{eq:gamma}), (\ref{eq:lambda}), and (\ref{eq:D-Value}) 
with $\mu = \mu[\eta; \mathrm{BN}]$, we have
$$
\begin{array}{rcl}
\gamma_i [v; \mu] &=& \sum\limits_{T\subseteq \mathbb{N}: i \in T} \eta^{|T|} (1-\eta)^{n-|T|} \big[v(T)-v(T\setminus \overline{i})\big], \\

\lambda_i [v;\mu] &=& \sum\limits_{Z\subseteq \mathbb{N}: i \not \in Z} \eta^{|Z|} (1-\eta)^{n-|Z|} \big[v(Z \cup \overline{i})-v(Z)\big]\\
&\stackrel{T=Z\cup\overline{i}}{=}& \sum\limits_{T\subseteq \mathbb{N}: i \in T} \eta^{|T|-1} (1-\eta)^{n-|T|+1} \big[v(T)-v(T\setminus \overline{i})\big],\\

\psi_i [v; \mu] 
&=&\sum\limits_{T\subseteq \mathbb{N}: i \in T} 
\left [\eta^{|T|} (1-\eta)^{n-|T|} + \eta^{|T|-1} (1-\eta)^{n-|T|+1} \right ]
 \big[v(T)-v(T\setminus \overline{i})\big]  \\
 
&=&\sum\limits_{T\subseteq \mathbb{N}: i \in T}  \eta^{|T|-1} (1-\eta)^{n-|T|} \big[v(T)-v(T\setminus \overline{i})\big].  
\end{array}
$$
Therefore, $\psi [v; \mu] \equiv \frac{1}{\eta} \gamma [v; \mu] \equiv \frac{1}{1-\eta}\lambda [v;\mu]$. \hspace{1cm} $\blacksquare$

\subsection*{SA-5. Proof of Theorem \ref{thm:divide_expectation_lambda}}
\noindent
For any $\mu \in \mathscr{F}$,
$$
\mathbb{E} v(\mathbf{S}) 
= \sum\limits_{T\subseteq \mathbb{N}}  P_{_T} v(T)
= \sum\limits_{T\subseteq \mathbb{N}}  \frac{(|T|)! (n-|T|)!}{n!} \delta_{_{|T|}} v(T).
$$

\noindent
\fbox{Part I:} We write the total expected marginal loss in terms of $v(T)$,
$$
\begin{array}{rcl}
\sum\limits_{i\in \mathbb{N}} \lambda_i[v;\mu]
&=&
\sum\limits_{i\in \mathbb{N}} \ \sum\limits_{T \subseteq \mathbb{N}: T\not= \mathbb{N}, i\not \in T} 
P_{_T} \left [ v(T \cup \overline{i}) - v(T) \right ]  \\

&=&
\sum\limits_{i\in \mathbb{N}} \ \sum\limits_{T \subseteq \mathbb{N}: T\not= \mathbb{N}, i\not \in T} 
P_{_T} v(T \cup \overline{i}) 
- 
\sum\limits_{i\in \mathbb{N}} \ \sum\limits_{T \subseteq \mathbb{N}: T\not= \mathbb{N}, i\not \in T} 
P_{_T} v(T) \\

&\stackrel{Z=T \cup \overline{i}}{=}&
\sum\limits_{i\in \mathbb{N}} \ \sum\limits_{Z \subseteq \mathbb{N}: Z\not=\emptyset, i\in Z}
P_{_{Z\setminus \overline{i}}} v(Z)
- 
\sum\limits_{T \subseteq \mathbb{N}: T\not= \mathbb{N}} v(T)
\sum\limits_{i \in \mathbb{N} \setminus T} P_{_T}  \\

&\stackrel{T=Z}{=}&
\sum\limits_{T \subseteq \mathbb{N}: T\not=\emptyset} v(T) 
\sum\limits_{i \in T} P_{_{T\setminus \overline{i}}}
- 
\sum\limits_{T \subseteq \mathbb{N}: T\not= \mathbb{N}} 
\frac{(n-|T|)(|T|)!(n-|T|)!\delta_{_{|T|}}}{n!}
v(T)  \\

&=&
\sum\limits_{T \subseteq \mathbb{N}: T\not=\emptyset, T\not = \mathbb{N}} 
\frac{|T| (|T|-1)! (n-|T|+1)!\delta_{_{|T|-1}}}{n!}
v(T) +  \delta_{n-1} v(\mathbb{N}) \\

&&
- \sum\limits_{T \subseteq \mathbb{N}: T\not= \mathbb{N}, T\not=\emptyset} 
\frac{(n-|T|)(|T|)!(n-|T|)!\delta_{_{|T|}}}{n!}
v(T) -  
n\delta_0 v(\emptyset) \\

&=&
\sum\limits_{T \subseteq \mathbb{N}: T\not=\emptyset, T\not=\mathbb{N}} 
\frac{(|T|)! (n-|T|)! \left [(n-|T|+1)\delta_{_{|T|-1}}-(n-|T|)\delta_{_{|T|}} \right ]}{n!}
v(T) \\

&&
+ 
\delta_{n-1} v(\mathbb{N}) 
-  
n\delta_0 v(\emptyset).
\end{array}
$$
\noindent
For (\ref{eq:divide_expectation_lambda}) to hold, we compare the coefficients of $v(T)$ on both sides of
(\ref{eq:divide_expectation_lambda}):
\begin{equation} \label{eq:total_lambda_mu}\tag{A.9}
\left \{
\begin{array}{rcl}
\delta_{n-1} 
&=& \delta_n, \\

-n\delta_0
&=&
\delta_0 - 1, \\

(n-t+1)\delta_{t-1}-(n-t)\delta_t
&=&
\delta_t, \hspace{1.5cm} \forall \ 1\le t \le n-1.
\end{array}
\right .
\end{equation}
The solution to the above system is $\delta_0 = \delta_1 = \cdots = \delta_n = \frac{1}{n+1}$, i.e.,
$\mu = \mu[0; \mathrm{SV}]$.

\noindent
\fbox{Part II:} We write the total expected marginal gain in terms of $v(T)$,
$$
\begin{array}{rcl}
\sum\limits_{i\in \mathbb{N}} \gamma_i[v;\mu]
&=&
\sum\limits_{i\in \mathbb{N}} \ \sum\limits_{T \subseteq \mathbb{N}: T\not= \emptyset, i \in T} 
P_{_T} \left [ v(T) - v(T \setminus  \overline{i}) \right ]  \\

&=&
\sum\limits_{i\in \mathbb{N}} \ \sum\limits_{T \subseteq \mathbb{N}: T\not= \emptyset, i \in T} 
P_{_T} v(T) 
- 
\sum\limits_{i\in \mathbb{N}} \ \sum\limits_{T \subseteq \mathbb{N}: T\not=\emptyset, i\in T} 
P_{_T} v(T\setminus  \overline{i}) \\

&\stackrel{Z=T \setminus \overline{i}}{=}&
\sum\limits_{i\in \mathbb{N}} \ \sum\limits_{T \subseteq \mathbb{N}: T\not=\emptyset, i\in T}
P_{_{T}} v(T)
- 
\sum\limits_{Z \subseteq \mathbb{N}: Z\not = \mathbb{N}} v(Z)
\sum\limits_{i \in \mathbb{N} \setminus Z} P_{_{Z\cup  \overline{i}}}  \\

&\stackrel{T=Z}{=}&
\sum\limits_{T \subseteq \mathbb{N}: T\not=\emptyset} v(T) 
\sum\limits_{i \in T} P_{_{T}}
- 
\sum\limits_{T \subseteq \mathbb{N}: T\not= \mathbb{N}} 
\frac{(n-|T|)(|T|+1)!(n-|T|-1)!\delta_{_{|T|+1}}}{n!}
v(T)  \\

&=&
\sum\limits_{T \subseteq \mathbb{N}: T\not=\emptyset, T\not = \mathbb{N}} 
\frac{|T| (|T|)! (n-|T|)!\delta_{_{|T|}}}{n!} v(T) + n \delta_n v(\mathbb{N}) \\

&&
- 
\sum\limits_{T \subseteq \mathbb{N}: T\not= \mathbb{N}, T\not=\emptyset} 
\frac{(|T|+1)!(n-|T|)!\delta_{_{|T|+1}}}{n!}v(T) 
-  \delta_1 v(\emptyset) \\

&=&
\sum\limits_{T \subseteq \mathbb{N}: T\not=\emptyset, T\not=\mathbb{N}} 
\frac{(|T|)! (n-|T|)! \left [ |T| \delta_{_{|T|}} - (|T|+1)\delta_{_{|T|+1}} \right ]}{n!}
v(T) \\

&&
+ n \delta_n v(\mathbb{N})  -  \delta_1 v(\emptyset).
\end{array}
$$
\noindent
For (\ref{eq:divide_expectation_gamma}) to hold, we compare the coefficients of $v(T)$ on both sides of
(\ref{eq:divide_expectation_gamma}):
\begin{equation} \label{eq:total_gamma_mu}\tag{A.10}
\left \{
\begin{array}{rcl}
n \delta_n
&=& \delta_n, \\

-\delta_1
&=&
\delta_0 - 1, \\

t \delta_t - (t+1)\delta_{t+1}
&=&
\delta_t, \hspace{1.5cm} \forall \ 1\le t \le n-1.
\end{array}
\right .
\end{equation}
The solution to the above system is $\delta_0 + \delta_1 = 1$ and $\delta_2 = \delta_3 = \cdots = \delta_n = 0$.

\noindent
\fbox{Part III:}
To prove the last part, we capitalize on the relation $\sum\limits_{i\in \mathbb{N}} \psi_i [v; \mu] = \sum\limits_{i\in \mathbb{N}} \gamma_i[v;\mu] + \sum\limits_{i\in \mathbb{N}} \lambda_i[v;\mu].$
For (\ref{eq:divide_expectation_psi}) to hold, we use (\ref{eq:total_lambda_mu}) and (\ref{eq:total_gamma_mu}) to get
\begin{equation} \label{eq:total_psi_mu}\tag{A.11}
\left \{
\begin{array}{rcl}
\delta_{n-1} + n \delta_n
&=& \delta_n, \\

-  n\delta_0  - \delta_1 
&=&
\delta_0 - 1, \\

 (n-t+1)\delta_{t-1} - (n-t)\delta_t  + t \delta_t - (t+1)\delta_{t+1}
&=&
\delta_t, \hspace{1cm} \forall \ 1\le t \le n-1.
\end{array}
\right .
\end{equation}
The first equation in (\ref{eq:total_psi_mu}), together with $\delta_n\ge 0$ and $\delta_{n-1}\ge 0$, implies that  $\delta_n = \delta_{n-1}=0$ unless $n=1$. 
We then apply $\delta_n=\delta_{n-1}=0$ to the third equation in (\ref{eq:total_psi_mu}), 
i.e. $\delta_{t-1}=\frac{(t+1)\delta_{t+1}-(2t-n-1)\delta_t}{n-t+1}$,  for $t=n-1, n-2,\cdots, 1$.
We get all $\delta_t = 0$, which contradicts to the second equation in (\ref{eq:total_psi_mu}).
Therefore, there exists no solution $\mu \in \mathscr{F}$ which solves (\ref{eq:divide_expectation_psi}).  \hspace{1cm} $\blacksquare$

\subsection*{SA-6. Proof of Theorem \ref{thm:expected_SV}}
\noindent
If $i\not \in Z \subseteq \mathbb{N}$, then for any $T\subseteq \mathbb{N}$,
$$
\begin{array}{rcl}
v_{_Z}(T\cup \overline{i})- v_{_Z}(T)
&=&
v \large (Z \cap (T\cup \overline{i}) \large ) - v (Z \cap T ) \\
&=&
v \large ((Z \cap T) \cup (Z \cap \overline{i}) \large  ) - v (Z \cap T ) \\
&=&
v \large ((Z \cap T) \cup \emptyset) \large ) - v (Z \cap T ) \\
&=&
0.
\end{array}
$$
Thus, $i$ is a dummy player in $v_{_Z}$ and $\Psi_i[v_{_Z}]=0$.
When $i \in Z$, its Shapley value in $v_{_Z}$ is
$$
\begin{array}{rcl}
\Psi_i[v_{_Z}]
&=&
\sum\limits_{W\subseteq \mathbb{N}: i\in W}  \frac{(|W|-1)!(n-|W|)!}{n!} \left [ v_{_Z}(W) - v_{_Z} (W \setminus \overline{i}) \right ] \\
&=&
\sum\limits_{W\subseteq \mathbb{N}: i\in W}  \frac{(|W|-1)!(n-|W|)!}{n!} \left [ v(Z\cap W) - v ((Z\cap W) \setminus \overline{i}) \right ]\\
&\stackrel{T=Z\cap W}{=}&
\sum\limits_{T\subseteq Z: i\in T} \frac{v(T) - v (T \setminus \overline{i})}{n!} \sum\limits_{W\subseteq \mathbb{N}:W \cap Z = T}  (|W|-1)!(n-|W|)! \\
&\stackrel{U=W\setminus T}{=}&
\sum\limits_{T\subseteq Z: i\in T} \frac{v(T) - v (T \setminus \overline{i})}{n!} \sum\limits_{U\subseteq \mathbb{N}\setminus Z}  (|T|+|U|-1)!(n-|T|-|U|)! \\
&=&
\sum\limits_{T\subseteq Z: i\in T} \frac{v(T) - v (T \setminus \overline{i})}{n!} \sum\limits_{u=0}^{n-|Z|}\sum\limits_{U\subseteq \mathbb{N}\setminus Z: |U|=u}  (|T|+u-1)!(n-|T|-u)! \\
&=&
\sum\limits_{T\subseteq Z: i\in T} \frac{v(T) - v (T \setminus \overline{i})}{n!}\ \sum\limits_{u=0}^{n-|Z|}\ (|T|+u-1)!(n-|T|-u)! 
\left ( \begin{array}{c} n-|Z| \\ u \end{array} \right ) \\
&=&
\sum\limits_{T\subseteq Z: i\in T} \frac{(n-|Z|)! (|T|-1)! (|Z|-|T|)!}{n!}\left [v(T)-v (T \setminus \overline{i}) \right ] \\
&&
\sum\limits_{u=0}^{n-|Z|}\ \left ( \begin{array}{c} |T|+u -1 \\ u \end{array} \right ) \left ( \begin{array}{c} n-|T|-u \\ n-|Z| - u\end{array}   \right )\\
&=&
\sum\limits_{T\subseteq Z: i\in T}   \frac{(n-|Z|)! (|T|-1)! (|Z|-|T|)!}{n!}\left [v(T)-v (T \setminus \overline{i}) \right ] 
\left ( \begin{array}{c} n \\ n-|Z| \end{array}   \right )  \\
&=&
\sum\limits_{T\subseteq Z: i \in T} \frac{(|T|-1)!(|Z|-|T|)!}{(|Z|)!} \left [ v(T) - v(T\setminus \overline{i})\right ].
\end{array}
$$
In the next to last equality, we use Lemma 1 in Hu (2006) to simplify the sum of products of two binomials.

Given the prior $\mu[0; \mathrm{SV}]$ for $\mathbf{S}$, we apply the above identities of $\Psi_i[v_{_Z}]$ to get
$$
\begin{array}{rcl}
\mathbb{E} \Psi_i [v_{_\mathbf{S}}] 
&=&
\sum\limits_{Z\subseteq \mathbb{N}} P_{_Z} \Psi_i[v_{_Z}] \\
&=&
\sum\limits_{Z\subseteq \mathbb{N}: i \not \in Z} P_{_Z} \Psi_i[v_{_Z}] + \sum\limits_{Z\subseteq \mathbb{N}: i \in Z} P_{_Z} \Psi_i[v_{_Z}] \\
&=&
0 + \sum\limits_{Z\subseteq \mathbb{N}: i\in Z} \frac{(|Z|)!(n-|Z|)!}{(n+1)!}\ \sum\limits_{T\subseteq Z: i \in T} \frac{(|T|-1)!(|Z|-|T|)!}{(|Z|)!} \left [ v(T) - v(T\setminus \overline{i})\right ] \\
&=&
\sum\limits_{T\subseteq \mathbb{N}: i \in T} \frac{(|T|-1)!}{(n+1)!} \left [v(T) - v(T\setminus \overline{i})\right ]\ \sum\limits_{Z\subseteq \mathbb{N}: Z \supseteq T} (n-|Z|)! (|Z|-|T|)!  \\
&=&
\sum\limits_{T\subseteq \mathbb{N}: i \in T} \frac{(|T|-1)!}{(n+1)!} \left [v(T) - v(T\setminus \overline{i})\right ] \sum\limits_{z=|T|}^n \ \sum\limits_{Z\subseteq \mathbb{N}: |Z|=z, Z \supseteq T} (n-z)! (z-|T|)!  \\
&=&
\sum\limits_{T\subseteq \mathbb{N}: i \in T} \frac{(|T|-1)!}{(n+1)!} \left [v(T) - v(T\setminus \overline{i})\right ] \sum\limits_{z=|T|}^n  (n-z)! (z-|T|)!
\left ( 
\begin{array}{c}
n-|T| \\
z-|T|
\end{array}  
\right )
\\
&=&
\sum\limits_{T\subseteq \mathbb{N}: i \in T} \frac{(|T|-1)!(n-|T|+1)!}{(n+1)!} \left [ v(T) - v(T\setminus \overline{i})\right ].  
\end{array}
$$
Also, using $\mu[0; \mathrm{SV}]$ and (\ref{eq:lambda}), we have
$$
\begin{array}{rcl}
\lambda_i [v, \mu[0; \mathrm{SV}] ]
&=&
\sum\limits_{Z\subseteq \mathbb{N}: i \not \in Z} P_{_Z} [v(Z\cup \overline{i}) - v(Z)] \\
&\stackrel{T=Z\cup \overline{i}}{=}&
\sum\limits_{T\subseteq \mathbb{N}: i \in T} P_{_{T\setminus \overline{i}}} [v(T) - v(T\setminus \overline{i})] \\
&=&
\sum\limits_{T\subseteq \mathbb{N}: i \in T} \frac{(|T|-1)!(n-|T|+1)!}{(n+1)!} \left [ v(T) - v(T\setminus \overline{i})\right ].  
\end{array}
$$
Therefore, $\mathbb{E} \Psi_i [v_{_\mathbf{S}}]=\lambda_i [v, \mu[0; \mathrm{SV}] ].$ 
Finally, the sum of the weights  for the marginal gain $\left [ v(T) - v(T\setminus \overline{i})\right ]$ is
$$
\begin{array}{rcl}
\sum\limits_{T\subseteq \mathbb{N}: i \in T} \frac{(|T|-1)!(n-|T|+1)!}{(n+1)!} 
&=&
\sum\limits_{t=1}^n \ \sum\limits_{T\subseteq \mathbb{N}: |T|=t, i \in T} \frac{(|T|-1)!(n-|T|+1)!}{(n+1)!}  \\
&=&
\sum\limits_{t=1}^n  \frac{(t-1)!(n-t+1)!}{(n+1)!}  
\left ( 
\begin{array}{c}
n-1 \\
t-1
\end{array}  
\right )
\\
&=&
\frac{1}{n(n+1)} \sum\limits_{t=1}^n \left ( n-t+1 \right ) \\
&=&
\frac{1}{2}.  \hspace{1cm} \blacksquare
\end{array}
$$

\subsection*{SA-7. Proof of  Lemma \ref{lm:bias}}
$$
\begin{array}{rcl}
\kappa_i [v;\mu] 
&=& \sum\limits_{T\subseteq \mathbb{N}: i \in T} P_{_T} \left [ v(T) - v(T\setminus \overline{i}) \right ]
- \sum\limits_{Z\subseteq \mathbb{N}: i \not \in Z} P_{_Z} \left [ v(Z\cup \overline{i}) - v(Z) \right ] \\

&\stackrel{T=Z\cup \overline{i}}{=}&
\left [ \sum\limits_{T\subseteq \mathbb{N}: i \in T} P_{_T}  v(T) 
+  \sum\limits_{Z\subseteq \mathbb{N}: i \not \in Z} P_{_Z} v(Z) \right ]
-  \sum\limits_{T\subseteq \mathbb{N}: i \in T}   P_{_T} v(T\setminus \overline{i}) \\
&&
- \sum\limits_{T\subseteq \mathbb{N}: i \in T} P_{_{T\setminus \overline{i}}} v(T) \\

&\stackrel{Z=T\setminus \overline{i}}{=}& 
\sum\limits_{T\subseteq \mathbb{N}} P_{_T}  v(T)
-  \sum\limits_{Z\subseteq \mathbb{N}:i\not \in Z}   P_{_{Z\cup \overline{i}}} v(Z)
- \sum\limits_{T\subseteq \mathbb{N}} P_{_{T\setminus \overline{i}}} v(T) 
+ \sum\limits_{T\subseteq \mathbb{N}:i\not\in T} P_{_{T\setminus \overline{i}}} v(T) \\

&=& 
\sum\limits_{T\subseteq \mathbb{N}} \left [P_{_T} - P_{_{T\setminus \overline{i}}} \right ] v(T)
+ \left(\sum\limits_{Z\subseteq \mathbb{N}: i \in Z} -  \sum\limits_{Z\subseteq \mathbb{N}} \right) P_{_{Z\cup \overline{i}}} v(Z) 
+ \sum\limits_{T\subseteq \mathbb{N}:i\not\in T} P_{_{T}} v(T) \\

&\stackrel{T=Z}{=}& 
\sum\limits_{T\subseteq \mathbb{N}} \left [P_{_T}  - P_{_{T\setminus \overline{i}}} -  P_{_{T\cup \overline{i}}} \right ] v(T)
+ \left[\sum\limits_{Z\subseteq \mathbb{N}: i \in Z}   P_{_{Z}} v(Z) + \sum\limits_{T\subseteq \mathbb{N}:i\not\in T} P_{_{T}} v(T) \right] \\

&=& 
\sum\limits_{T\subseteq \mathbb{N}} \left [2 P_{_T} - P_{_{T \cup \overline{i}}} - P_{_{T \setminus \overline{i}}} \right ] v(T).  \hspace{1cm} \blacksquare
\end{array}
$$

\subsection*{SA-8. Proof of  Theorem \ref{thm:bias_Banzhaf}}
\noindent
\fbox{Part I:}
If $\mu = \mu [0; \mathrm{BV}]$, then $P_{_T} = \frac{1}{2^n}$ for any $T\subseteq \mathbb{N}$. 
Thus, $2 P_{_T} - P_{_{T \cup \overline{i}}} - P_{_{T \setminus \overline{i}}}=0$ 
for any $i\in \mathbb{N}$ and  $T\subseteq \mathbb{N}$. 
By Lemma \ref{lm:bias},
$$
\kappa_i [v;\mu] 
= \sum\limits_{T\subseteq \mathbb{N}} \left [2 P_{_T} - P_{_{T \cup \overline{i}}} - P_{_{T \setminus \overline{i}}} \right ] v(T)
= 0
$$
for any $i \in \mathbb{N}$.

On the other hand, if  $\kappa_i [v;\mu] \equiv 0$ for all $i\in \mathbb{N}$, then Lemma \ref{lm:bias} implies that
$2 P_{_T} - P_{_{T \cup \overline{i}}} - P_{_{T \setminus \overline{i}}} = 0$ for any $i\in \mathbb{N}$ and any $T\subseteq \mathbb{N}$. A simply implication is that 
$P_{_T}=P_{_{T \setminus \overline{i}}}$ for any $i\in T \subseteq \mathbb{N}$. 
For any $T\subseteq \mathbb{N}$ and $T\not = \emptyset$, we write $T$ as
$T = \overline{i_1  i_2 \cdots i_{_{|T|}}}$. Now
$$
P_{_T} 
= P_{_{\overline{i_2 i_3 \cdots i_{_{|T|}}}}} 
= P_{_{\overline{i_3 i_4 \cdots i_{_{|T|}}}}} 
= \cdots 
= P_{_{\overline{i_{_{|T|}}}}} 
= P_{_{\emptyset}}.
$$
As $\sum\limits_{T\subseteq \mathbb{N}} P_{_T}=1$,  we have $2^n P_{_{\emptyset}}=1$ and  $P_{_T} =  P_{_{\emptyset}} = \frac{1}{2^n}$ for any $T\subseteq \mathbb{N}$.
Therefore, $\mu = \mu [0; \mathrm{BV}]$.

\noindent
\fbox{Part II:}
If $\mu = \mu[0; \mathrm{BV}]$, then  Lemma \ref{lm:bias} implies that
$\kappa_i[v;\mu]=0$ for all $i \in \mathbb{N}$.
Therefore, $\sum\limits_{i\in \mathbb{N}} \kappa_i[v;\mu]=0$.

On the other hand, we assume $\sum\limits_{i\in \mathbb{N}} \kappa_i[v;\mu] \equiv 0$ and
$\mu\in \mathscr{F}$. Using $P_{_T} = \frac{(|T|)!(n-|T|)! \delta_{_{|T|}}}{n!}$
and Lemma \ref{lm:bias}, we simplify the aggregate bias:
\begin{equation} \label{eq:total_bias}\tag{A.12}
\begin{array}{rcl}
\sum\limits_{i\in \mathbb{N}} \kappa_i [v;\mu]
&=& 
\sum\limits_{T\subseteq \mathbb{N}} v(T) \sum\limits_{i\in \mathbb{N}} \left [2 P_{_T} - P_{_{T \cup \overline{i}}} - P_{_{T \setminus \overline{i}}} \right ] \\

&=& 
\sum\limits_{T\subseteq \mathbb{N}} v(T) \left[
2\sum\limits_{i\in \mathbb{N}} P_{_T} 
- \left(\sum\limits_{i\in T}+\sum\limits_{i\in \mathbb{N}\setminus T}\right) P_{_{T \cup \overline{i}}} 
- \left(\sum\limits_{i\in \mathbb{N}\setminus T}+\sum\limits_{i\in T}\right) P_{_{T \setminus \overline{i}}} 
\right ] \\

&=& \sum\limits_{T\subseteq \mathbb{N}} v(T) \left [ 2n P_{_T} - |T| P_{_{T}} 
-\sum\limits_{i \in \mathbb{N} \setminus T} P_{_{T \cup \overline{i}}} 
- (n-|T|) P_{_{T}} - \sum\limits_{i \in T} P_{_{T \setminus \overline{i}}} \right ] \\

&=& \sum\limits_{T\subseteq \mathbb{N}} v(T) \left [ n P_{_T} 
-\sum\limits_{i \in \mathbb{N} \setminus T} P_{_{T \cup \overline{i}}} 
- \sum\limits_{i \in T} P_{_{T \setminus \overline{i}}} \right ] \\

&=& \sum\limits_{T\subseteq \mathbb{N}:T\not = \mathbb{N}, T\not = \emptyset}  \frac{(|T|)!(n-|T|)! \left [ n \delta_{_{|T|}}
-(|T|+1) \delta_{_{|T|+1}} - (n-|T|+1) \delta_{_{|T|-1}} \right ] }{n!}  v(T) \\
&& + (n\delta_n - \delta_{n-1}) v(\mathbb{N}) + (n\delta_0 - \delta_{1}) v(\emptyset).  \\
\end{array}
\end{equation}

Therefore, by $\sum\limits_{i\in \mathbb{N}} \kappa_i[v;\mu] \equiv 0$,
$$
\left \{
\begin{array}{rcl}
n\delta_n - \delta_{n-1} &=& 0, \\
n\delta_0 - \delta_{1} &=& 0, \\
n \delta_t -(t+1) \delta_{_{t+1}} - (n-t+1) \delta_{_{t-1}}&=&0, \quad \forall \ t=1,2,...,n-1.
\end{array}
\right .
$$

By mathematical induction on $t$, it is not hard to see that $\delta_t = \frac{n!}{t!(n-t)!} \delta_0$ for any $t=1,2,...,n$. 
Finally, as  $\sum\limits_{t=0}^n \delta_t = 1$, we get $\delta_0 = \frac{1}{2^n}$ and
$$
P_{_T} =  \frac{(|T|)!(n-|T|)!}{n!}\delta_{_{|T|}} =  \frac{(|T|)!(n-|T|)!}{n!}\frac{n!}{(|T|)!(n-|T|)!}\delta_0 =  \frac{1}{2^n}.  
$$
Therefore, $\mu = \mu [0; \mathrm{BV}]$. \hspace{1cm} $\blacksquare$

\subsection*{SA-9. Proof of  Theorem \ref{thm:pi_omega}}
\noindent
For any $t=0,1,\cdots,n-1$, by the definitions of $\pi_t(v)$ and $ \omega_t(v)$, 
$$
\begin{array}{rcl}
\pi_t(v) &=& \frac{t!(n-t-1)!}{n!} \sum\limits_{T\subseteq \mathbb{N}: |T|=t}\ \sum\limits_{i\in \mathbb{N} \setminus T} \big[v(T\cup \overline{i}) - v(T) \big] \\

&\stackrel{Z = T\cup \overline{i}}{=}& \frac{ ((t+1)-1)!(n-(t+1))!}{n!} 
	\sum\limits_{Z\subseteq \mathbb{N}: |Z|=t+1}\ \sum\limits_{i \in Z} \big[v(Z) - v(Z\setminus \overline{i}) \big] \\
&=& \omega_{t+1}(v).
\end{array}
$$
Therefore, $\omega_t(v) \ge \pi_t(v)$ for all $1\le t\le n-1$ if and only if $\omega_t (v) \ge \omega_{t+1}(v)$ for all $1\le t\le n-1$;
equivalently, $\omega_t(v)$ is a decreasing function of $t$.
In the same fashion, $\omega_t (v) \ge \pi_t (v)$ for all $1\le t\le n-1$ if and only if $\pi_{t-1}(v) \ge \pi_t(v)$ for all $1\le t\le n-1$;
equivalently, $\pi_t(v)$ is a decreasing function of $t$. \hspace{1cm} $\blacksquare$

\subsection*{SA-10. Proof of  Theorem \ref{thm:diminishing}}
\noindent
\fbox{Part I:} \
As $\mu = \mu[0; \mathrm{SV}]$, $P_{_T} = \frac{(|T|)!(n-|T|)!}{(n+1)!}$ and  
$$
\begin{array}{rcl}
\sum\limits_{i\in \mathbb{N}} \kappa_i [v;\mu]
&=&
\sum\limits_{i\in N} \left \{ \sum\limits_{T\subseteq \mathbb{N}: i \in T} P_{_T}  \big[v(T) - v(T\setminus \overline{i})\big] 
-  \sum\limits_{Z\subseteq \mathbb{N}: i \not\in Z} P_{_Z}  \big[v(Z\cup \overline{i}) - v(Z)\big] \right \} \\

&=&  
\sum\limits_{T\subseteq \mathbb{N}: T\not = \emptyset} P_{_T} 
		\sum\limits_{i \in T} \big[v(T) - v(T\setminus \overline{i})\big] 
 - \sum\limits_{Z\subseteq \mathbb{N}: Z\not = \mathbb{N}}  P_{_Z} 
		\sum\limits_{i \in \mathbb{N} \setminus Z} \big[ v(Z \cup \overline{i}) - v(Z) \big] \\

&=&  
\sum\limits_{t=1}^n \ \sum\limits_{T\subseteq \mathbb{N}: |T| = t} \frac{(|T|)!(n-|T|)!}{(n+1)!} 
		\sum\limits_{i \in T} \big[v(T) - v(T\setminus \overline{i})\big] \\

& &
 - \sum\limits_{z=0}^{n-1} \ \sum\limits_{Z\subseteq \mathbb{N}: |Z|=z}  \frac{(|Z|)!(n-|Z|)!}{(n+1)!} 
		\sum\limits_{i \in \mathbb{N} \setminus Z} \big[ v(Z \cup \overline{i}) - v(Z) \big] \\

&=&  
\sum\limits_{t=1}^n \frac{t}{n+1} \omega_t(v)
	- \sum\limits_{z=0}^{n-1} \frac{n-z}{n+1} \pi_z(v)  \\

&=&  
\sum\limits_{t=1}^n \frac{t}{n+1} \omega_t(v) - \sum\limits_{t=1}^n \frac{n+1-t}{n+1} \omega_t (v)   \hspace{1cm} \mathrm{(by} \ \mathrm{Theorem} \ \ref{thm:pi_omega} \mathrm{)}\\

&=&
 \sum\limits_{t=1}^n \frac{2t}{n+1} \omega_t(v) - \sum\limits_{t=1}^n \omega_t(v) \\

&=& n \left [ \frac{1}{n} \sum\limits_{t=1}^n \left (\frac{2t}{n+1} \right ) \omega_t (v)
		- \left (\frac{1}{n} \sum\limits_{t=1}^n \frac{2t}{n+1} \right ) \left (\frac{1}{n}
		 \sum\limits_{t=1}^n \omega_t (v) \right )  \right  ]
\end{array}
$$
which is the sample covariance, multiplied by $n$, between the series $\left \{ \frac{2t}{n+1} \right \}_{t=1}^n$ and the series
$\left \{ \omega_t(v) \right \}_{t=1}^n$. 
As $v$ has diminishing marginality, $\omega_t(v)$ is decreasing in $t$.
But $\frac{2t}{n+1}$ is increasing in $t$. Therefore,
the sample covariance is non-positive, i.e., $\sum\limits_{i\in \mathbb{N}} \kappa_i [v; \mu] \le 0$.

\noindent
\fbox{Part II:} \
If $v$ is super-additive, then $v(\emptyset)=0$ (e.g., Megiddo, 1988). Denote 
$$
\begin{array}{rcl}
\Delta_1 &\stackrel{\mathrm{def}}{=}& \sum\limits_{T\subseteq \mathbb{N}: T\not = \emptyset} (|T|)!(n-|T|)! \sum\limits_{i \in T} 
\big[ v(T) - v(T\setminus \overline{i}) \big],  \\

\Delta_2 &\stackrel{\mathrm{def}}{=}& \sum\limits_{Z\subseteq \mathbb{N}: Z\not = \mathbb{N}} (|Z|)!(n-|Z|)! \sum\limits_{i \in \mathbb{N} \setminus Z} 
\big[ v(Z \cup \overline{i}) - v(Z) \big].
\end{array}
$$
Note that $\Delta_1$ equals
$$
\begin{array}{rcl}
&&\sum\limits_{T\subseteq \mathbb{N}: T\not = \emptyset} |T|(|T|)!(n-|T|)! v(T) -
\sum\limits_{T\subseteq \mathbb{N}: T\not = \emptyset}\ \sum\limits_{i \in T} (|T|)!(n-|T|)! v(T\setminus \overline{i})  \\

&\stackrel{Z=T\setminus\overline{i}}{=}&\sum\limits_{T\subseteq \mathbb{N}: T\not = \emptyset} |T|(|T|)!(n-|T|)! v(T)
 - \sum\limits_{Z\subseteq \mathbb{N}: Z\not = \mathbb{N}}\ \sum\limits_{i \in \mathbb{N} \setminus Z} (|Z|+1)!(n-|Z|-1)! v(Z) \\

&=&\sum\limits_{T\subseteq \mathbb{N}: T\not = \emptyset} |T|(|T|)!(n-|T|)! v(T) 
	- \sum\limits_{Z\subseteq \mathbb{N}: Z\not = \mathbb{N}}	(|Z|+1)!(n-|Z|)! v(Z),
\end{array}
$$
and $\Delta_2$ equals
$$
\begin{array}{rcl}
&&\sum\limits_{Z\subseteq \mathbb{N}: Z\not = \mathbb{N}} (|Z|)!(n-|Z|)! \sum\limits_{i \not \in Z} v(Z\cup \overline{i})
-\sum\limits_{Z\subseteq \mathbb{N}: Z\not = \mathbb{N}} (|Z|)!(n-|Z|)! (n-|Z|) v(Z) \\
&\stackrel{T=Z\cup \overline{i}}{=}&\sum\limits_{T\subseteq \mathbb{N}: T\not = \emptyset} (|T|-1)!(n-|T|+1)! \sum\limits_{i \in T} v(T) 
-\sum\limits_{Z\subseteq \mathbb{N}: Z\not = \mathbb{N}} (|Z|)!(n-|Z|)! (n-|Z|) v(Z)\\
&=&\sum\limits_{T\subseteq \mathbb{N}: T\not = \emptyset} (|T|)!(n-|T|+1)! v(T) 
-\sum\limits_{Z\subseteq \mathbb{N}: Z\not = \mathbb{N}} (|Z|)!(n-|Z|)! (n-|Z|) v(Z).
\end{array}
$$

Using $v(\emptyset)=0$ and $v(T)+v(\mathbb{N}\setminus T)\le v(\mathbb{N})$, we can write $\Delta_1 - \Delta_2$ as
$$
\begin{array}{rcl}
&& \sum\limits_{T\subseteq \mathbb{N}: T\not = \emptyset} (|T|)!(n-|T|)!(2|T|-n-1) v(T) \\
&&	+ \sum\limits_{Z\subseteq \mathbb{N}: Z\not = \mathbb{N}}	(|Z|)!(n-|Z|)!(n-2|Z|-1) v(Z) \\

&=& n!(n-1) v(\mathbb{N}) + \sum\limits_{T\subseteq \mathbb{N}:T\not = \emptyset, T\not = \mathbb{N}} (|T|)!(n-|T|)!(2|T|-n) v(T) \\

&&	+ \sum\limits_{Z\subseteq \mathbb{N}: Z\not = \mathbb{N}, Z\not = \emptyset} (|Z|)!(n-|Z|)!(n-2|Z|) v(Z)\\

&&  - \sum\limits_{T\subseteq \mathbb{N}:T\not = \emptyset, T\not = \mathbb{N}} (|T|)!(n-|T|)! v(T) 
	-\sum\limits_{Z\subseteq \mathbb{N}: Z\not = \mathbb{N}, Z\not = \emptyset} (|Z|)!(n-|Z|)! v(Z)\\

&=& n!(n-1) v(\mathbb{N}) - \sum\limits_{T\subseteq \mathbb{N}:T\not = \emptyset, T\not = \mathbb{N}} (|T|)!(n-|T|)! v(T) \\
&& -\sum\limits_{Z\subseteq \mathbb{N}: Z\not = \mathbb{N}, Z\not = \emptyset} (|Z|)!(n-|Z|)! v(Z)\\

&\stackrel{T=\mathbb{N}\setminus Z}{=}&  n!(n-1) v(\mathbb{N}) - \sum\limits_{T\subseteq \mathbb{N}: T\not = \emptyset, T\not = \mathbb{N}} (|T|)!(n-|T|)! \big[ v(T)+v(\mathbb{N}\setminus T) \big] \\

&\ge& n!(n-1) v(\mathbb{N}) - \sum\limits_{T\subseteq \mathbb{N}: T\not = \emptyset, T\not = \mathbb{N}} (|T|)!(n-|T|)! v(\mathbb{N})   
 \hspace{.3cm} \mathrm{(by} \ \mathrm{superadditivity)}\\

&=& n!(n-1) v(\mathbb{N}) - v(\mathbb{N}) \sum\limits_{t=1}^{n-1}\ \sum\limits_{T\subseteq \mathbb{N}: |T|=t} (|T|)!(n-|T|)! \\

&=& n!(n-1) v(\mathbb{N}) - v(\mathbb{N}) \sum\limits_{t=1}^{n-1} t!(n-t)! \left ( \begin{array}{c} n \\ t \end{array} \right )  \\

&=&
0.\\
\end{array}
$$
Finally, we plug $\Delta_1$ and $\Delta_2$ into the second equality in the proof of Part I to get
$$
\sum\limits_{i\in \mathbb{N}} \kappa_i[v;\mu] = \frac{\Delta_1}{(n+1)!} - \frac{\Delta_2}{(n+1)!} = \frac{\Delta_1-\Delta_2}{(n+1)!} \ge 0.  \hspace{1cm} \blacksquare
$$

\subsection*{SA-11. Proof of  Theorem \ref{thm:binomialbias}}
\noindent
Given $\mu = \mu[\eta; \mathrm{BN}]$, by Theorem \ref{thm:binomial} and (\ref{eq:def_alpha}), 
$$
\begin{array}{rclrcl}
\kappa_i [v;\mu]
&=& \gamma_i[v;\mu] - \lambda_i [v;\mu]   
\hspace{1cm}&
\alpha
&=& {\sum\limits_{i\in \mathbb{N}} \kappa_i[v;\mu]}\ / \ {\sum\limits_{i\in \mathbb{N}} \psi_i[v;\mu]} \\

&=& \eta \psi_i[v;\mu] - (1-\eta) \psi_i[v;\mu] \hspace{1cm}
&&=& {\sum\limits_{i\in \mathbb{N}} (2\eta-1) \psi_i[v;\mu]}\ / \ {\sum\limits_{i\in \mathbb{N}} \psi_i[v;\mu]} \\

&=& (2\eta-1) \psi_i[v;\mu], 
&&=& 2\eta-1. 
\end{array}
$$
Furthermore, by (\ref{eq:unbiasedValue}) and  Theorem \ref{thm:binomial},   
$$
\begin{array}{rcl}
\tilde \psi_i[v;\mu]
&=& \left [ 1-(2\eta-1)\right ] \gamma_i[v;\mu] + \left [ 1+(2\eta-1) \right ] \lambda_i [v;\mu] \\
&=& 2(1-\eta) \eta \psi_i[v;\mu] + 2\eta (1-\eta) \psi_i[v;\mu]   \\
&=& 4 \eta (1-\eta) \psi_i[v;\mu] \\
&=& 4 \sum\limits_{T\subseteq \mathbb{N}: i \in T} \eta^{|T|} (1-\eta)^{n-|T|+1} \left[v(T)-v(T\setminus \overline{i}) \right].
 \hspace{1cm} \blacksquare
\end{array}
$$

\subsection*{SA-12. Proof of  Theorem \ref{thm:UShapleyValueWeights}}
\noindent
\fbox{Part I:}\  The weight on the marginal gain $[v(T)-v(T\setminus \overline{i})]$ in
the unbiased Shapley value $\tilde \Psi_i[v]$ defined in (\ref{eq:unbiased_Shapley}) is $\frac{4(|T|)!(n-|T|+1)!}{(n+2)!}$ 
for any $T \ni i$. The total weight is then
$$
\begin{array}{rcl}
\sum\limits_{T\subseteq \mathbb{N}: i \in T} \frac{4(|T|)!(n-|T|+1)!}{(n+2)!}
&=&  \sum\limits_{t=1}^n\ \sum\limits_{T\subseteq \mathbb{N}: |T|=t, i \in T} \frac{4t!(n-t+1)!}{(n+2)!}   \\
&=&  \sum\limits_{t=1}^n  \frac{4t!(n-t+1)!}{(n+2)!}  \left ( \begin{array}{c} n-1 \\ t-1 \end{array}  \right ) \\
&=&  \sum\limits_{t=1}^n  \frac{4(tn-t^2+t)}{n(n+1)(n+2)}\\
&=&  \frac{4n(n+1)n}{2n(n+1)(n+2)}  - \frac{4n(n+1)(2n+1)}{6n(n+1)(n+2)} + \frac{4n(n+1)}{2n(n+1)(n+2)} \\
&=& \frac{2}{3}. \hspace{1cm} 
\end{array}
$$

\noindent
\fbox{Part II:}\  Using (\ref{eq:unbiased_Shapley}), we write $\sum\limits_{i \in \mathbb{N}} \tilde \Psi_i[v]$ as
$$
\begin{array}{rcl}
&& 
\sum\limits_{i \in \mathbb{N}}\ \sum\limits_{T\subseteq \mathbb{N}: i \in T} \frac{4 (|T|)! (n-|T|+1)!}{ (n+2)!} \big[ v(T)-v(T\setminus \overline{i}) \big] \\

&\stackrel{Z=T\setminus \overline{i}}{=}& 
\sum\limits_{T\subseteq \mathbb{N}: T\not = \emptyset}\ \sum\limits_{i \in T} \frac{4 (|T|)! (n-|T|+1)!}{(n+2)!} v(T) 
- \sum\limits_{Z\subseteq \mathbb{N}: Z\not = \mathbb{N}}\ \sum\limits_{i \not \in Z} \frac{4 (|Z|+1)! (n-|Z|)!}{(n+2)!} v(Z) \\

&\stackrel{T=Z}{=}& 
\sum\limits_{T\subseteq \mathbb{N}: T\not = \emptyset} \frac{4 |T| (|T|)! (n-|T|+1)!}{(n+2)!} v(T) 
- \sum\limits_{T\subseteq \mathbb{N}: T\not = \mathbb{N}} \frac{4 (n-|T|) (|T|+1)! (n-|T|)!}{(n+2)!} v(T) \\

&=& 
\frac{4 n n!}{(n+2)!} \left [v(\mathbb{N}) - v(\emptyset) \right ] 
+ \sum\limits_{T\subseteq \mathbb{N}: T\not = \mathbb{N}, T\not = \emptyset} \frac{4 (|T|)! (n-|T|)! (2|T|-n)!}{(n+2)!} v(T) \\

&=& 
\frac{4 n n!}{(n+2)!} \left [v(\mathbb{N}) - v(\emptyset) \right ] 
+ \left [\sum\limits_{T: \frac{n}{2} < |T| < n} + \sum\limits_{T: 0 < |T| < \frac{n}{2}}  \right ]\frac{4 (|T|)! (n-|T|)! (2|T|-n)!}{(n+2)!} v(T) \\

&\stackrel{Z=\mathbb{N}\setminus T}{=}& 
\frac{4 n n!}{(n+2)!} \left [v(\mathbb{N}) - v(\emptyset) \right ] 
+ \sum\limits_{T\subseteq \mathbb{N}: \frac{n}{2} < |T| < n} \frac{4 (|T|)! (n-|T|)! (2|T|-n)!}{(n+2)!} v(T) \\
&&
+ \sum\limits_{Z\subseteq \mathbb{N}: \frac{n}{2} < |Z| < n} \frac{4 (n-|Z|)! (|Z|)! (n-2|Z|)!}{(n+2)!} v(\mathbb{N}\setminus Z) \\

&\stackrel{T=Z}{=}&
\sum\limits_{T \subseteq \mathbb{N}: |T| > \frac{n}{2}}  \frac{4(|T|)!(n-|T|)!(2|T|-n)}{(n+2)!} \left [ v(T) - v(\mathbb{N} \setminus T)\right ]. 
\end{array}
$$
In addition, the weight $g(|T|) \eqdef \frac{4(|T|)!(n-|T|)!(2|T|-n)}{(n+2)!}$ increases in $|T|$ as the ratio
$$
\frac{g(t+1)}{g(t)} = \frac{4 (t+1)! (n-t-1)!(2t+2-n)}{4 t! (n-t)! (2t-n)} =  \frac{t+1}{n-t}\left [ 1+\frac{2}{2t-n}\right ] > 1 * 1 =  1
$$
when $t > \frac{n}{2}$. Besides, the sum of the weights on $\left[ v(T) - v(\mathbb{N} \setminus T)\right ]$ is
$$
\begin{array}{rcl}
\sum\limits_{T \subseteq \mathbb{N}: |T| > \frac{n}{2}}  \frac{4(|T|)!(n-|T|)!(2|T|-n)}{(n+2)!} 
&=&
\sum\limits_{\frac{n}{2}<t\le n}  \ \sum\limits_{T \subseteq \mathbb{N}: |T|=t}  \frac{4t!(n-t)!(2t-n)}{(n+2)!}  \\
&=&
\sum\limits_{\frac{n}{2}<t\le n}   \frac{4t!(n-t)!(2t-n)}{(n+2)!}  \left ( \begin{array}{c} n \\ t \end{array} \right ) \\
&=&
\sum\limits_{\frac{n}{2}<t\le n}  \frac{4(2t-n)}{(n+1)(n+2)} \\
&=&
\left \{
\begin{array}{rl}
\frac{n}{n+1}, & \mathrm{if}\ n \ \mathrm{is\ even;}\\
\frac{n+1}{n+2}, & \mathrm{if}\ n \ \mathrm{is\ odd.}\\
\end{array}
\right .
\end{array}
$$
Therefore, the sum is close to $1$ when $n$ is large.\hspace{1cm} $\blacksquare$

\subsection*{SA-13. Proof of  Theorem \ref{thm:ShapleyValue}}
\noindent
\fbox{(v) $\Longrightarrow$ (i) :} \
If $\theta=\rho=1$, then by (\ref{eq:shapley_value}) and (\ref{eq:binomialBeta}), $\psi_i[v;\mu] \equiv \Psi_i[v]$ for all $i\in \mathbb{N}$.

\noindent
\fbox{(i) $\Longrightarrow$ (v) :} \
By (\ref{eq:shapley_value}) and (\ref{eq:binomialBeta}), $\psi [v; \mu] \equiv \Psi [v]$ if and only if
\begin{equation}\label{eq:ocean}\tag{A.13}
\frac{\beta(\theta+|T|-1, \rho+n-|T|)}{\beta (\theta, \rho)} = \frac{(|T|-1)!(n-|T|)!}{n!}
\end{equation}
for all $T\subseteq  \mathbb{N}, T\not = \emptyset$. 
For any $T\subseteq \mathbb{N}$ with $1\le |T|\le n-1$ and any $i \not \in T$, we apply  $Z = T\cup \overline{i}$ 
to equation (\ref{eq:ocean}),
\begin{equation}\label{eq:ocean1}\tag{A.14}
\frac{\beta(\theta+|Z|-1, \rho+n-|Z|)}{\beta (\theta, \rho)} = \frac{(|Z|-1)!(n-|Z|)!}{n!}
\end{equation}
Using $|Z|=|T|+1$, we write the ratios between (\ref{eq:ocean}) and (\ref{eq:ocean1}),
$$
\frac{ \beta (\theta+|T|-1, \rho+n-|T|)}{\beta  (\theta+|T|, \rho+n-|T|-1)} 
= \frac{(|T|-1)!(n-|T|)!} {(|T|)!(n-|T|-1)!}.
$$
Using the property of beta function, we simplify the above ratios to get
$
n (\theta-1) = |T| (\theta + \rho -2)
$
for all $T\subseteq \mathbb{N}$ with $T\not = \emptyset$ and $T\not = \mathbb{N}$. 
Therefore,  $\theta=\rho=1$.

\noindent
\fbox{(i) $\Longrightarrow$ (ii) :} \ This follows from the efficiency property of the Shapley value.

\noindent
\fbox{(ii) $\Longrightarrow$ (i) :}  \ As $\mu \in \mathscr{F}$, (i) follows from Theorem  \ref{thm:sv}.

\noindent
\fbox{(v) $\Longrightarrow$ (iii) :} \ By (\ref{eq:probabilitydensity}), 
the aggregate expected marginal loss $\sum\limits_{i\in \mathbb{N}} \lambda_i [v;\mu]$ is
\begin{equation}\label{eq:value_of_unemployed}\tag{A.15}
\begin{array}{rcl}
&&
\sum\limits_{i\in \mathbb{N}}\ \sum\limits_{T\subseteq \mathbb{N}\setminus \overline{i}} 
\frac{\beta(\theta+|T|, \rho+n-|T|)}{\beta(\theta,\rho)}[v(T \cup \overline{i})-v(T)] \\

&\stackrel{Z=T\cup\overline{i}}=&
\sum\limits_{Z\subseteq \mathbb{N}: Z\not = \emptyset}\ \sum\limits_{i\in Z} \frac{\beta(\theta+|Z|-1, \rho+n-|Z|+1)}{\beta(\theta,\rho)}v(Z)
- \sum\limits_{T\subseteq \mathbb{N}}\ \sum\limits_{i\in \mathbb{N}\setminus T} \frac{\beta(\theta+|T|, \rho+n-|T|)}{\beta(\theta,\rho)} v(T) \\

&\stackrel{T=Z}=& \sum\limits_{T\subseteq \mathbb{N}: T\not = \emptyset} 
\frac{|T| \beta(\theta+|T|-1, \rho+n-|T|+1) - (n-|T|)\beta(\theta+|T|, \rho+n-|T|)}{\beta(\theta,\rho)} v(T) 
 - \frac{n\beta(\theta,\rho+n)}{\beta(\theta,\rho)}v(\emptyset)\\

&=& \sum\limits_{T\subseteq \mathbb{N}: T\not = \emptyset} 
\left [ \frac{|T|(\rho+n-|T|)}{\theta+|T|-1} - (n - |T|) \right ]
\frac{\beta(\theta+|T|, \rho+n-|T|)}{\beta(\theta,\rho)} v(T)
- \frac{n\beta(\theta,\rho+n)}{\beta(\theta,\rho)}v(\emptyset).
\end{array}
\end{equation}

When $\theta=\rho=1$,
$$
\begin{array}{rcl}
\sum\limits_{i\in \mathbb{N}} \lambda_i[v;\mu] 
&=&
\sum\limits_{T\subseteq \mathbb{N}: T\not = \emptyset} \frac{ \beta (|T|+1, n+1-|T|)}{\beta (1,1)} v(T)  
-  \frac{n \beta(1, n+1)}{\beta (1,1)} v(\emptyset) \\

&=& 
\sum\limits_{T\subseteq \mathbb{N}: T\not = \emptyset} \frac{(|T|)!(n-|T|)!}{(n+1)!} v(T) 
- \frac{n}{n+1} v(\emptyset) \\
&=& \sum\limits_{T\subseteq \mathbb{N}} \frac{(|T|)!(n-|T|)!}{(n+1)!} v(T) - v(\emptyset).
\end{array}
$$

 Note that  by (\ref{eq:probabilitydensity}),

\begin{equation}\label{eq:totalexpectedvalue}\tag{A.16}
\mathbb{E} v(\mathbf{S})= \sum\limits_{T\subseteq \mathbb{N}} \frac{\beta (\theta+|T|, \rho+n-|T|)}{\beta (\theta, \rho)} v(T).
\end{equation}

When $\theta = \rho = 1$, it reduces to 
$$
\mathbb{E} v(\mathbf{S}) = \sum\limits_{T\subseteq \mathbb{N}} \frac{(|T|)!(n-|T|)!}{(n+1)!} v(T)
= \sum\limits_{i\in \mathbb{N}} \lambda_i[v;\mu]  +  v(\emptyset).
$$

\noindent
\fbox{(iii) $\Longrightarrow$ (v) :} \ 
For  $\sum\limits_{i\in \mathbb{N}} \lambda_i[v;\mu] \equiv \mathbb{E} v(\mathbf{S}) -v(\emptyset)$  to hold,  by (\ref{eq:value_of_unemployed}) and (\ref{eq:totalexpectedvalue}), we need 
$$
\left \{
\begin{array}{rcll}
\frac{|T|(\rho+n-|T|)}{\theta+|T|-1} - (n - |T|) &\equiv&1, \qquad & \forall\ T\subseteq \mathbb{N}, T \not = \emptyset; \\
- \frac{n\beta (\theta, \rho+n)}{\beta (\theta, \rho)} &\equiv& \frac{\beta (\theta, \rho+n)}{\beta (\theta, \rho)} - 1 .&
\end{array}
\right .
$$
The solution to the above system is $\theta = \rho=1$.

\noindent
\fbox{(iii) $\Longrightarrow$ (iv) :} \  From the above proof, we have already shown the equivalence of (i), (ii), (iii) and (v). Therefore, (iii) implies that (ii). Thus
$$
\sum\limits_{i\in \mathbb{N}} \gamma_i[v;\mu] = \sum\limits_{i\in \mathbb{N}} \psi_i[v;\mu] - \sum\limits_{i\in \mathbb{N}} \lambda_i[v;\mu] 
= [v(\mathbb{N}) - v(\emptyset)] - [\mathbb{E}  v(\mathbf{S})  - v(\emptyset)]  = v(\mathbb{N}) - \mathbb{E} v(\mathbf{S}).
$$

\noindent
\fbox{(iv) $\Longrightarrow$ (v) :} \
By (\ref{eq:probabilitydensity}), the aggregate expected marginal gain
$\sum\limits_{i\in \mathbb{N}}  \gamma_i [v;\mu]$ is
\begin{equation}\label{eq:value_of_employed}\tag{A.17}
\begin{array}{rcl}
&&
\sum\limits_{i\in \mathbb{N}}\ \sum\limits_{T\subseteq \mathbb{N}: i \in T} \frac{\beta(\theta+|T|, \rho+n-|T|)}{\beta(\theta,\rho)}[v(T)-v(T\setminus\overline{i})] \\

&\stackrel{Z=T\setminus\overline{i}}=&
\sum\limits_{T\subseteq \mathbb{N}}\ \sum\limits_{i\in T} \frac{\beta(\theta+|T|, \rho+n-|T|)}{\beta(\theta,\rho)}v(T)
- \sum\limits_{Z\subseteq \mathbb{N}: Z\not = \mathbb{N}}\ \sum\limits_{i \in \mathbb{N}\setminus Z}  \frac{\beta(\theta+|Z|+1, \rho+n-|Z|-1)}{\beta(\theta,\rho)} v(Z) \\

&\stackrel{T=Z}=& 
\sum\limits_{T\subseteq \mathbb{N}}\frac{|T| \beta(\theta+|T|, \rho+n-|T|)}{\beta(\theta,\rho)}v(T)
- \sum\limits_{T\subseteq \mathbb{N}: T\not = \mathbb{N}} \frac{(n-|T|) \beta(\theta+|T|+1, \rho+n-|T|-1)}{\beta(\theta,\rho)} v(T) \\

&=& 
\sum\limits_{T\subseteq \mathbb{N}: T\not = \mathbb{N}}\left [  |T| - \frac{(n-|T|) (\theta+|T|)}{\rho+n-|T|-1} \right ] \frac{\beta(\theta+|T|, \rho+n-|T|)}{\beta(\theta,\rho)}v(T) + \frac{n\beta(\theta+n,\rho)}{\beta(\theta,\rho)}v(\mathbb{N}).

\end{array}
\end{equation}

Thus, for $\sum\limits_{i\in \mathbb{N}}  \gamma_i [v;\mu] \equiv v(\mathbb{N}) - \mathbb{E} v(\mathbf{S})$ to hold, by (\ref{eq:totalexpectedvalue}) and (\ref{eq:value_of_employed}), we need
$$
\left \{
\begin{array}{rcll}
\frac{n\beta(\theta+n,\rho)}{\beta (\theta, \rho)} &\equiv&1 - \frac{\beta(\theta+n,\rho)}{\beta(\theta,\rho)},& \\
|T| - \frac{(n-|T|) (\theta+|T|)}{\rho+n-|T|-1}        &\equiv&-1, \hspace{2cm} & \forall \ T\subseteq \mathbb{N}, T\not = \mathbb{N}.
\end{array}
\right .
$$
The solution to the above system is $\theta=\rho=1$.  \hspace{1cm} $\blacksquare$

\subsection*{SA-14. Proof of  Theorem \ref{thm:sq}}
\noindent
As the ordering $\tau$ has the uniform distribution over $\Omega$, each ordering occurs with probability $\frac{1}{n!}$. 
Moreover, there are $(|\Xi_i^\tau|)!$ permutations in $\Xi_i^\tau$ and $(n-1-|\Xi_i^\tau|)!$ permutations in $\mathbb{N}\setminus \Xi_i^\tau \setminus \overline{i}$, the set of elements
preceded by $i$ in the ordering $\tau$.
Thus, the probability of $\Xi_i^\tau = T$ is $\frac{(|\Xi_i^\tau|)!(n-1-|\Xi_i^\tau|)!}{n!} = \frac{(|T|)!(n-1-|T|)!}{n!}$.
Using (\ref{eq:variant_tau_phi}) and the law of total expectation, we have
$$
\begin{array}{rcl}
\mathbb{E} [\tilde \phi^\tau_i] 
&=& 
\sum\limits_{T\subseteq \mathbb{N}\setminus \overline{i}} \mathrm{Prob} ( \Xi_i^\tau=T) \mathbb{E} \left [ \tilde \phi^\tau_i  \  | \ \Xi_i^\tau=T   \right ] \\

&=&  
\sum\limits_{T\subseteq \mathbb{N}\setminus \overline{i}}  \frac{(|T|)!(n-1-|T|)!}{n!}  \frac{n! (P_{_T}+P_{T \cup \overline{i}})}{(|T|)!(n-|T| - 1)!} \left [v(T \cup \overline{i}) - v(T) \right ] \\

&=& 
\sum\limits_{T\subseteq \mathbb{N}\setminus \overline{i}}  (P_{_T}+P_{_{T \cup \overline{i}}}) \left [v(T \cup \overline{i}) - v(T) \right ] \\

&=& 
\sum\limits_{T\subseteq \mathbb{N}\setminus \overline{i}}  P_{_T} \left [v(T \cup \overline{i}) - v(T) \right ] 
+ \sum\limits_{T\subseteq \mathbb{N}\setminus \overline{i}} P_ {_{T \cup \overline{i}}} 
\left [v(T \cup \overline{i}) - v(T) \right ]  \\

&\stackrel{Z=T\cup\overline{i}}=& 
\lambda_i [v;\mu] 
+ \sum\limits_{Z\subseteq \mathbb{N}: i \in Z} P_{_Z} \left [v(Z) - v(Z \setminus \overline{i}) \right ]  \\

&=& 
\lambda_i [v;\mu] + \gamma_i [v;\mu] \\

&=&
\psi_i[v;\mu] .
\end{array}
$$
The above proof also implies (\ref{eq:sequel_phi}), by considering the two probability components in (\ref{eq:variant_tau_phi}) seperately. 

When $\mu \in \mathscr{F}$,
$$
\left \{
\begin{array}{rcl}
\frac{n! P_{_{\Xi_i^\tau \cup \overline{i}}}}{(|\Xi_i^\tau|)!(n-|\Xi_i^\tau| - 1)!}
&=& \frac{n! \frac{ (|\overline{i} \cup \Xi_i^\tau|)! (n-|\overline{i} \cup \Xi_i^\tau|)!}{n!} \delta_{|\overline{i} \cup \Xi_i^\tau|} }{(|\Xi_i^\tau|)!(n-|\Xi_i^\tau| - 1)!}  = (|\Xi_i^\tau|+1) \delta_{|\Xi_i^\tau|+1},  \\ \medskip

\frac{n! P_{_{\Xi_i^\tau}}}{(|\Xi_i^\tau|)!(n-|\Xi_i^\tau| - 1)!}
&=&\frac{n! \frac{ (| \Xi_i^\tau|)! (n-| \Xi_i^\tau|)!}{n!}   \delta_{|\Xi_i^\tau|}   }{(|\Xi_i^\tau|)!(n-|\Xi_i^\tau| - 1)!} = (n-|\Xi_i^\tau|) \delta_{|\Xi_i^\tau|}.
\end{array}
\right .
$$

For the Shapley value, let $\mu = \mu [\mathrm{SV}, 0]$. Then,
$$
\left \{
\begin{array}{rcl}
\frac{n! P_{_{\Xi_i^\tau \cup \overline{i}}}}{(|\Xi_i^\tau|)!(n-|\Xi_i^\tau| - 1)!}
&=& \frac{n! \frac{ (|\overline{i} \cup \Xi_i^\tau|)! (n-|\overline{i} \cup \Xi_i^\tau|)!}{(n+1)!}}{(|\Xi_i^\tau|)!(n-|\Xi_i^\tau| - 1)!}  = \frac{|\Xi_i^\tau|+1}{n+1},  \\ \medskip
\frac{n! P_{_{\Xi_i^\tau}}}{(|\Xi_i^\tau|)!(n-|\Xi_i^\tau| - 1)!}
&=&\frac{n! \frac{ (| \Xi_i^\tau|)! (n-| \Xi_i^\tau|)!}{(n+1)!}}{(|\Xi_i^\tau|)!(n-|\Xi_i^\tau| - 1)!} = \frac{n-|\Xi_i^\tau|}{n+1}.
\end{array}
\right .
$$

For the unbiased Shapley value, the right-hand side of (\ref{eq:unbiasedShapleyValue}) is
$$
\begin{array}{rcl}
&& 
4 \sum\limits_{T\subseteq \mathbb{N}\setminus \overline{i}} \mathrm{Prob} ( \Xi_i^\tau=T)\ \mathbb{E} \left \{ \frac{(|\Xi_i^\tau|+1)(n-|\Xi_i^\tau|)}{(n+1)(n+2)} [v(\Xi_i^\tau \cup \overline{i}) - v(\Xi_i^\tau) ]   \Bigm\vert  \Xi_i^\tau=T   \right \} \\

&=& 
4 \sum\limits_{T\subseteq \mathbb{N}\setminus \overline{i}}  \frac{(|T|)!(n-1-|T|)!}{n!} \frac{(|T|+1)(n-|T|)}{(n+1)n+2)} \left [v(T \cup \overline{i}) - v(T) \right ] \\

&\stackrel{Z=T\cup \overline{i}}{=}&  
4 \sum\limits_{Z\subseteq \mathbb{N}: i \in Z}  \frac{(|Z|)!(n-|Z|+1)!}{(n+2)!}  \left [v(Z) - v(Z\setminus \overline{i}) \right ] \\

&=&
\tilde \Psi_i[v]. \hspace{1cm} \blacksquare
\end{array}
$$
\end {document}